\documentclass[10pt,twocolumn,letterpaper]{article}

\usepackage{iccv}
\usepackage{times}
\usepackage{epsfig}
\usepackage{graphicx}
\usepackage{amsmath}
\usepackage{amssymb}
\usepackage{caption}
\usepackage[accsupp]{axessibility}  

\usepackage[pagebackref=true,breaklinks=true,letterpaper=true,colorlinks,bookmarks=false]{hyperref}

\usepackage{xcolor,colortbl}
\definecolor{LightCyan}{rgb}{0.92,1,1}
\definecolor{LightGreen}{rgb}{0.92,1,0.92}
\definecolor{LightPurple}{rgb}{1,0.92,1}
\definecolor{LightOrange}{rgb}{0.99,0.84,0.69}

\newcolumntype{a}{>{\columncolor{LightCyan}}c}
\newcolumntype{b}{>{\columncolor{LightGreen}}c}
\newcolumntype{d}{>{\columncolor{LightPurple}}c}
\newcolumntype{f}{>{\columncolor{LightOrange}}c}

\usepackage{enumitem}
\usepackage{wrapfig}
\usepackage{amssymb}

\iccvfinalcopy 

\def\iccvPaperID{11046} 

\ificcvfinal\fi

\begin{document}

\title{Prototype-based Dataset Comparison}

\author{Nanne van Noord\\
University of Amsterdam\\
{\tt\small n.j.e.vannoord@uva.nl}
}

\twocolumn[{
\maketitle
\ificcvfinal\thispagestyle{empty}\fi

{\setlength{\tabcolsep}{0.4em} 
\begin{center}
    \captionsetup{type=figure}
{\sffamily
    \begin{tabular}{dd!{\color{white}\vrule width 4pt}bb!{\color{white}\vrule width 4pt}aaaa}
    \multicolumn{4}{c}{Dataset-specific Prototypes} & \multicolumn{4}{c}{Shared Prototype} \\
    \includegraphics[width=0.11\linewidth]{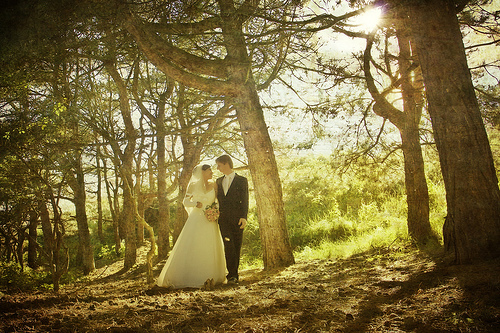} &
    \includegraphics[width=0.11\linewidth]{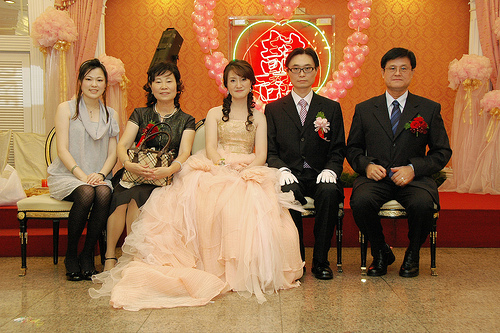} &
    \includegraphics[width=0.11\linewidth]{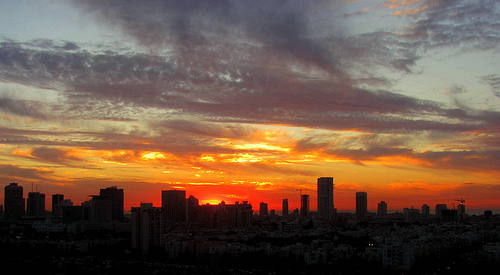} &
    \includegraphics[width=0.10\linewidth]{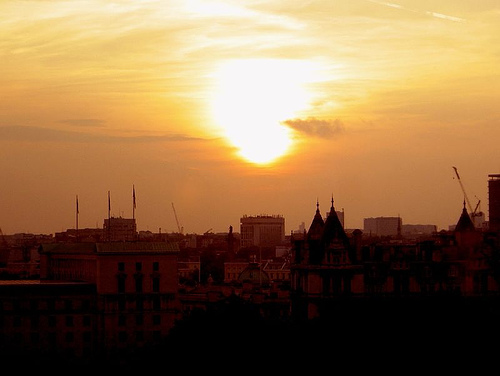} &
    \includegraphics[width=0.11\linewidth]{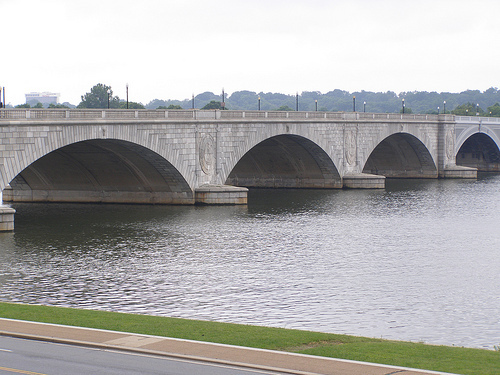} &
    \includegraphics[width=0.11\linewidth]{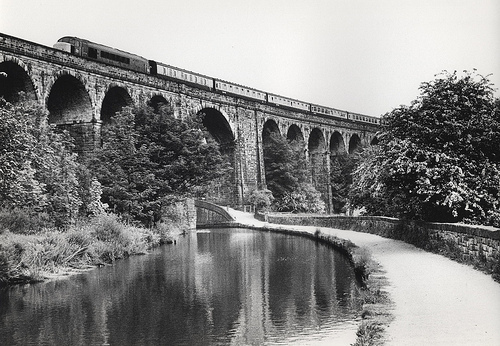} &
    \includegraphics[width=0.08\linewidth]{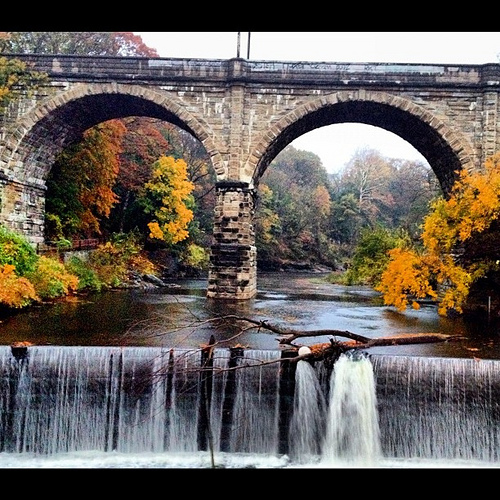} &
    \includegraphics[width=0.10\linewidth]{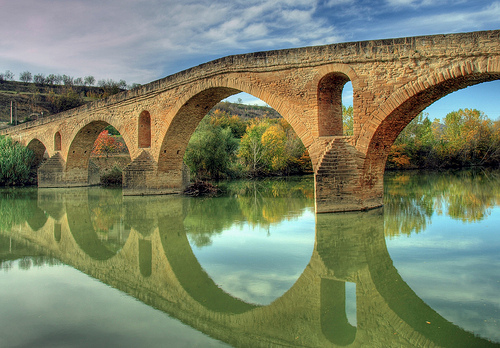} \\
    \includegraphics[width=0.11\linewidth]{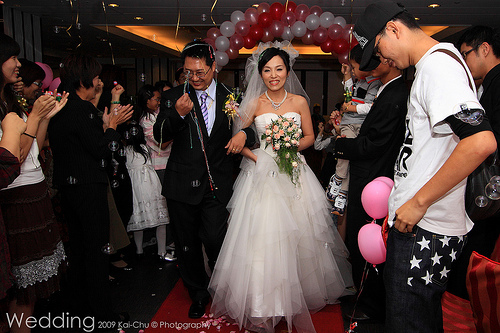} &
    \includegraphics[width=0.11\linewidth]{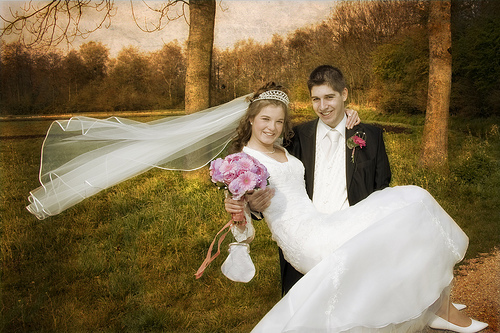} &
    \includegraphics[width=0.11\linewidth]{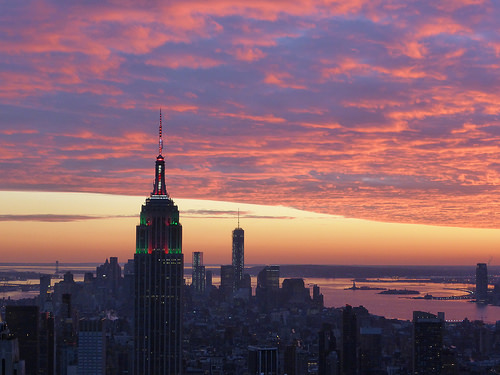} &
    \includegraphics[width=0.11\linewidth]{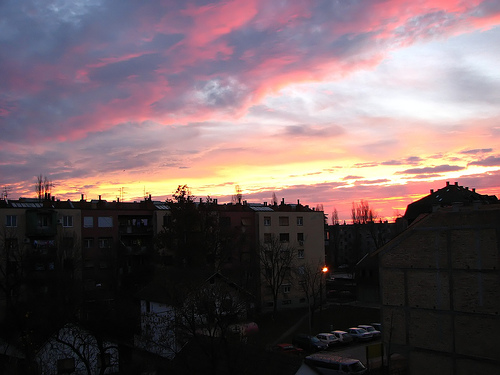} &
    \includegraphics[width=0.11\linewidth]{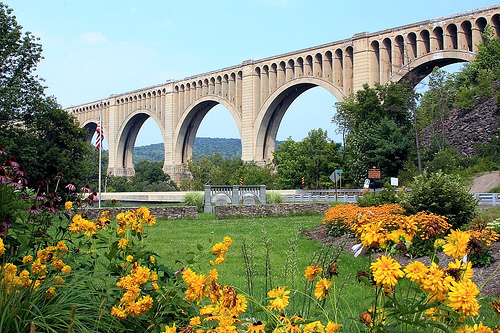} &
    \includegraphics[width=0.11\linewidth]{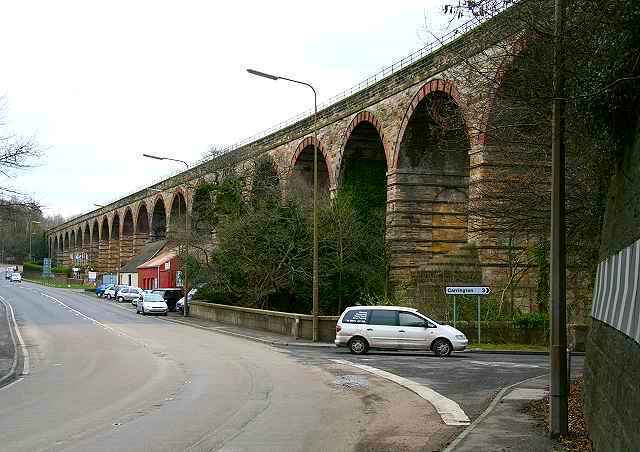} &
    \includegraphics[width=0.10\linewidth]{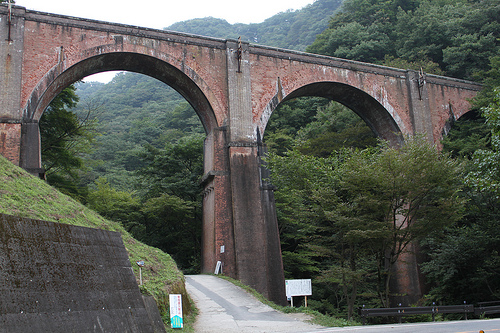} &
    \includegraphics[width=0.10\linewidth]{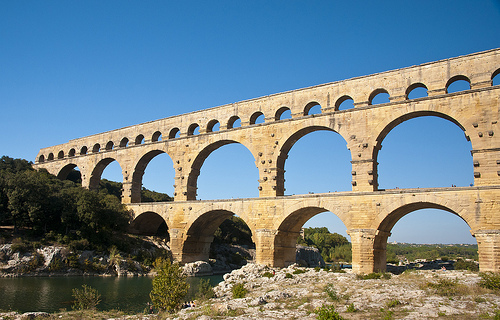} \\
    \multicolumn{2}{c}{ImageNet} & \multicolumn{2}{c}{PASS} & \multicolumn{2}{c}{ImageNet} & \multicolumn{2}{c}{PASS} \\
    \end{tabular}}
    \vspace{2pt}
    \captionof{figure}{\textbf{Prototypes discovered with ProtoSim}. By training on both datasets ProtoSim is able to discover prototypes that are dataset specific or shared. From left to right: ImageNet specific ``Wedding'', PASS specific ``sunset skyline'', and shared ``viaduct'' prototypes.}
    \label{fig:teaser}
    \end{center}
}
}]

\begin{abstract}
Dataset summarisation is a fruitful approach to dataset inspection. However, when applied to a single dataset the discovery of visual concepts is restricted to those most prominent. We argue that a comparative approach can expand upon this paradigm to enable richer forms of dataset inspection that go beyond the most prominent concepts.

To enable dataset comparison we present a module that learns concept-level prototypes across datasets. We leverage self-supervised learning to discover these prototypes without supervision, and we demonstrate the benefits of our approach in two case-studies. Our findings show that dataset comparison extends dataset inspection and we hope to encourage more works in this direction. Code and usage instructions available at \url{https://github.com/Nanne/ProtoSim}
\end{abstract}

\section{Introduction}
\label{sec:intro}

Image datasets are crucial for Computer Vision and due to the algorithms' need for more data they are ever growing in size. At the same time datasets are a major source of bias leading to negative social impact \cite{paulladaDataItsDis2021,prabhuLargeImageDatasets2021}. Unfortunately, it is challenging to determine what is contained in a dataset as their large size combined with the visual nature of the data makes manual inspection infeasible. To support users and developers of large-scale datasets in ensuring that the datasets match their usage and design goals it is necessary to develop better tools for dataset inspection.

A promising direction for generic dataset inspection is found in highly effective approaches that have been proposed for summarisation \cite{sinhaSummarizationPersonalPhotologs2011,doerschWhatMakesParis2012,goelVisualHashtagsVisualSummarization2017,yuHierarchicallyAttentiveRNNAlbum2017,liMidLevelDeepPattern2015}. A major benefit of these approaches is that they enable explorative dataset inspection without needing supervised pretraining. 
However, a limitation of these approaches is that they use frequency as a proxy for importance, and on a single dataset therefore only discover those visual concepts which are most prominent. As such, we argue that a comparative approach, which enables discovery of a wider and more diverse range of concepts, is necessary to effectively perform dataset inspection. 
For instance, the PASS dataset \cite{asanoPASSImageNetReplacement2021} is designed as an ImageNet \cite{russakovskyImagenetLargeScale2015} alternative whilst containing no people, as such it provides us with a testable hypothesis that is comparative in nature. Namely, when comparing these two datasets there should be a disjoint set of visual people-centric concepts that are only found in ImageNet. In a case study we will verify this hypothesis, and demonstrate that dataset comparison can lead to new insights. 

Moreover, an additional limitation of existing summarisation approaches is that they decouple the summarisation process from representation learning, and treat these as two distinct steps by performing the summarisation on a pre-defined feature basis, such as GIST descriptors in \cite{sinhaSummarizationPersonalPhotologs2011} or LDA clusters in \cite{rematasDatasetFingerprintsExploring2015}. 
As recent work on incorporating prototypes into a network's reasoning process has been shown to aid interpretability \cite{liDeepLearningCaseBased2018,chenThisLooksThat2019,rymarczykProtoPSharePrototypeSharing2021,ghorbani2019, yuksekgonul2023}, we propose that this may also be a promising direction for end-to-end dataset comparison. Therefore, we introduce a method for prototype-based dataset comparison, which discovers the visual concepts in a dataset in an end-to-end fashion.

Dataset prototypes are similar to cluster centroids in that they represent latent points in the feature space that are to be discovered. However, in clustering, similar to prior work on dataset summarisation, these centroids are discovered in a step that is decoupled from learning the feature representations. Whilst some recent works have explored forms of deep clustering \cite{yang2017, gao2019, li2021}, they still involve two separate optimisation goals. For example, by having one loss focused on the feature representation and another on clustering \cite{gao2019}. Instead, we propose a simple module \textit{ProtoSim} that can be integrated into a deep network without changing how it is optimised. To demonstrate this, we add ProtoSim to a Vision Transformer (ViT) \cite{dosovitskiyImageWorth16x162020} and show that it can effectively discover visual concepts across datasets in a self-supervised setting. 

Overall, we make the following contributions:
\begin{itemize}[topsep=2pt]
    \setlength\itemsep{0.1em}
    \item We introduce dataset comparison, a new approach for inspecting datasets.
    \item To enable dataset comparison we present ProtoSim, a module for integrated learning of dataset prototypes.
    \item With two case-studies we demonstrate how dataset comparison can be used to gain new insight into datasets.
\end{itemize}

\section{Related work}
\label{sec:rel}

Dataset comparison through prototypes is a largely unexplored area of research, but it is similar in spirit to dataset distilliation \cite{zhaoDatasetCondensationDistribution2021,cazenavetteDatasetDistillationMatching2022} which seeks to distill the knowledge from a large dataset into a small dataset. Most works in dataset distillation generate synthetic examples, as opposed to selecting a set of data instances, to construct the small dataset. These synthetic examples could be considered dataset prototypes, as they represent an aggregate of information that is central to the dataset. A key difference between such prototypes and the prototypes considered in this work, is that the distilled synthetic examples are instance-level prototypes (i.e., complete images that may represent multiple concepts), whereas in this work the aim is to discover concept-level prototypes, which may correspond to global concepts or part-level concepts. 

In the remainder of this section we will further explore  how dataset comparison relates to dataset summarisation, prototype learning, and deep clustering.

\subsection{Dataset Summarisation}

Manual browsing is a reliable manner to determine what is contained in a dataset, however, browsing even a subset of the data can be time-consuming, which has led to a line of work on dataset summarisation \cite{rematasDatasetFingerprintsExploring2015,doerschWhatMakesParis2012,goelVisualHashtagsVisualSummarization2017,yuHierarchicallyAttentiveRNNAlbum2017,liMidLevelDeepPattern2015}. The aim of dataset summarisation is to find underlying structure or patterns that are shared between data instances to give an overview of what is contained in a dataset. 

An important modeling choice for dataset summarisation is the basis on top of which the patterns are found. Work on mid-level pattern mining aims to find discriminative and representative patches and use these as visual primitive for further analysis \cite{singhUnsupervisedDiscoveryMidLevel2012}. Mid-level pattern mining aims to distinguish itself from ``visual word'' approaches \cite{sivicVideoGoogleText2003}, which were found to mainly capture low-level patterns such as edges and corners \cite{singhUnsupervisedDiscoveryMidLevel2012,doerschWhatMakesParis2012}. Two notable patch-based approaches are the dataset fingerprinting approach by \cite{rematasDatasetFingerprintsExploring2015} and the \textit{what makes Paris look like Paris?} approach by \cite{doerschWhatMakesParis2012}. Patch-based approaches aim to circumvent the limitations of visual words by starting with larger visual structure (i.e., patches at multiple scales). 
Due to the limitations of pixel-space representations patches cannot be used directly for analysis, instead initial patch-based approaches used feature extractors such as HOG \cite{singhUnsupervisedDiscoveryMidLevel2012,doerschWhatMakesParis2012} and Convolutional Neural Network (CNN) activations in later works \cite{liMidLevelDeepPattern2015}. 
Using patches has been shown to help steer the results towards more semantically meaningful visual elements, yet they are still restrictive in representational power and in terms of shape (i.e., patches must be rectangular). Moreover, although the starting point is a patch, the features extracted largely determine what the pattern represents and the features are not intrinsically explainable, necessitating post-hoc explanations. In contrast, we propose to add ProtoSim to a ViT architecture, which can learn to recognise visual elements of arbitrary shapes and sizes, whilst still focusing on semantically meaningful concepts. 

\subsection{Prototype Learning} 

Prototype-based approaches can be categorised into two areas, the first aims to learn prototypes which represent high-level semantic concepts (\ie, classes) to support zero or few-shot learning \cite{jetleyPrototypicalPriorsImproving2015,mettesHypersphericalPrototypeNetworks2019,snellPrototypicalNetworksFewshot2017}. The second area is concerned with finding recurring visual elements and are 
increasingly used for model explainability \cite{liDeepLearningCaseBased2018,chenThisLooksThat2019,kraftSPARROWSemanticallyCoherent2021,trinhInterpretableTrustworthyDeepfake2021,ghorbani2019, koh2020, yeh2020, yuksekgonul2023}. As opposed to post-hoc explainability methods, prototypes are integral to the network and provide explanations of the network's reasoning process \cite{chenThisLooksThat2019}. Our approach is part of this second area and we will focus the discussion on integral prototypes for recurring visual elements. 

Integral prototypes have been shown to be effective for supervised multi-class classification \cite{chenThisLooksThat2019, ghorbani2019, koh2020, kraftSPARROWSemanticallyCoherent2021}, deepfake detection \cite{trinhInterpretableTrustworthyDeepfake2021}, and semantic segmentation \cite{zhou2022}. In the literature we can recognise two branches of work on integral prototypes, the first builds on ProtoPnet \cite{liDeepLearningCaseBased2018,chenThisLooksThat2019} and the second on Concept Bottleneck Models (CBM) \cite{ghorbani2019,koh2020}. Common across these approaches is that they add a layer preceding the output layer that maps the extracted features to prototypes (or concepts) and produce the final output based on the affinity to these prototypes. The logic is thus that because the prototypes are recognisable entities, the prediction is explained by the affinity to the prototypes. The main difference between these branches is that in ProtoPnet-like approaches the prototypes are latent vectors in the embedding space that are learnt end-to-end, whereas CBM approaches use a pre-defined concept library. 

However, these prior approaches all focus on a supervised setting and learn per-class prototypes or concepts, which is severely limiting for dataset summarisation. Instead, we propose to leverage a self-supervised loss to learn prototypes that are not class specific and can represent \textit{any} visual concept that is present in the data. In particular, our prototypes may represent class-level concepts or segment-level concepts as in CBM \cite{ghorbani2019}, but we learn them without concept-level supervision.
Previous works learned prototypes for the features at each location in the convolutional output of a Convolutional Neural Network (CNN) backbone
\cite{chen2019,kraftSPARROWSemanticallyCoherent2021,trinhInterpretableTrustworthyDeepfake2021}. This restricts the spatial extent of the prototypes to size of the receptive field. Additionally, in CNN the output is typically spatially averaged to achieve a global image representation, however, as this is not done for CNN-based prototype approaches they can only learn local and not global prototypes. Instead, we use Vision Transformers (ViT) \cite{dosovitskiyImageWorth16x162020}, as they offer solutions for both aforementioned issues capturing local and global concepts of various spatial extents.

\subsection{Deep Clustering}

Clustering relates to learning prototypes or concepts in that it has been used to define the library of concept CBM \cite{ghorbani2019}, and to an extent the underlying principles match the goal of clustering. In particular, in deep clustering \cite{yang2017, gao2019, li2021} the centroids are learned iteratively, resulting in them being conceptually close to prototypes or CBM-concepts.
However, in prior works on deep clustering \cite{yang2017, gao2019, li2021} the clustering objective is separate from the feature representation, whereas in prototype-based approaches the final output is determined by projecting the input to the prototype subspace. 

Another key difference is that by making the prototypes integral to the reasoning process, as in CBM \cite{ghorbani2019} or ProtoPnet \cite{chenThisLooksThat2019}, it is ensured that the prototypes are of meaningful to the final output. Post-hoc clustering, or as part of a separate objective, may lead to overrepresentation of clusters which capture meaningless information, \eg, orthogonal to the objective that was used to obtain the feature representation.

\section{Dataset Comparison}
\label{sec:method} 

\begin{figure}[!ht]
    \centering
    \includegraphics[width=\linewidth]{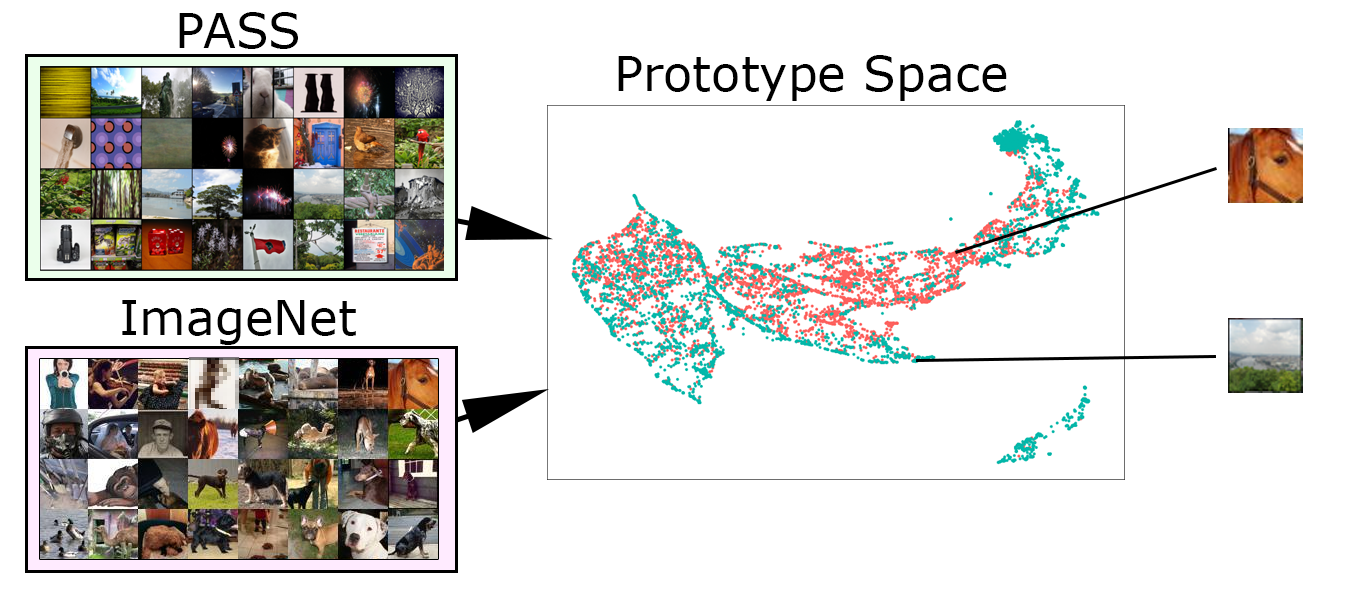}
    \caption{Overview of the dataset comparison workflow. Datasets are mapped to the prototype embedding space, from where we can inspect and compare individual prototypes.}
    \label{fig:dcworkflow}
\end{figure}

Given a collection of datasets $ X = \{\chi_i\}^{|X|}_{i=1} $ the aim of prototype-based dataset comparison is to discover a set of prototypes $ p \in \mathbb{R}^{K \times D} $ that represent visual concepts that may occur predominantly in only one dataset (\ie dataset-specific prototypes) or visual concepts which occur in two or more datasets (\ie shared prototypes). An illustration of the dataset comparison workflow is shown in Figure~\ref{fig:dcworkflow}.

We argue that a comparative approach to dataset summarisation, \ie, dataset comparison is vital for gaining insights into datasets. In particular, if $C_{\ast}$ is the set of \textit{all} possible visual concepts, and $C_i$ the visual concepts in ImageNet, then $C_i \subset C_{\ast}$. Given these sets, $C_i$ is a perfect description of what is contained in ImageNet,
and the difference $C_{\ast} - C_i$ perfectly describes what is not contained in ImageNet. However, as we do not have access to $C_{\ast}$ and we have no guaranteed manner of discovering $C_i$, we can at best learn a set of prototypes $\hat{C}_i$ that approximates $C_i$. 

As it is an approximation it cannot be concluded from $\hat{C}_i$ that a concept is not in $C_i$. For example, PASS is designed to
not contain humans (i.e., no humans in $C_p$), but based on $\hat{C}_p$ it cannot be stated with certainty that no humans are found in PASS, only that no human-centric concepts were discovered.
To overcome this limitation we propose to jointly learn $\hat{C}_i$ and $\hat{C}_p$. Thereby, if we know a concept is in $\hat{C}_i$ and not in $\hat{C}_p$, we can reasonably conclude that the concept is 
not in $C_p$ either. 

Through dataset comparison we are thus able to answer questions that could not be answered by only considering a single dataset, thereby creating new possibilities for dataset inspection for hypothesis verification or comparative exploration.

\subsection{Prototype Evaluation}

How to evaluate the prototypes learnt for dataset comparison (or summarisation) is an open problem, as the ground-truth for which visual concepts are contained in the dataset are unknown. Nonetheless, we can leverage the comparative approach by verifying whether the prototypes found match the design goals of the datasets. For instance, as PASS was designed not to contain humans we can explore whether the human-centric prototypes learnt from ImageNet indeed do not occur in PASS. This manner of evaluation forms the basis of our first case study.

Generally, we consider the prototypes $p$ to be of high-quality when they are \textit{distinct} and \textit{meaningful}. Distinct prototypes are able to independently represent a visual concept, this is in contrast to, for example, sparse coding \cite{zhang2013} where the aim is to find basis vectors that represent the input through linear combinations. The set of prototypes is meaningful when it can be used to discriminate images within and across the datasets from one another. Requiring that the prototypes are discriminative helps avoid trivial solutions. We can quantitatively measure how discriminative the prototypes are by evaluating them with downstream tasks. 

To ensure that the prototypes are distinct we design our method ProtoSim to perform hard assignment of the prototypes, and by optimising the prototypes with contrastive SSL we ensure that they are discriminative. This alignment between the design of the method (more details in Section~\ref{sec:protosim}) and the goal of dataset comparison helps us make certain the prototypes are of high-quality.

\subsection{Types of Prototypes}

Within the set of prototypes we can recognise different types of prototypes depending on how they occur across the datasets. For instance, a prototype $p_i$ may be considered \textit{dataset-specific} when over $95\%$ of its occurrences are only found in a single dataset. By definition, any prototype which is not dataset-specific would then be considered \textit{shared}. However, various degrees of `sharedness' may occur depending on the overlap between datasets. 

In theory it may be possible for the datasets in $X$ to be fully distinct, resulting in only dataset-specific prototypes. In practice, we find that prototypes may represent various basic visual properties (see Figure~\ref{fig:motionblur}), resulting in shared prototypes even when the datasets differ semantically.

\begin{figure}
    \centering
    \includegraphics[width=\linewidth]{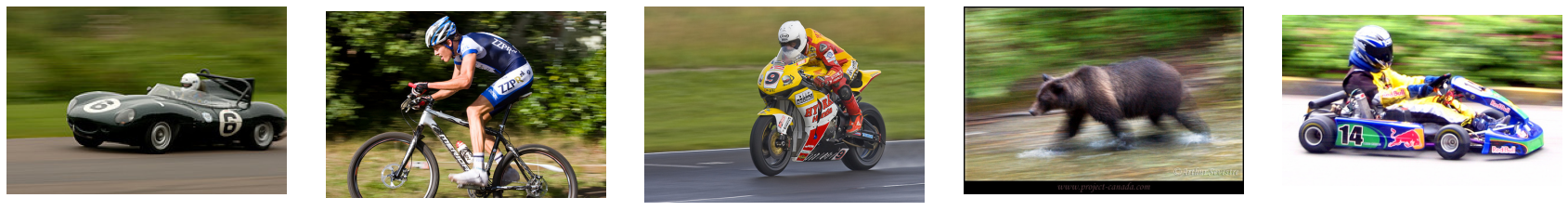}
    \includegraphics[width=\linewidth]{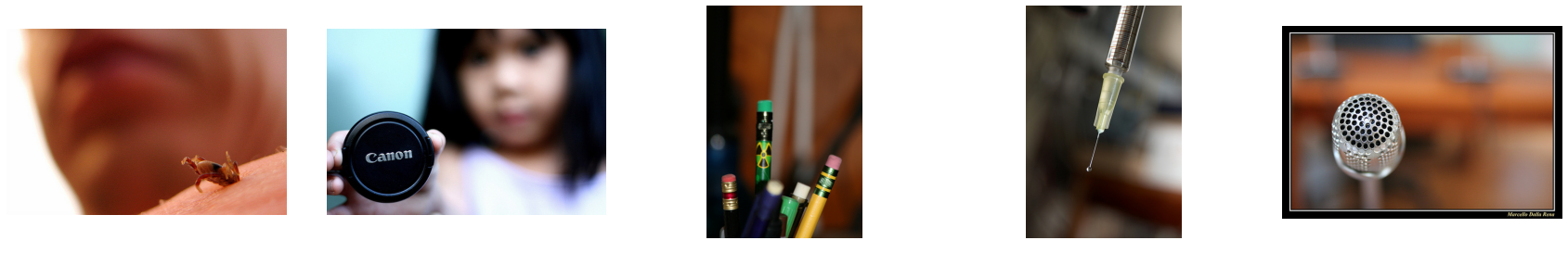}
    \caption{Examples of low-level visual properties discovered with ProtoSim: motion blur (top) and a shallow depth of field (bottom).} 
    \label{fig:motionblur}
\end{figure}

\section{Prototype-based Learning}
\label{sec:protosim}

We propose \textit{ProtoSim} a simple module that can inserted in ViT architectures to enable dataset comparison through integral prototype-based learning. ProtoSim differs from previous prototype layers in three ways, firstly, the prototypes learned with ProtoSim are distinct (as opposed to linear mixtures), secondly, ProtoSim is specifically designed for ViT architectures and learns prototypes from a set of tokens, rather than a single embedding vector. Lastly, the prototypes in ProtoSim are not class-specific, instead we learn a single pool of prototypes that may occur in any image.

\subsection{ProtoSim}

In order to learn which visual concepts are present in a dataset we aim to optimise a set of $K$ learnable prototypes $ p \in \mathbb{R}^{K \times D} $. By designing the prototypes to be part of a prototype layer (\ie ProtoSim) we can directly optimise them in and end-to-end fashion as part of ViT that we add the layer to. Specifically, given the token embeddings ${z \in \mathbb{R}^{{N+1} \times D}}$ produced by a ViT, ProtoSim maps these to the prototype embeddings $ \hat{z} \in \mathbb{R}^{N+1 \times D} $, where $D$ is the token vector size. During this mapping each token is replaced with the most similar prototype in $p$.

In \cite{liDeepLearningCaseBased2018,chenThisLooksThat2019} the similarity to each of the class-specifics prototypes is determined by calculating the squared $L^2$ distance and selecting the closest prototype with a max operation. 
Because our aim is to find prototypes that recur in the entire dataset, instead of finding class-specific prototypes, we propose a surprisingly simple formulation of this process by using a reversed version of dot-product attention \cite{luongEffectiveApproachesAttentionbased2015} to efficiently calculate attention across all the prototypes: 
\begin{align}
    a &= {softmax}(p z^\intercal) \label{eq:attention} \\
    \hat{z} & = a^\intercal p.
\end{align}
Here the attention mask $a$ represents the soft-assignment of prototypes to tokens, as such each token embedding $\hat{z}$ is a linear combination of prototypes. This is not desirable as our aim is to learn distinct prototypes which independently represent a visual concept, as this enables better summarisation and greater interpretability. To obtain a hard assignment we replace the softmax operation in Equation~\ref{eq:attention} with gumbel-softmax \cite{huijben2021}: 
\begin{align}
    a &= {gumbel\text{-}softmax}(p z^\intercal). \label{eq:gumbel} 
\end{align}
For simplicity we drop the temperature parameter, as it empirically does not meaningfully influence performance and can be kept fixed at $1$. The benefit of gumbel-softmax is that it maintains gradient-flow whilst enabling hard attention. Based on this modification $a$ now represents a hard assignment matrix which is used to replace each token with a prototype. As a visual concept can occur across multiple spatial locations there is no constraint on how often a prototype can be assigned. 

\subsection{Backbone}
Whilst ProtoSim functions as a generic network module, with the only requirement being an a two-dimensional input of $K \times D$, we specifically apply it to ViT in this work as these offer some benefits over CNN. 
In particular, in ViT all tokens can share information with all other tokens, as such there is no restriction on the spatial extent of the learned prototypes. Moreover, whilst most tokens in a ViT represent some spatial region of the image (\ie patch tokens), the class token takes on the role of a global image representation. This allows us to learn prototypes that are local (\ie only assigned to patch tokens) and global (\ie only assigned to the class token) without modifying the architecture.

ProtoSim is placed at the end of the backbone ViT in the network architecture. Formally,  the input $ x \in \mathbb{R}^{H \times W \times C}$ is divided into flattened patch tokens $ x_p \in \mathbb{R}^{N \times (P^2 \cdot C)}$, where $H$, $W$, and $C$ are the height, width, and channel dimensionality of the input and $P$ the patch size hyperparameter. In addition to $N$ patch tokens, a learnable \textit{class} token $x_{class}$ is prepended to input. No class labels are used during training, but we retain the ViT terminology of referring to this as a class token. After the embedding phase the tokens are passed through a series of transformer blocks resulting in the token embeddings ${z \in \mathbb{R}^{{N+1} \times D}}$, where $D$ is the embedded token vector size. $z$ is passed through the ProtoSim layer resulting in the prototype embeddings $\hat{z}$. 

\subsection{Training Objective}

The training objective determines the prototypes. For instance, a classification-based objective results in prototypes that are indicative of the classes (as in ProtoPnet \cite{chenThisLooksThat2019} or CBM \cite{koh2020}). Yet, such an objective may result in a poor coverage of concepts that are not part of the annotated classes. Because we strive for good coverage we focus on a Contrastive Self-Supervised Learning (Contrastive SSL) objective. In particular, because Constrastive SSL optimises for fine-grained discrimination between images (rather than classes) and has been shown to be perform well on a variety of downstream tasks \cite{chenSimpleFrameworkContrastive2020}.

From the recently proposed Contrastive SSL methods \cite{heMomentumContrastUnsupervised2020, grillBootstrapYourOwn2020, chenSimpleFrameworkContrastive2020, chenExploringSimpleSiamese2021, caronUnsupervisedLearningVisual2020, caronEmergingPropertiesSelfSupervised2021}, only DINO \cite{caronEmergingPropertiesSelfSupervised2021} has been specifically designed to work with ViT. As such, we minimise the DINO loss as the training objective:
\begin{equation}
	\min_{\theta_s} \sum_{x \in \{x^{g}_1, x^{g}_2\}} \quad \sum_{\substack{x' \in V\\x'\neq \, x}} \quad H(P_t(x), P_s(x')),
\label{eq:loss}
\end{equation}
where $H(a, b) = - a \log b$, $x^g_1$ and $x^g_2$ are two large crops from the input, $V$ a set of data-augmented smaller crops, and $P_t$ and $P_s$ are the teacher and student networks respectively, each with a ProtoSim layer added to the backbone. We refer to the excellent work by \cite{caronEmergingPropertiesSelfSupervised2021} for further details.

\section{Case Studies}
\label{sec:exp}

In this section we present two case studies, the first is a comparison between two datasets (\ie, two-way dataset comparison) aimed at determining how the PASS \cite{asanoPASSImageNetReplacement2021} dataset differs from ImageNet \cite{russakovskyImagenetLargeScale2015}. Because the design goal of PASS was to present an alternative for ImageNet that does not contain humans we test whether this is achieved successfully. The second case study focuses on a three-way dataset comparison scenario wherein we explore how three artwork datasets differ. These three datasets were designed for different tasks and contain images from different museum collections, as such based on these meta properties of the datasets it is challenging to determine how they actually relate. With these two case studies we put the two main applications of dataset comparison to the test, verifying a design goal of a dataset, and exploring unknown datasets. In addition, in Supplementary Section~A we demonstrate how ProtoSim can also be used on a single dataset. 

\subsection{Datasets}

For the first case study we focus on ImageNet and PASS, we briefly describe both:

\begin{itemize}
    \item \textbf{ImageNet} \cite{russakovskyImagenetLargeScale2015} is a widely used dataset available in two versions: ImageNet-21K and ImageNet-1K. In this work we focus on  ImageNet-1K consisting of approximately 1.2 million training images and 1000 classes. Despite how widely used it is, and how influential it has been, ImageNet-1K has also been criticised for its biased depictions of persons \cite{prabhuLargeImageDatasets2021}. 

    \item \textbf{PASS} \cite{asanoPASSImageNetReplacement2021} is a recently proposed alternative for ImageNet with the intention of not containing persons to avoid the issues in ImageNet. PASS contains approximately 1.4 million unlabeled images and can only be used for self-supervised training.

\end{itemize}

In the second case study we focus on three artwork datasets that have been used for a variety of different tasks:

\begin{itemize}
    \item \textbf{MET} \cite{ypsilantis2021} is a dataset designed for instance-level recognition, its training set contains over 400k images from more than 224k different objects held by the Metropolitan Museum of Art in New York.

    \item \textbf{Rijksmuseum} \cite{mensinkRijksmuseumChallengeMuseumcentered2014} presents a set of tasks which are museum-centric, focusing on predicting various types of metadata (\ie, artist, object-type, material usage, and creation year). It consists of 110k images obtained from collections held by the Rijksmuseum in Amsterdam.

    \item \textbf{SemArt} \cite{garciaHowReadPaintings2019} consists of images collected from the Web Gallery of Art a virtual museum with predominantly European fine-art. The SemArt dataset contains 21k images described with similar metadata as the Rijksmuseum dataset, but in addition also has textual descriptions in the form of artistic comments.
\end{itemize}

The available metadata for these three artwork datasets differs, but notably none of these datasets have been described with semantic classes as, for example, found in ImageNet. As such, we do not have prior information about what type of visual concepts are represented in the artwork datasets.

\subsection{Experimental setup}
The backbone used is the DeIT-S model \cite{touvronTrainingDataefficientImage2021} with a patch size of $16 \times 16$. The parameters of the backbone are fixed during prototype learning and training is done on a single NVIDIA RTX 3090 with a batch size of $128$ and learning rate of $5\mathrm{e}{-5}$ for $20$ epochs. For the first $15$ epochs we train with soft gumbel-softmax, after which we switch the student to the hard `straight-through' gumbel-softmax, to ensure more independent prototypes. In line with previous works (\eg, \cite{baoBEiTBERTPreTraining2022, maoDiscreteRepresentationsStrengthen2022}) we set $K$ to $8192$. 


For quantitative evaluation we follow the standard procedure of training a linear classifier on frozen features \cite{caronEmergingPropertiesSelfSupervised2021, heMomentumContrastUnsupervised2020}. We freeze the backbone and the ProtoSim layer and then train a linear classifier for $20$ epochs with a learning rate of $0.001$ and a batch size of $256$. 

\subsection{Two-way Dataset Comparison}

{\setlength{\tabcolsep}{0.4em} 
\begin{figure*}[!ht]
\centering
{\sffamily
    \begin{tabular}{dddd!{\color{white}\vrule width 2pt}bb}
    \multicolumn{6}{c}{ImageNet prototypes} \\
    \includegraphics[width=0.13\linewidth]{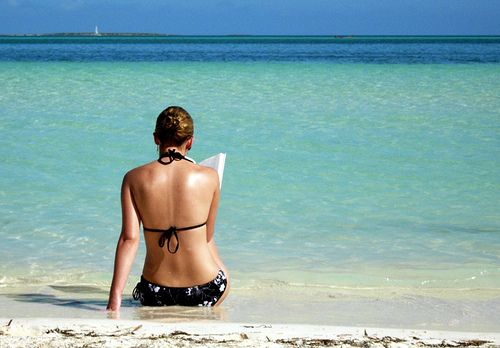} &
    \includegraphics[width=0.11\linewidth]{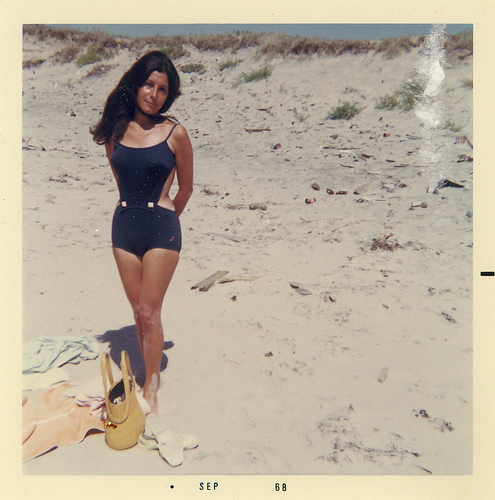} &
    \includegraphics[width=0.13\linewidth]{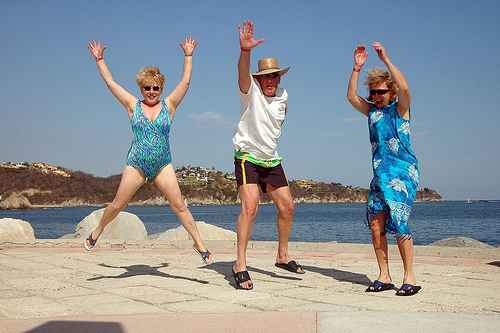} &
    \includegraphics[width=0.13\linewidth]{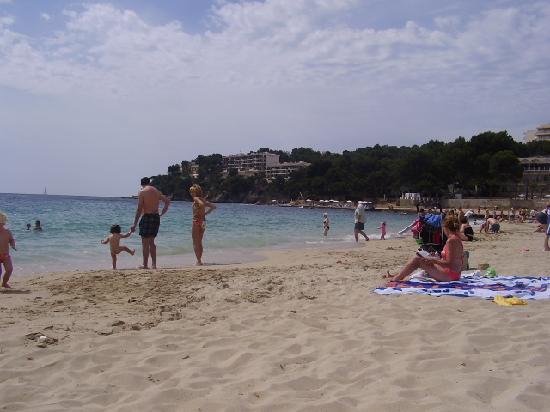} &
    \includegraphics[width=0.08\linewidth]{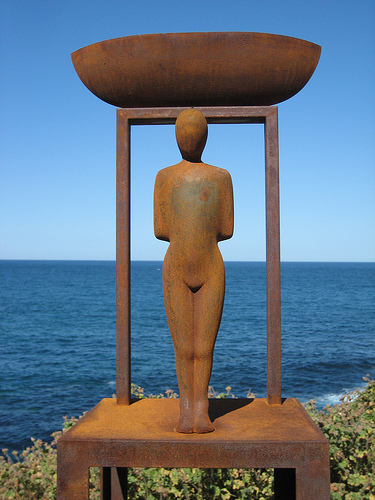} &
    \includegraphics[width=0.13\linewidth]{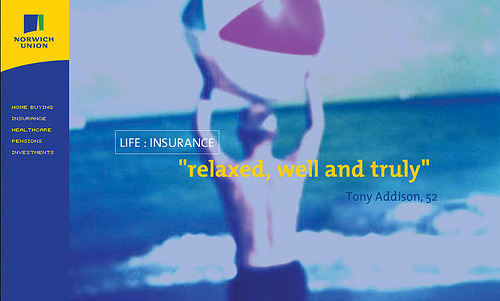} \\
    \includegraphics[width=0.13\linewidth]{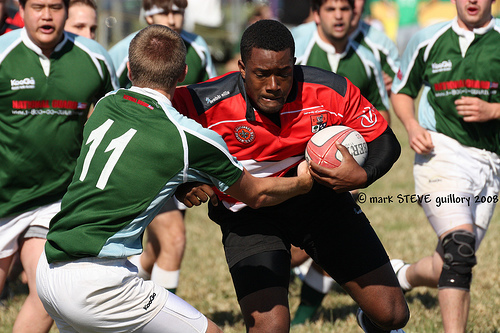} &
    \includegraphics[width=0.13\linewidth]{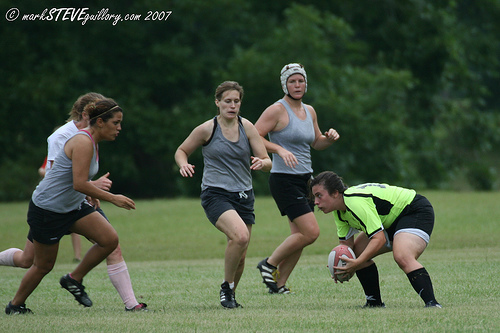} &
    \includegraphics[width=0.13\linewidth]{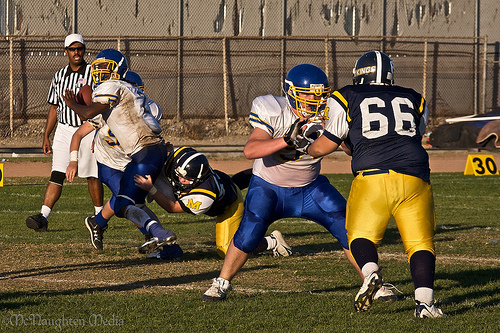} &
    \includegraphics[width=0.13\linewidth]{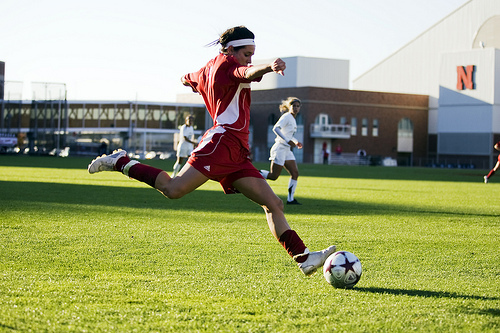} &
    \includegraphics[width=0.13\linewidth]{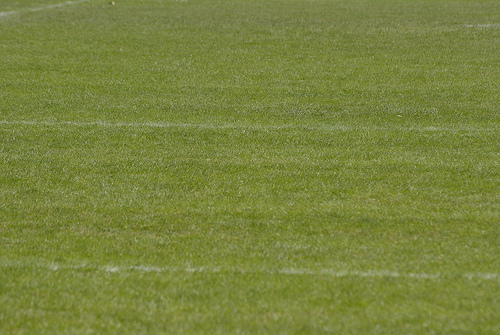} &
    \includegraphics[width=0.13\linewidth]{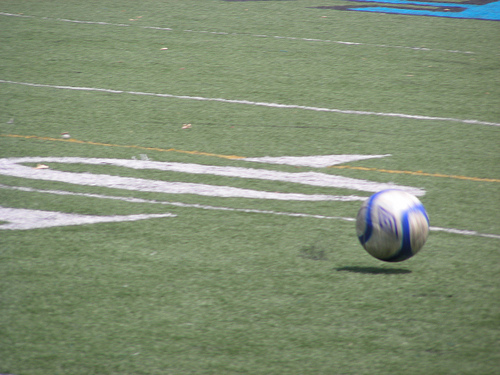} \\
    \includegraphics[width=0.06\linewidth]{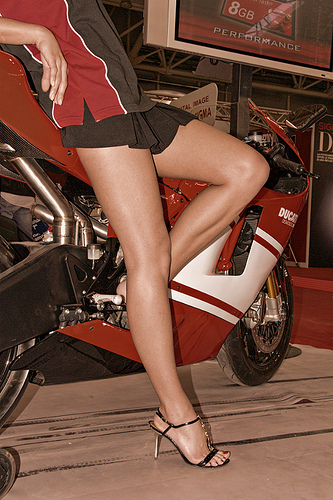} &
    \includegraphics[width=0.13\linewidth]{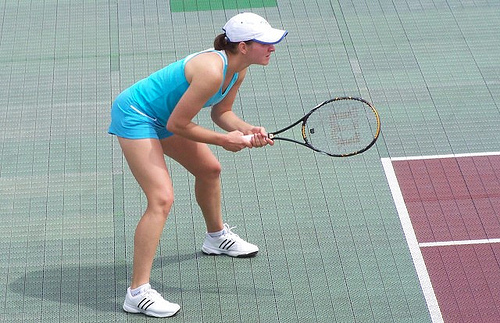} &
    \includegraphics[width=0.13\linewidth]{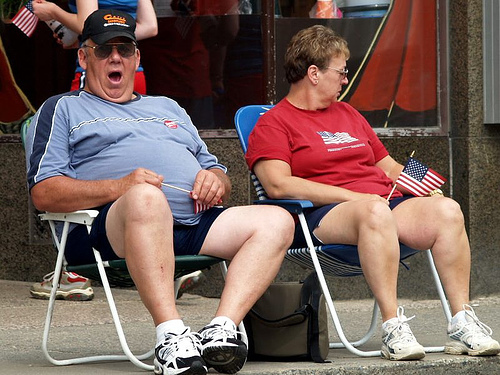} &
    \includegraphics[width=0.13\linewidth]{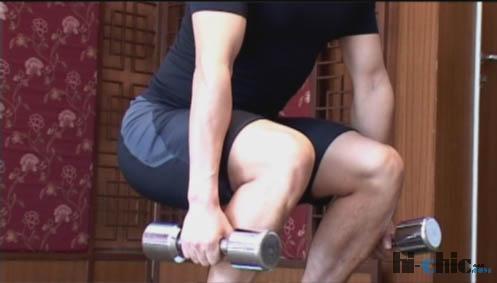} &
    \includegraphics[width=0.06\linewidth]{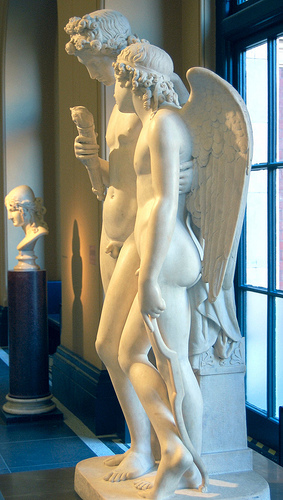} &
    \includegraphics[width=0.07\linewidth]{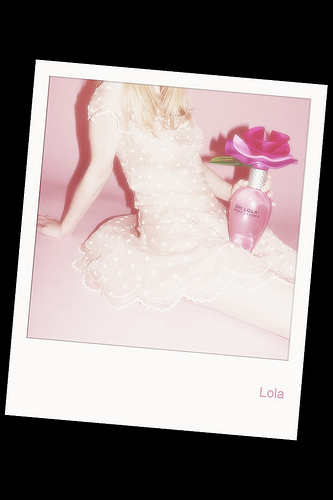} \\
    \end{tabular}
    \vspace{1em}
    \begin{tabular}{bbbb!{\color{white}\vrule width 4pt}dd}
    \multicolumn{6}{c}{PASS prototypes} \\
    \includegraphics[width=0.13\linewidth]{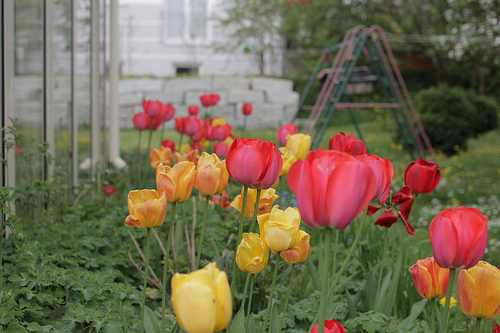} &
    \includegraphics[width=0.13\linewidth]{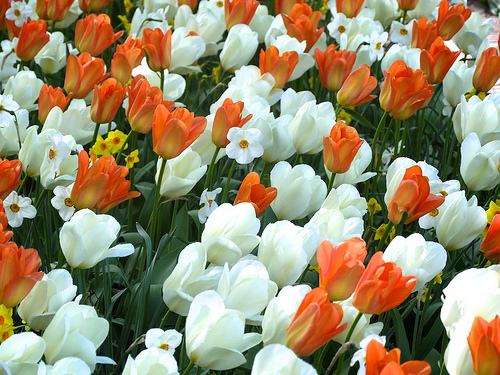} &
    \includegraphics[width=0.13\linewidth]{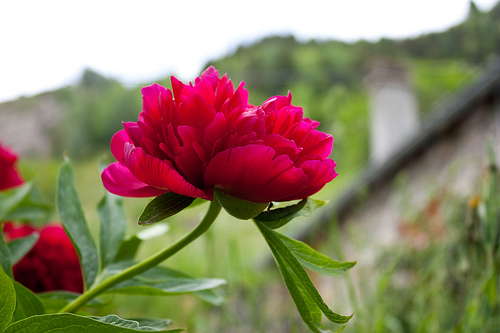} &
    \includegraphics[width=0.13\linewidth]{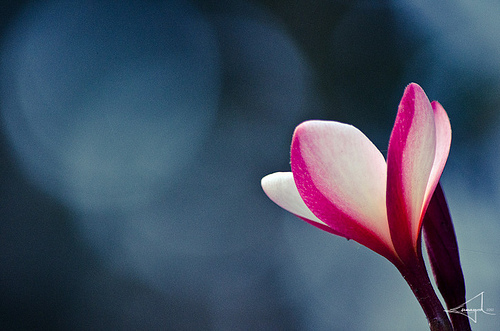} &
    \includegraphics[width=0.13\linewidth]{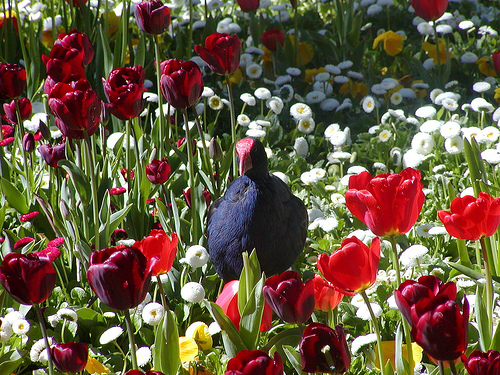} &
    \includegraphics[width=0.13\linewidth]{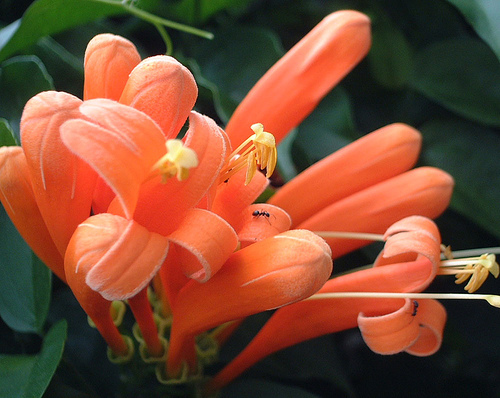} \\
    \includegraphics[width=0.11\linewidth]{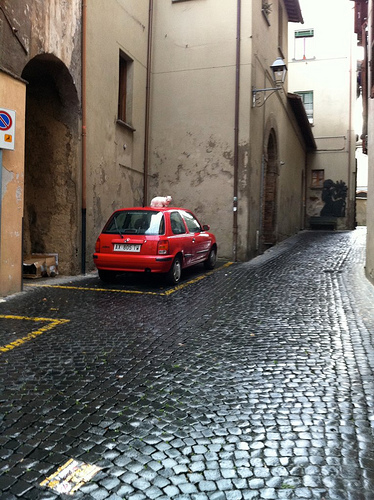} &
    \includegraphics[width=0.11\linewidth]{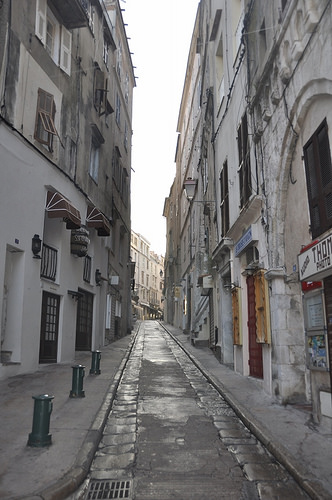} &
    \includegraphics[width=0.11\linewidth]{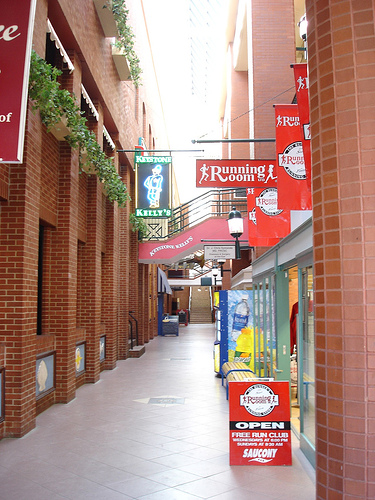} &
    \includegraphics[width=0.11\linewidth]{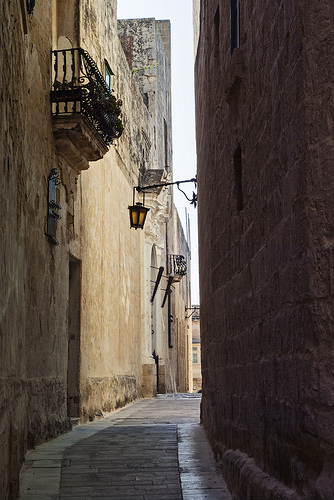} &
    \includegraphics[width=0.11\linewidth]{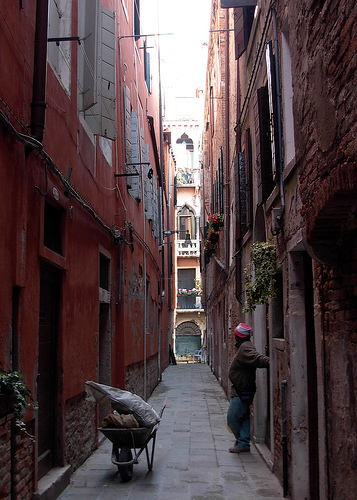} &
    \includegraphics[width=0.11\linewidth]{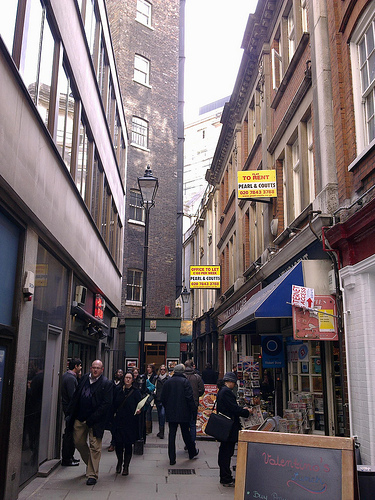} \\
    \includegraphics[width=0.13\linewidth]{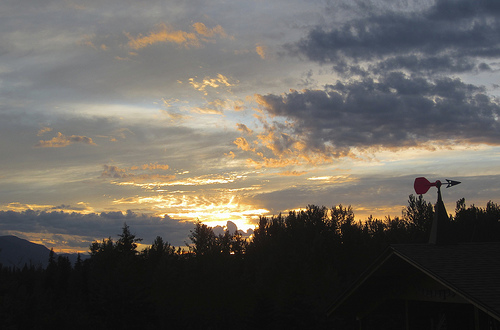} &
    \includegraphics[width=0.13\linewidth]{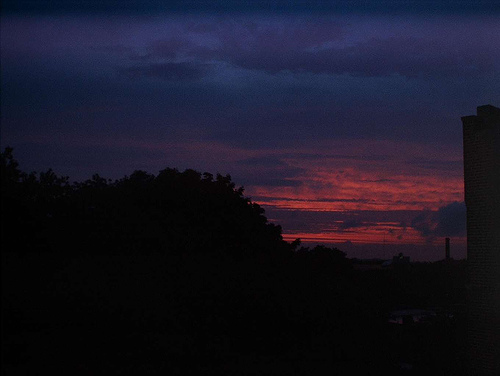} &
    \includegraphics[width=0.13\linewidth]{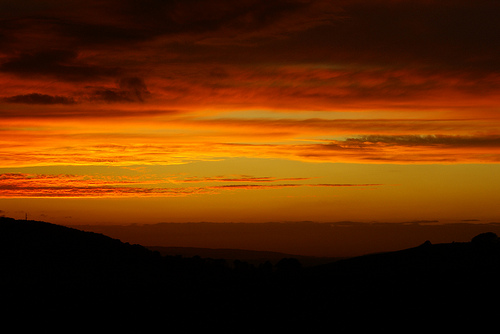} &
    \includegraphics[width=0.13\linewidth]{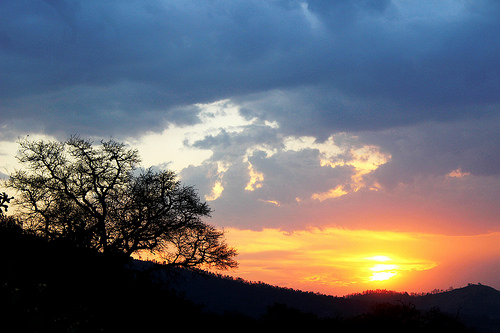} &
    \includegraphics[width=0.13\linewidth]{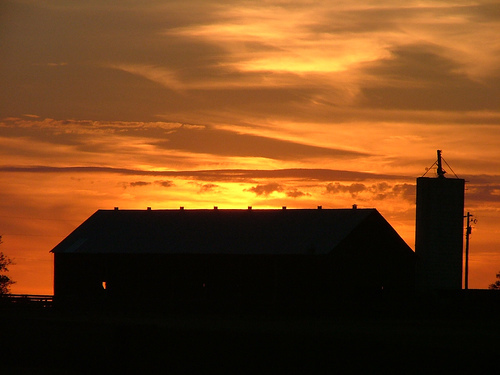} &
    \includegraphics[width=0.13\linewidth]{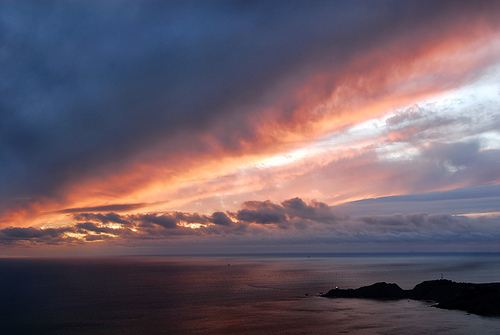} \\
    \end{tabular}
    \vspace{0.5em}
    \begin{tabular}{ddd!{\color{white}\vrule width 4pt}bbb}
    \multicolumn{6}{c}{Shared prototypes} \\
    \includegraphics[width=0.13\linewidth]{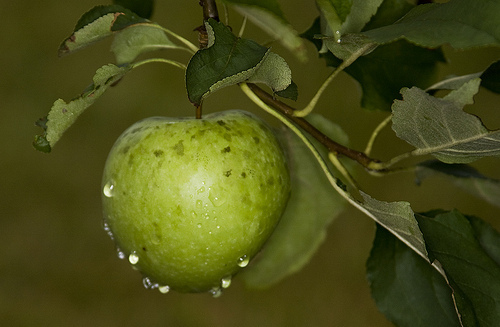} &
    \includegraphics[width=0.13\linewidth]{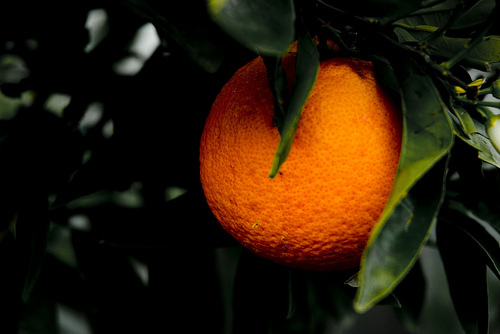} &
    \includegraphics[width=0.13\linewidth]{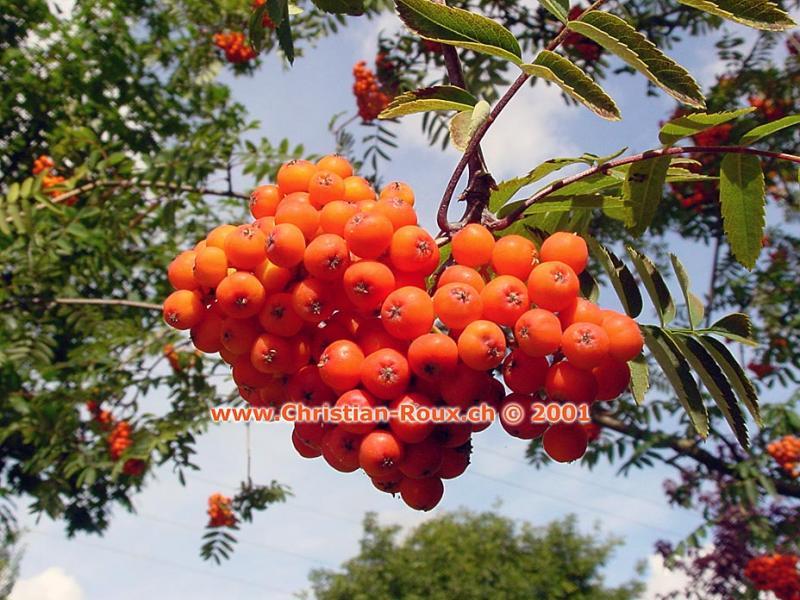} &
    \includegraphics[width=0.13\linewidth]{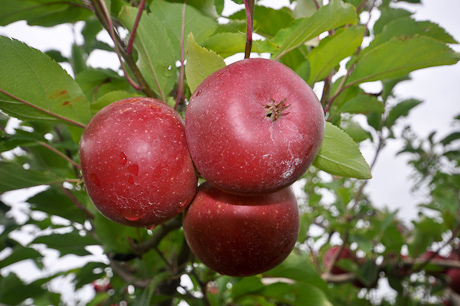} &
    \includegraphics[width=0.13\linewidth]{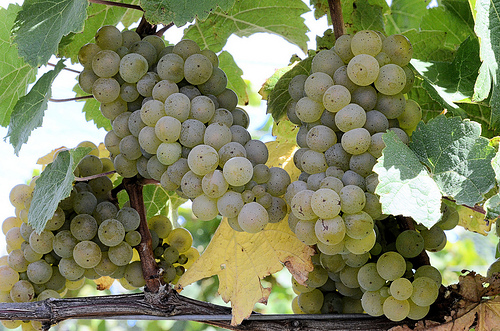} &
    \includegraphics[width=0.13\linewidth]{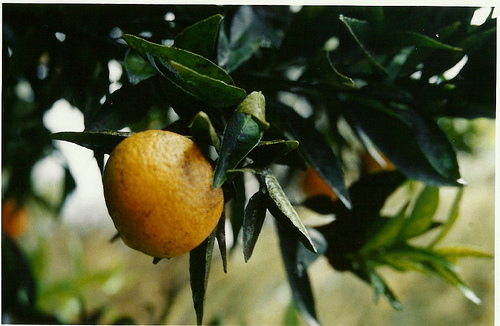} \\
    \includegraphics[width=0.12\linewidth]{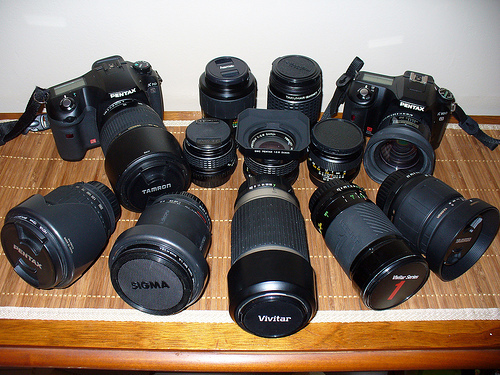} &
    \includegraphics[width=0.13\linewidth]{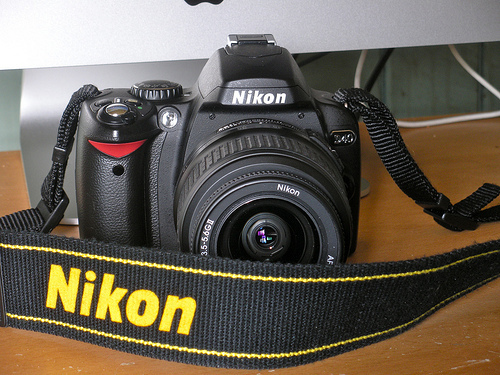} &
    \includegraphics[width=0.13\linewidth]{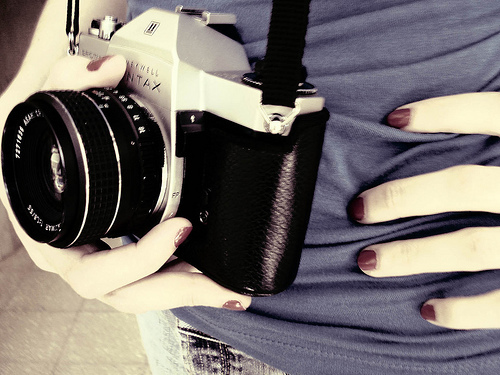} &
    \includegraphics[width=0.13\linewidth]{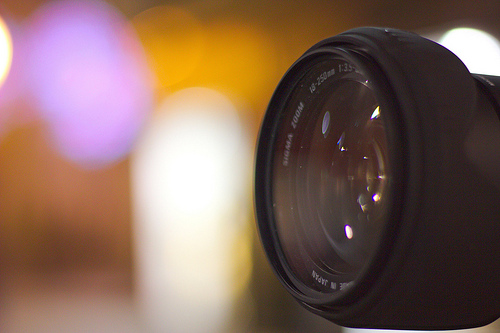} &
    \includegraphics[width=0.13\linewidth]{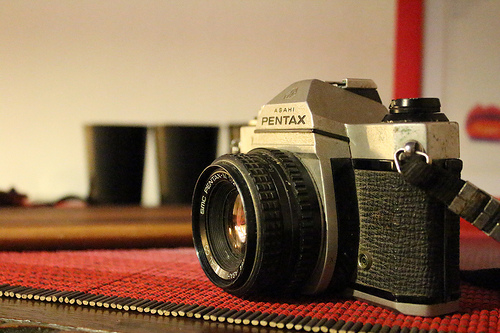} &
    \includegraphics[width=0.13\linewidth]{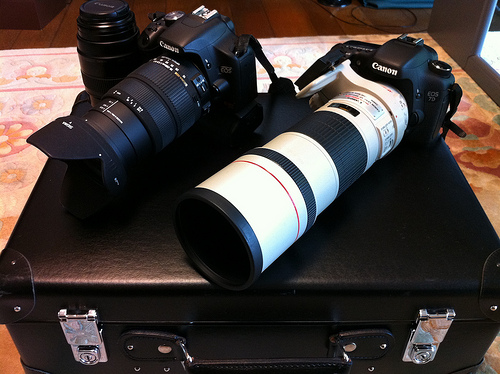} \\
    \includegraphics[width=0.13\linewidth]{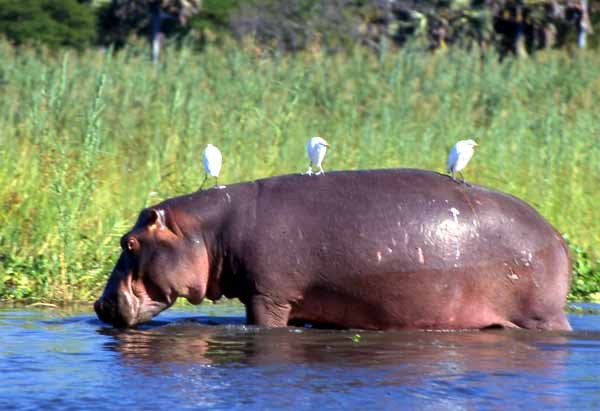} &
    \includegraphics[width=0.13\linewidth]{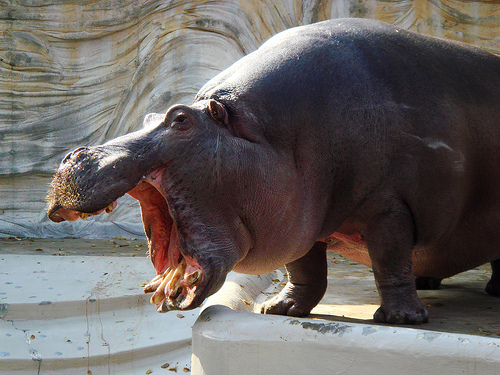} &
    \includegraphics[width=0.13\linewidth]{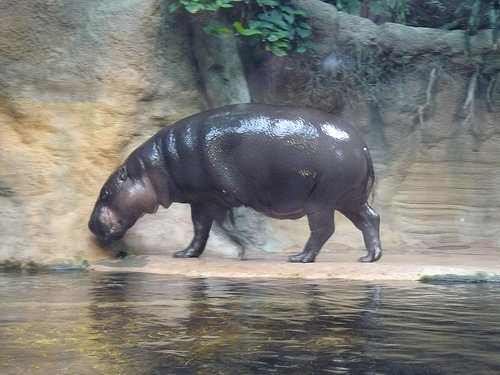} &
    \includegraphics[width=0.13\linewidth]{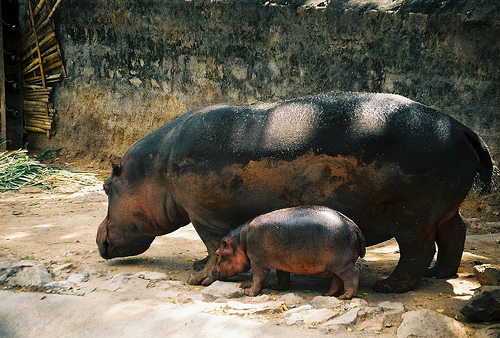} &
    \includegraphics[width=0.13\linewidth]{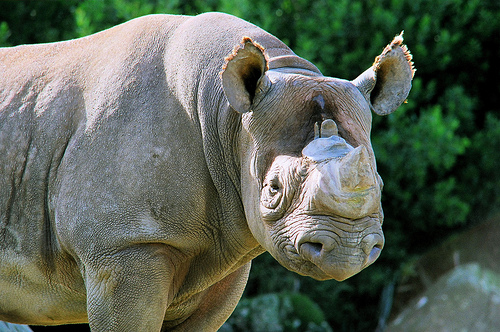} &
    \includegraphics[width=0.13\linewidth]{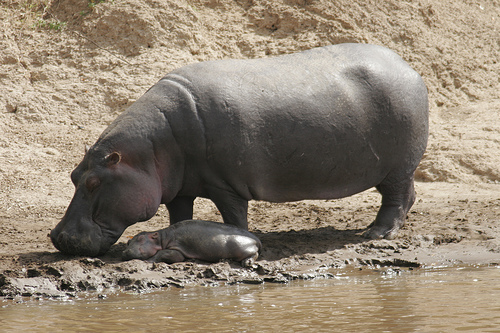} \\
    \end{tabular}}
    \caption{Example images for nine prototypes, each three rows showing predominantly ImageNet, predominantly PASS, and Shared prototypes respectively. The last two columns for rows one through six show some of the few examples found in the other dataset. The last three rows show prototypes common in both with examples of ImageNet in the left three columns, and examples in the right three.} 
    \label{fig:examples}
\end{figure*}
}

For the two-way dataset comparison case study we focus on the PASS and ImageNet datasets, to this end we learn prototypes on the union of these two datasets, which we refer to as the \textit{PassNet} dataset, and present the results in this section. All results in this section are obtained with ProtoSim trained on top of the DINO backbone pre-trained on ImageNet. A key finding of this comparison is that ProtoSim learns both dataset-specific prototypes (i.e., prototypes that predominantly activate for one dataset) and shared prototypes (i.e., prototypes that occur frequently in both datasets). In Figure~\ref{fig:examples} we show nine prototypes which are predominantly found in ImageNet, PASS, or are shared equally among both datasets. Additional examples of prototypes are shown in the supplementary material.

The prototypes that are predominantly found in ImageNet contain people (\ie, over 99\% of images for which they activate are found in ImageNet). However, despite its design goal we do find persons for these prototypes in PASS. The \textit{Persons at the beach} prototype (first row) contains a man holding a beach-ball photographed from the back, and the \textit{Uncovered legs/arms} prototype (third row) is found in (nude) statues and a faded Polaroid of a person. The only two PASS images for the \textit{Sports-ball} prototype (second row) just depict the field and a ball, suggesting that the prototype captures more than just persons.

The prototypes that are predominantly found in PASS depict concepts or objects that are not found among the $1000$ ImageNet categories. Nonetheless, these predominantly PASS prototypes can be found in some ImageNet images where the annotated object is a (small) part of the image. From left to right for row four to six we find the following ImageNet categories: European Gallinule, Ant, Wheelbarrow, Turnstile, Barn, and Headland. 
We would describe the prototypes in these rows as capturing \textit{Flowers}, a \textit{View down narrow alley}, and \textit{Sunset with silhouetted foreground} respectively. In general, the PASS prototypes that predominantly concern vista or landscape images, which on occasion contain objects from ImageNet categories. 

\begin{figure}[!ht]
\centering
    \begin{tabular}{cccc}
    \includegraphics[width=0.13\linewidth]{figures/examples/main-img/2714/n03770439_22301.JPEG} &
    \includegraphics[width=0.20\linewidth]{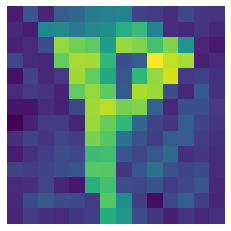} &
    \includegraphics[width=0.20\linewidth]{figures/examples/main-img/2714/n04039381_22632.JPEG} &
    \includegraphics[width=0.20\linewidth]{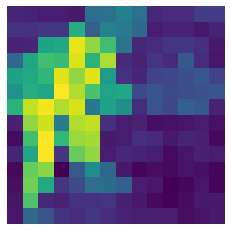} \\
    \includegraphics[width=0.20\linewidth]{figures/examples/main-img/2714/n03376595_10011.JPEG} &
    \includegraphics[width=0.20\linewidth]{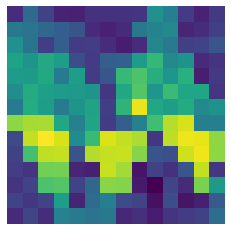} &
    \includegraphics[width=0.20\linewidth]{figures/examples/main-img/2714/n03255030_6099.JPEG} &
    \includegraphics[width=0.20\linewidth]{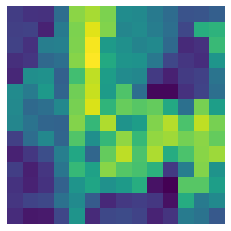} \\
    \end{tabular}
    
    \caption{Attention maps for the for the four ImageNet images for the \textit{Uncovered legs/arms} prototype in Figure~\ref{fig:examples}.}
    \label{fig:heatmaps}
\end{figure}

For the prototype in the second row of Figure~\ref{fig:examples} we measure how it is spatially distributed and visualise this for the four ImageNet images in Figure~\ref{fig:heatmaps}. Visualising the attention maps of this prototype shows that the main activations are indeed for the leg and arm areas of the persons depicted. Other parts of the persons receive much less attention than these uncovered areas. Although this is a predominantly ImageNet prototype, and is therefore commonly found in instances of the `miniskirt' and `bikini' categories, it is also found in PASS, including in images of mannequins, artworks, and anime of primarily feminine characters.

Overall, we can conclude that the two-way comparison is highly successful, not only are we able to verify that depictions of humans in PASS are practically non-existent, but by edge-cases we are able to demonstrate the strength of the prototypes. Moreover, based on the discovered prototypes we also gain insights into how these datasets differ, with PASS containing many more vistas and landscapes, whereas ImageNet is strongly object-centric. As such we can conclude that while there is an overlap, PASS differs in more respects from ImageNet than just the presence of humans - which may cast doubt on its status as an ImageNet replacement. 

\begin{table}[!ht]
    \caption{Performance for linear classifier on ImageNet.} 
    \label{tab:quanres}
    \begin{tabular}{cccc}
        \hline
         Train set & Backbone & $K$ & Accuracy \\
         \hline
         ImageNet &  DINO \cite{caronEmergingPropertiesSelfSupervised2021} & $8192$ & $64.8$ \\ 
         PASS & PASS-DINO \cite{asanoPASSImageNetReplacement2021} & $8192$ &  $20.4$ \\ 
         PassNet (I+P) &  DINO \cite{caronEmergingPropertiesSelfSupervised2021} & $8192$ & $53.2$ \\ 
         PassNet (P+I) & PASS-DINO \cite{asanoPASSImageNetReplacement2021} & $8192$ &  $15.9$ \\ 
         \hline
         ImageNet &  DINO \cite{caronEmergingPropertiesSelfSupervised2021} & $1024$ & $54.7$ \\ 
         ImageNet &  DINO \cite{caronEmergingPropertiesSelfSupervised2021} & $4096$ & $63.7$ \\ 
         \hline
    \end{tabular}
\end{table}

\textbf{Quantitative analysis.}
We perform two quantitative analyses. Firstly, we explore the loss value of each model during training and compare it to the diversity of the prototypes, as measured by the average cosine distance between the prototype weights. 
Secondly, we train a linear classifier on top of each of the backbones to perform ImageNet classification. For these evaluations we compare between two different backbones, one pre-trained on ImageNet \cite{caronEmergingPropertiesSelfSupervised2021} and one pre-trained on PASS \cite{asanoPASSImageNetReplacement2021}. Results reported on PassNet are described as either (I+P) for the ImageNet pre-trained backbone, or P+I for the PASS pre-trained backbone.  

Table~\ref{tab:quanres} shows the performance on the ImageNet validation set for a linear classifier trained on top of fixed backbones. We observe that pre-training on ImageNet (unsurprisingly) leads to the best downstream performance on ImageNet, and that training on PASS leads to lower performance. Similarly, for the two backbones we find that the ImageNet pre-trained backbone performs much better. Nonetheless, we see degraded performance for training on PassNet versus ImageNet. The last two rows of Table~\ref{tab:quanres} show that increasing the $K$ value benefits performance, due to computational complexity and diminishing returns no values above $8192$ were tested. In conclusion, our quantative evaluation confirms our modelling choices of using the ImageNet pre-trained backbone with $8192$ prototypes, and we will use this configuration for the second case study as well.

\begin{figure}[!ht]
    \centering
    \includegraphics[height=7\baselineskip]{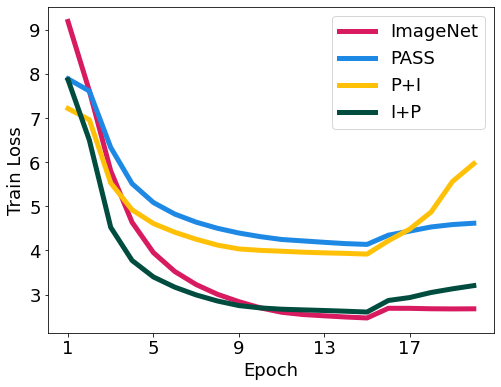}
    \includegraphics[height=7\baselineskip]{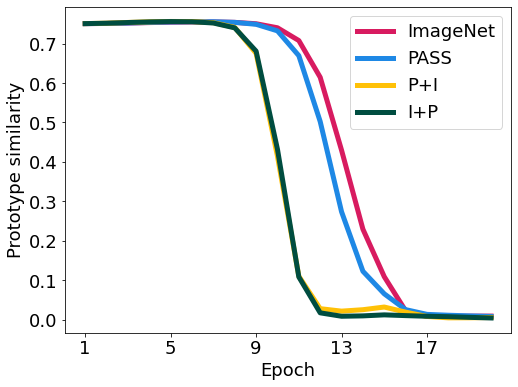}
    \caption{Training loss (left) and average cosine similarity (right) during prototype training. PassNet results P+I are obtained PASS-DINO backbone and I+P with the DINO backbone.}
    \label{fig:loss}
\end{figure}

The results in Figure~\ref{fig:loss} show the training loss and the average cosine similarity between prototypes as training progresses. In both plots we see two groupings of two lines that follow the same pattern. For all models we can see that switching to the hard gumbel-softmax leads to a bump in the loss, but except for P+I this bump levels off quickly. 

\subsection{Three-way Dataset Comparison}

In the second case study we perform a three-way dataset comparison between the MET, Rijksmuseum, and SemArt datasets. The aim of this comparison is highly exploratory, unlike ImageNet, none of these datasets have pre-defined classes, as such it is not obvious what visual concepts are represented. In Figure~\ref{fig:artexamples} we show nine prototypes discovered with ProtoSim, the first three pairs of prototypes are dataset-specific and are predominantly found in the MET, Rijksmuseum, and SemArt respectively. These prototypes were randomly selected from the most dataset-specific prototypes and give a good indication of what is contained in these datasets. The MET dataset contains many 3D objects and often multiple photographs are taken from different angles, as such the first prototype represents \textit{Top-down views of vases}. The second MET prototype depicts \textit{Egpytian wall carvings}, which were photographed under raking light to emphasise the height differences across the surface.

The Rijksmuseum dataset contains many prints or drawings on paper, whilst this is something that can be learnt from the material type overview in \cite{mensinkRijksmuseumChallengeMuseumcentered2014}, the two prototypes depicted show two different groups of visual concepts. The first Rijksmuseum prototype depicts \textit{natural history drawings}, whereas the second is found in \textit{side view portraits}. SemArt on the other hand consists of mostly fine-art paintings, with the first prototype capturing a stylistic visual concept of \textit{Renaissance paintings}, the second prototype is a more concrete semantic concept, namely \textit{dogs}.

Just by exploring the dataset-specific datasets we already get a much better understanding of what is contained in these datasets and what makes them unique. In addition, when we then analyse the shared prototypes we can see that these differences largely persists across different visual concepts. The first shared prototype is of various \textit{animals}, notably we see that the prototype is not sensitive to material differences, returning both drawings and paintings. The same observation is made for the second shared prototype of \textit{baskets}, which is matched to drawings, paintings, and photographs of actual baskets. This ability to generalise across modalities is also observed in the third shared prototype of \textit{brass instruments}, which matches actual instruments and realistic paintings containing similar instruments.

Based on the three-way dataset comparison we find that there are degrees of overlap between the datasets, generally we would see that the MET and Rijksmuseum datasets are most similar in types of objects, where MET has a greater focus on physical objects and Rijksmuseum on prints and drawings. Because these two datasets have a relatively small proportion of paintings the similarity with the SemArt dataset is less, nonetheless, we can observe that in terms of visual concepts these datasets represent similar things. In Appendix Section~C we show additional prototypes discovered for these artwork datasets.

{\setlength{\tabcolsep}{0.4em} 
\begin{figure}[!ht]
\centering
{\sffamily
    \begin{tabular}{dddd}
    \multicolumn{4}{c}{MET prototypes} \\
    \includegraphics[width=0.21\linewidth]{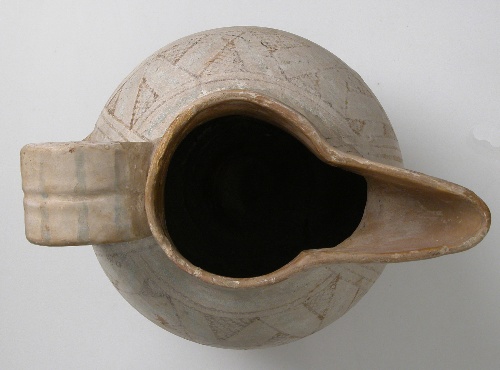} &
    \includegraphics[width=0.21\linewidth]{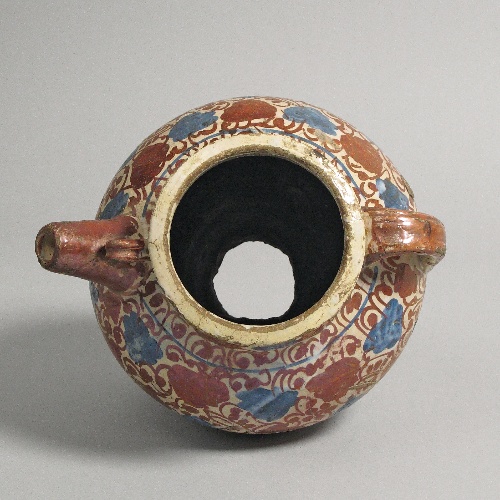} &
    \includegraphics[width=0.21\linewidth]{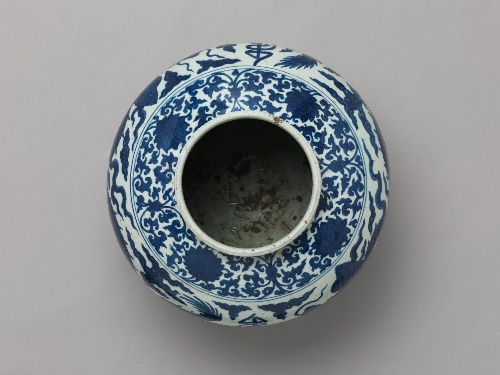} &
    \includegraphics[width=0.21\linewidth]{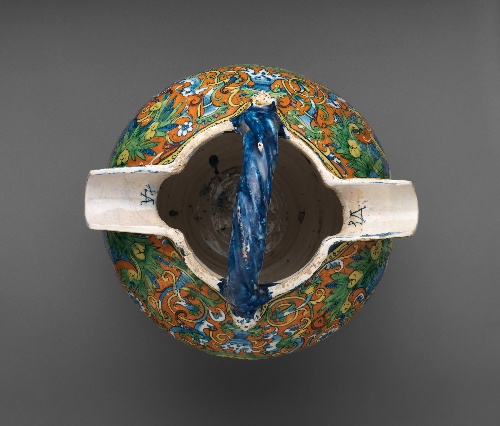} \\
    \includegraphics[width=0.21\linewidth]{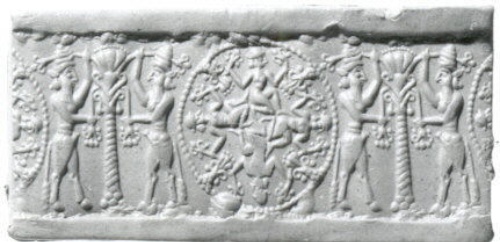} &
    \includegraphics[width=0.21\linewidth]{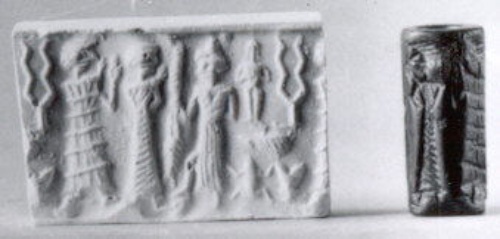} &
    \includegraphics[width=0.11\linewidth]{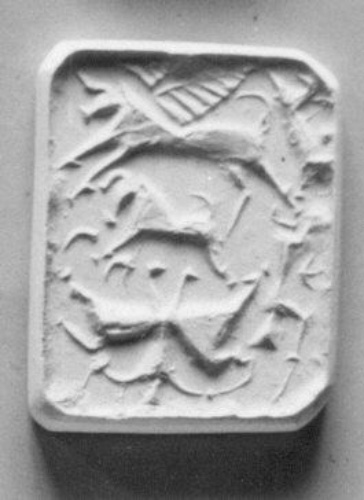} &
    \includegraphics[width=0.21\linewidth]{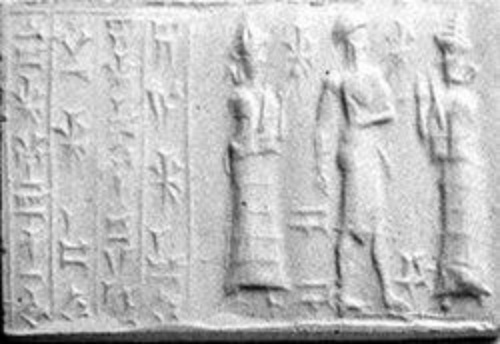} \\
    \end{tabular}
    \vspace{1em}
    \begin{tabular}{bbbb}
    \multicolumn{4}{c}{Rijksmuseum prototypes} \\
    \includegraphics[width=0.21\linewidth]{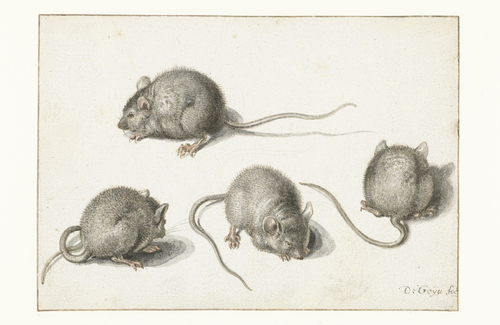} &
    \includegraphics[width=0.21\linewidth]{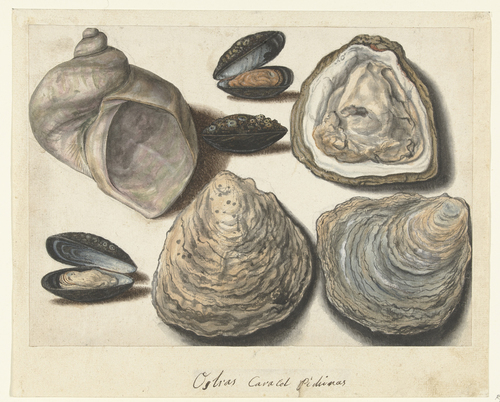} &
    \includegraphics[width=0.21\linewidth]{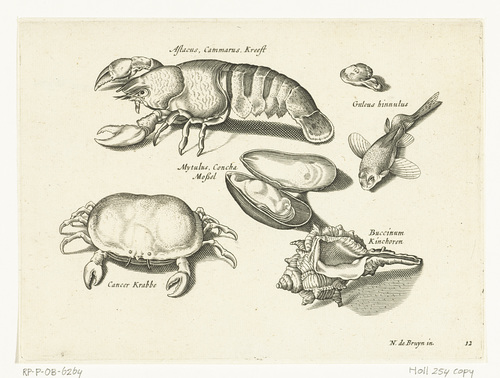} &
    \includegraphics[width=0.21\linewidth]{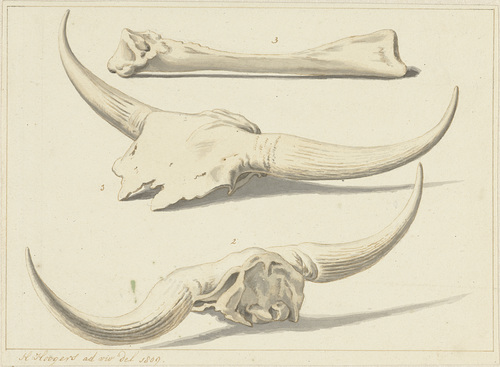} \\
    \includegraphics[width=0.21\linewidth]{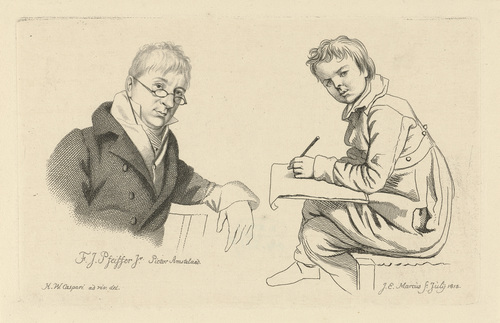} &
    \includegraphics[width=0.15\linewidth]{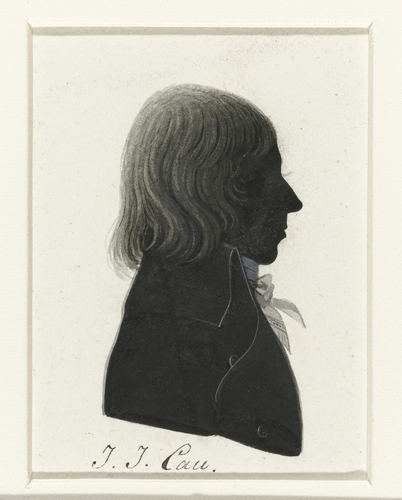} &
    \includegraphics[width=0.15\linewidth]{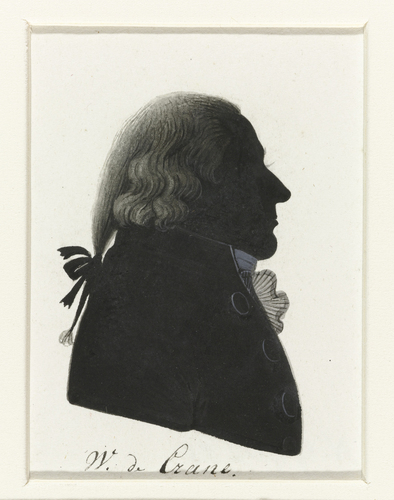} &
    \includegraphics[width=0.12\linewidth]{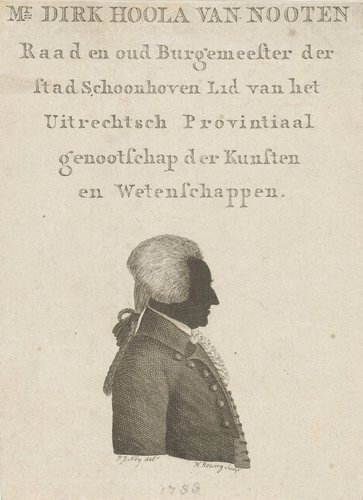} \\
    \end{tabular}
    \vspace{1em}
    \begin{tabular}{aaaa}
    \multicolumn{4}{c}{SemArt prototypes} \\
    \includegraphics[width=0.18\linewidth]{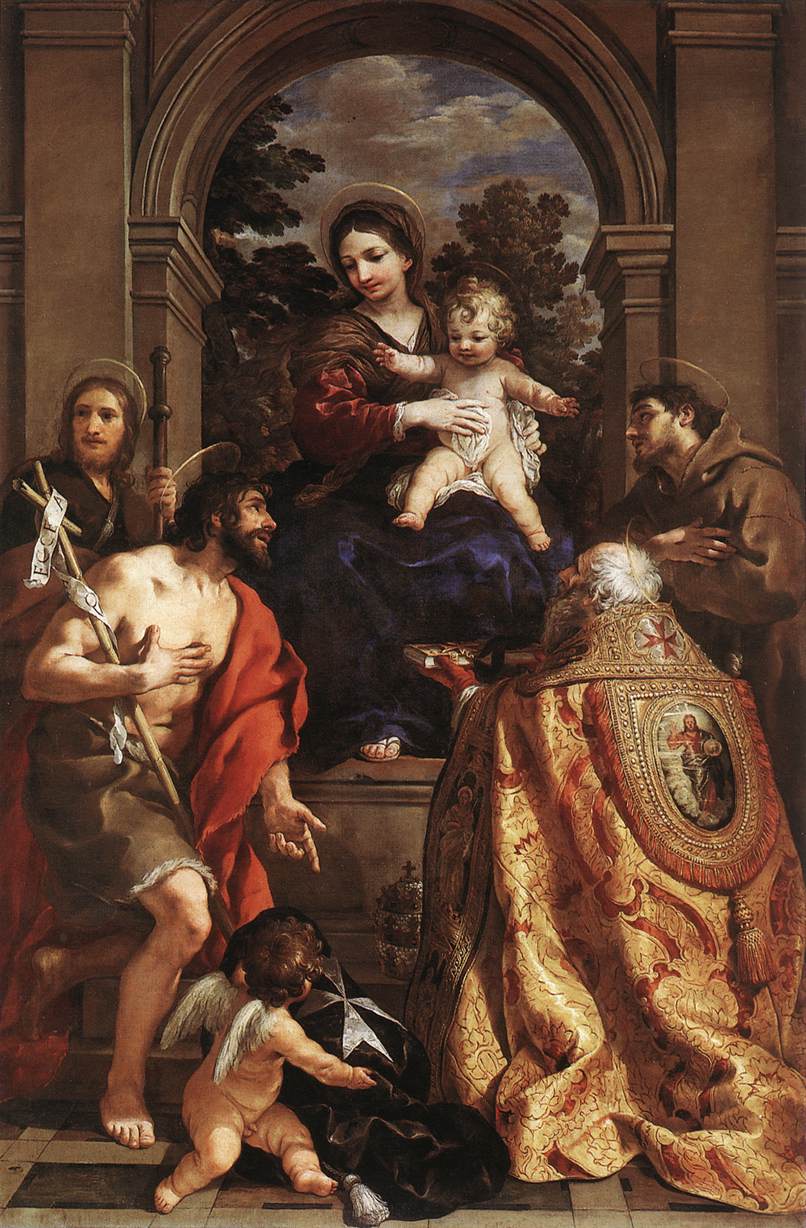} &
    \includegraphics[width=0.21\linewidth]{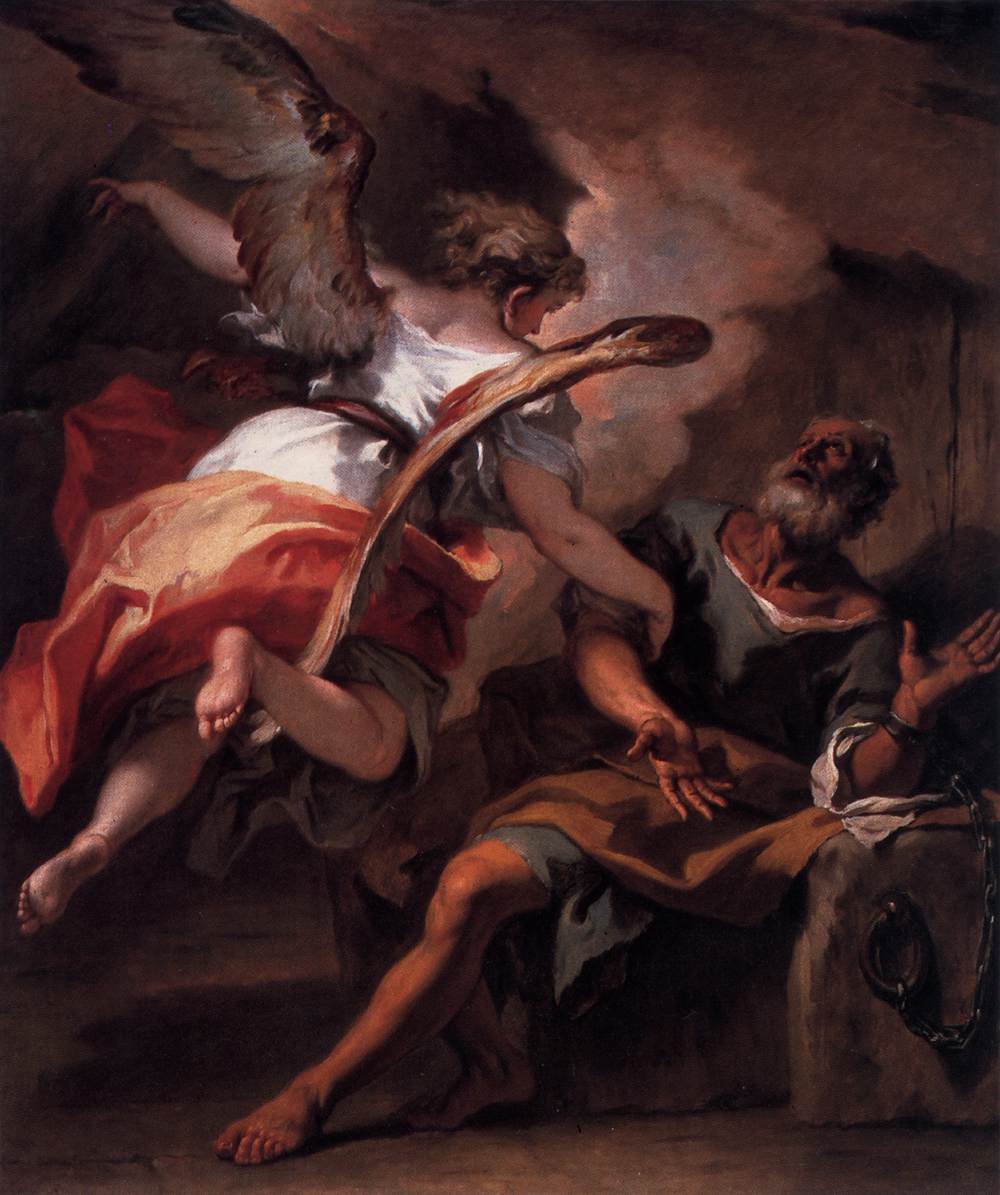} &
    \includegraphics[width=0.21\linewidth]{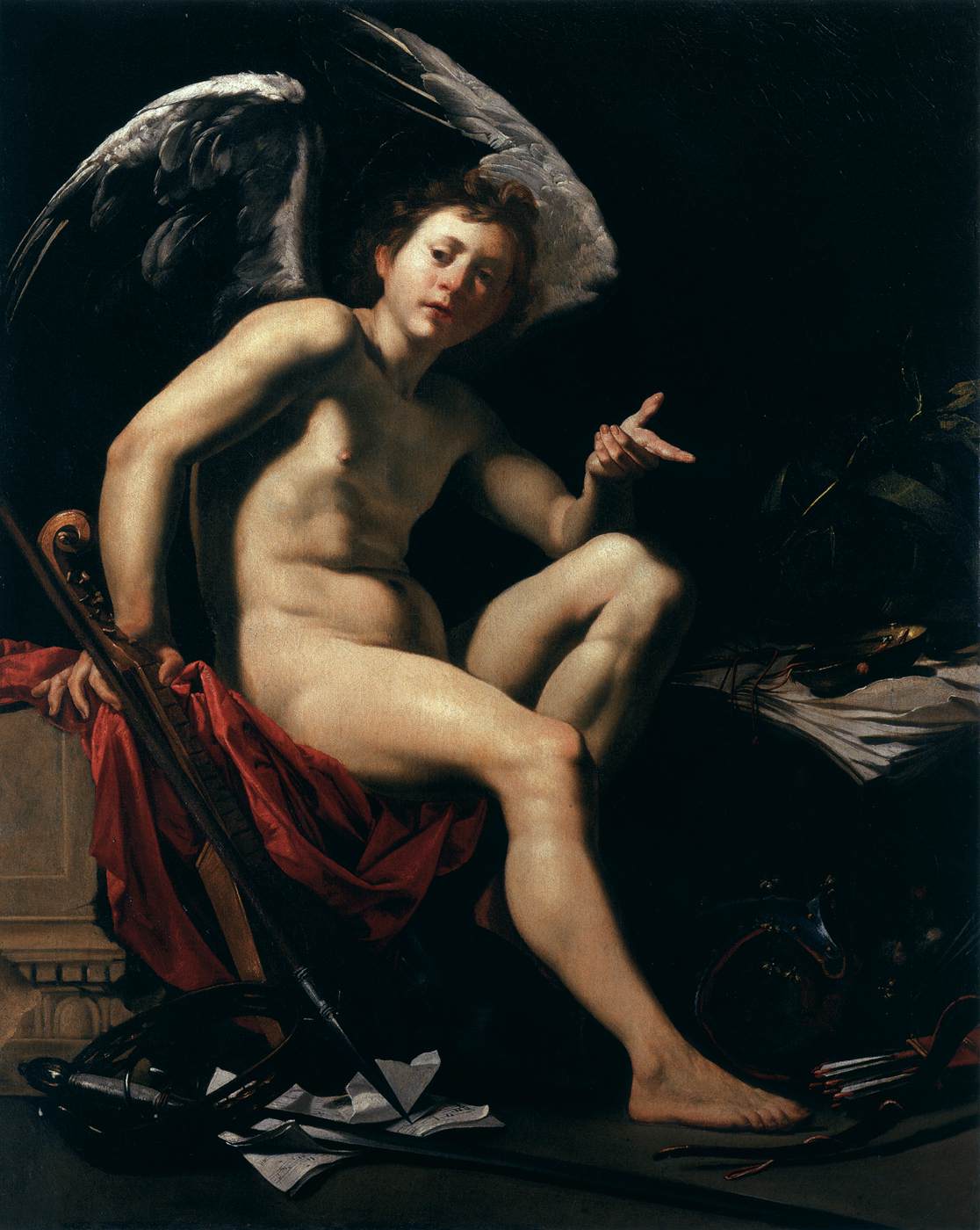} &
    \includegraphics[width=0.21\linewidth]{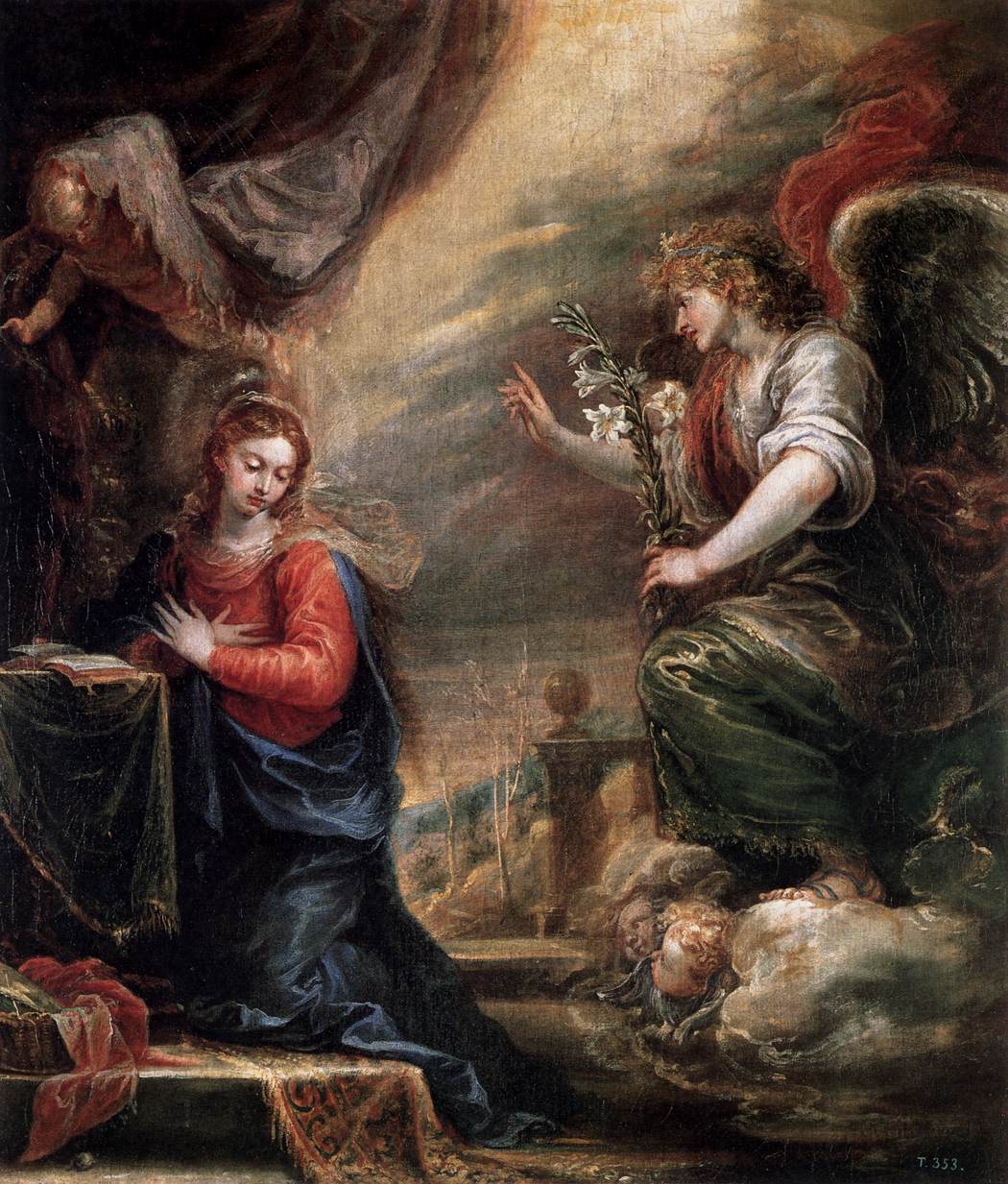} \\
    \includegraphics[width=0.21\linewidth]{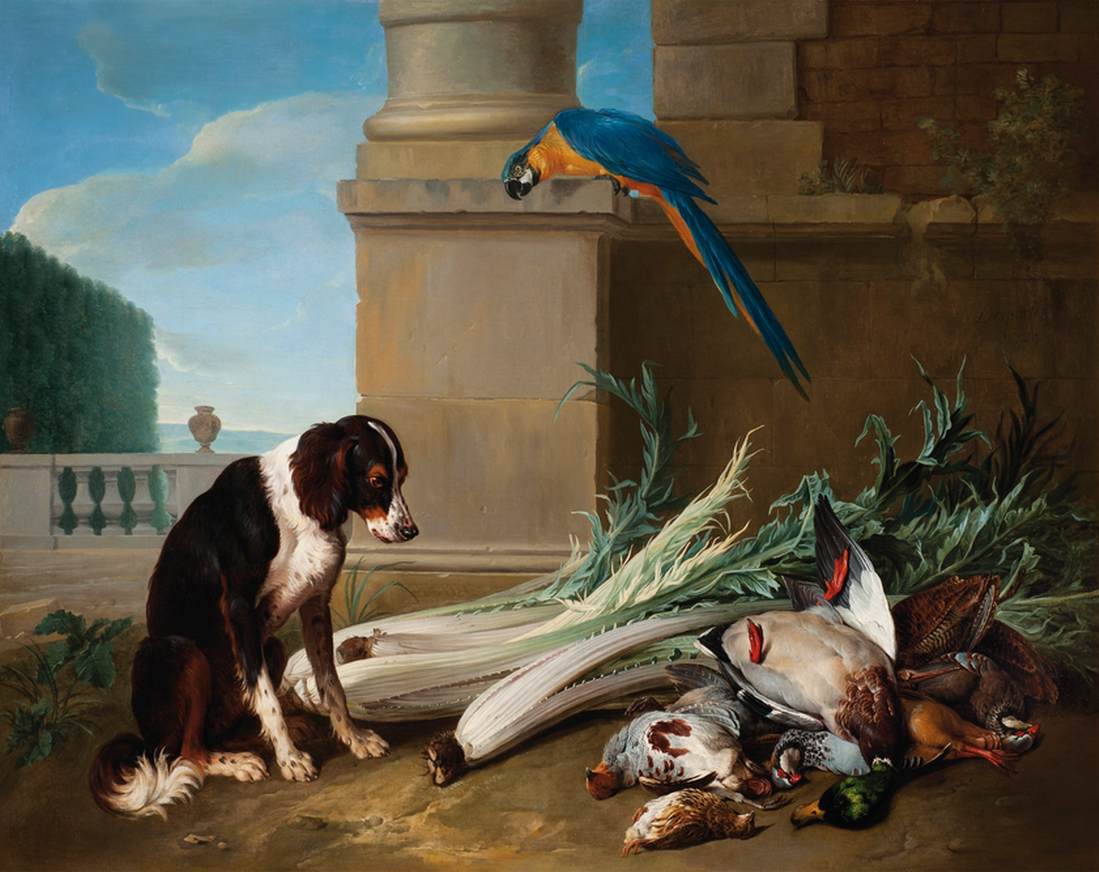} &
    \includegraphics[width=0.21\linewidth]{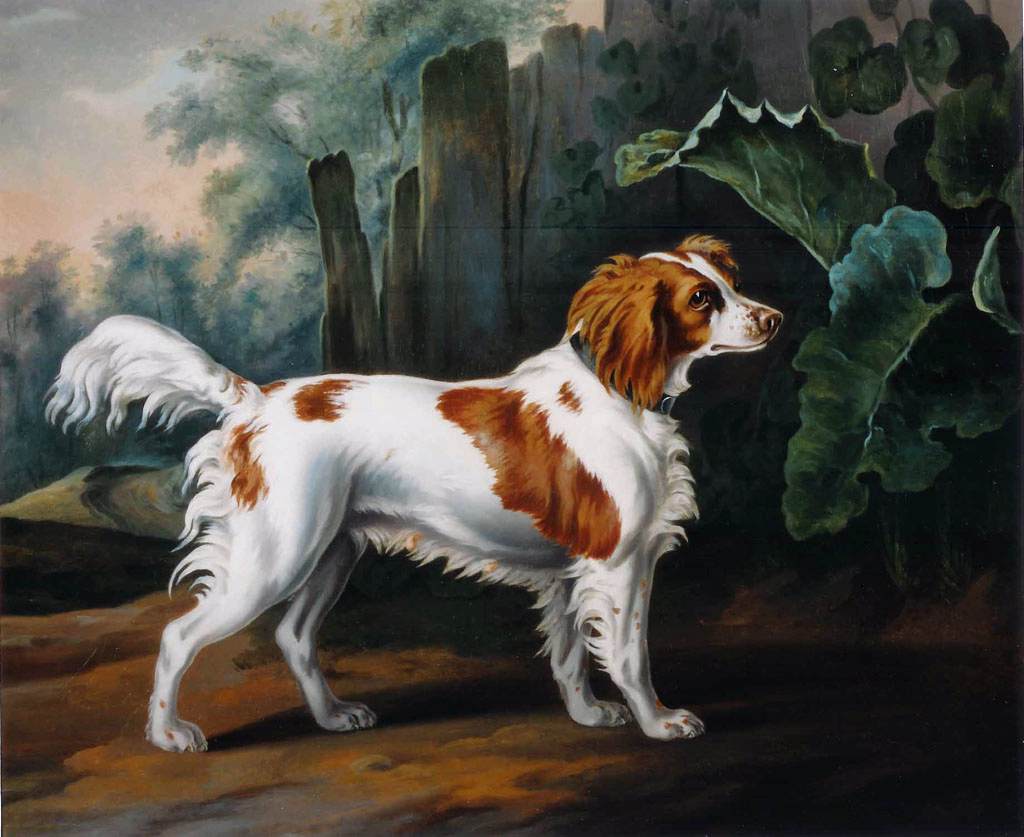} &
    \includegraphics[width=0.21\linewidth]{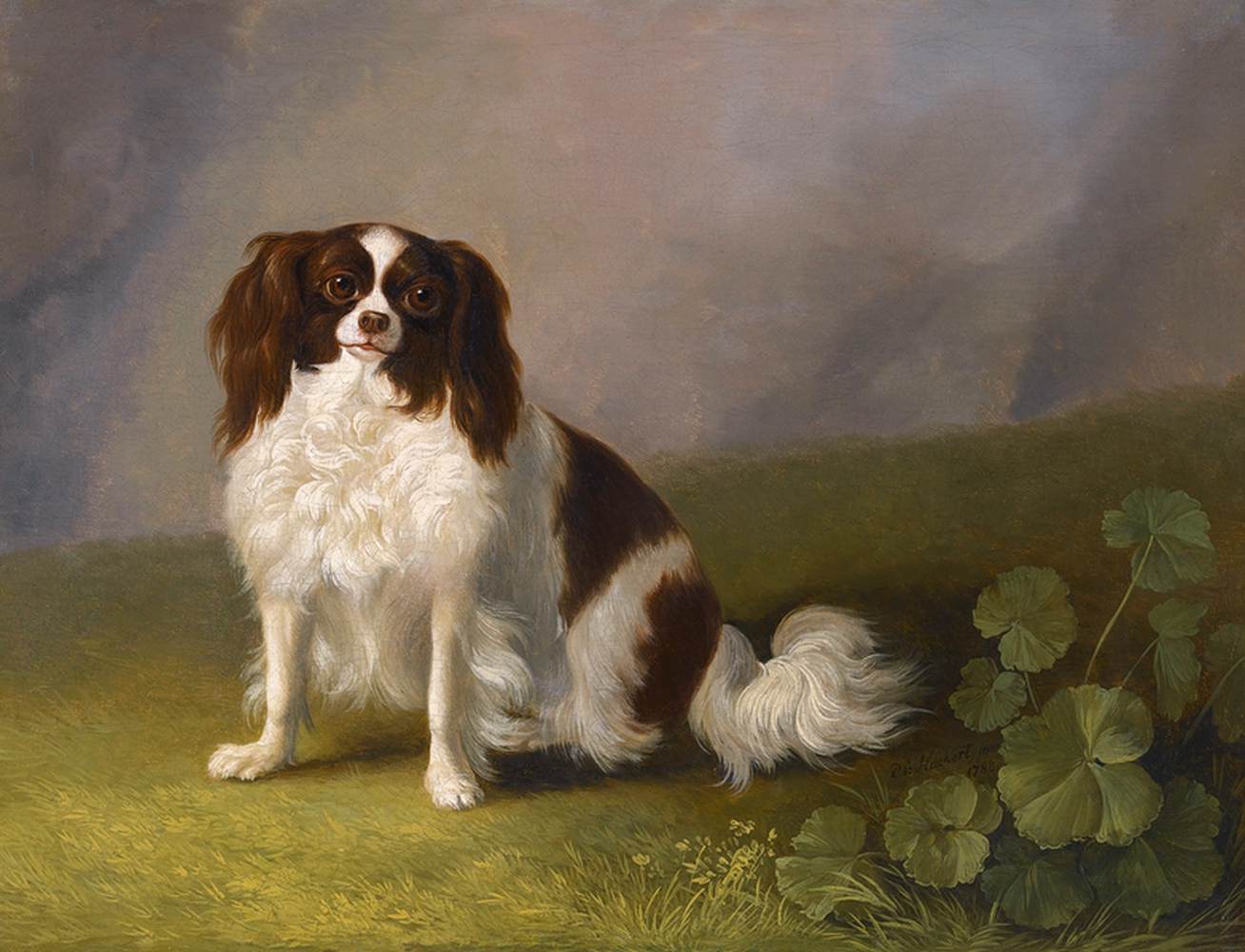} &
    \includegraphics[width=0.21\linewidth]{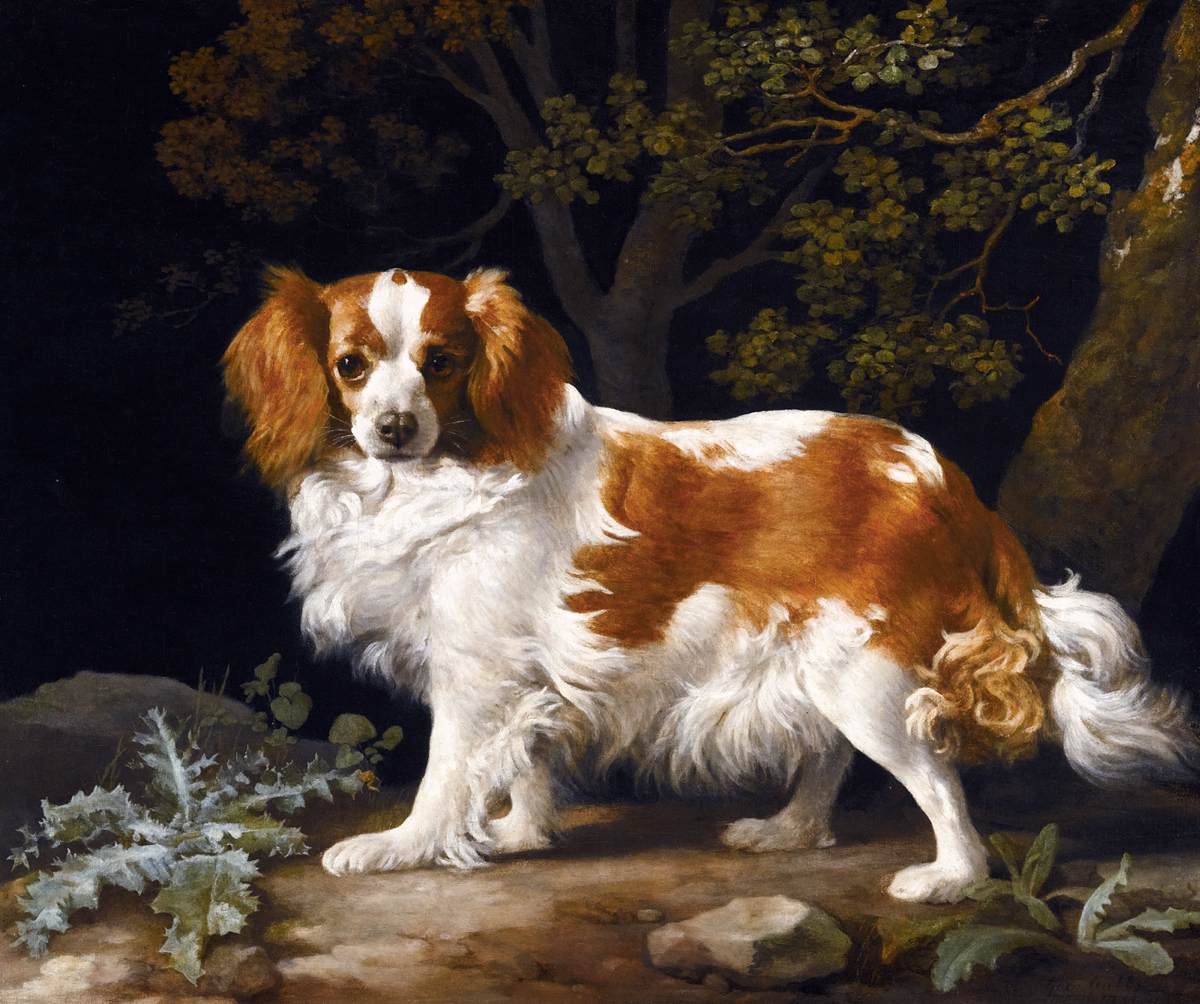} \\  
    \end{tabular}
    \vspace{1em}
    \begin{tabular}{cccc}
    \multicolumn{4}{c}{Shared prototypes} \\
    \cellcolor{LightPurple}\includegraphics[width=0.21\linewidth]{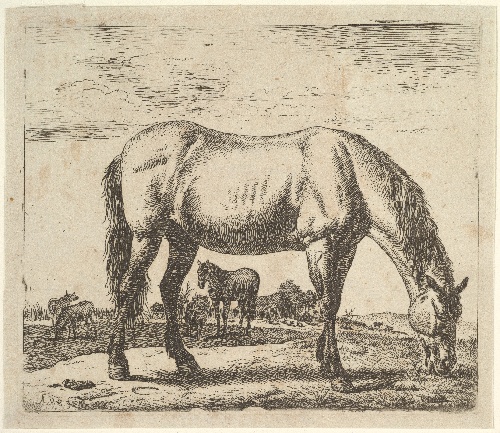} &
    \cellcolor{LightGreen}\includegraphics[width=0.21\linewidth]{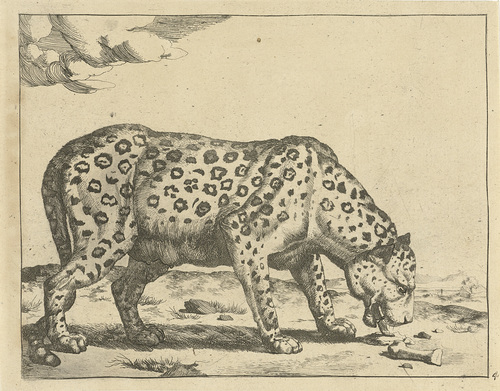} &
    \cellcolor{LightCyan}\includegraphics[width=0.21\linewidth]{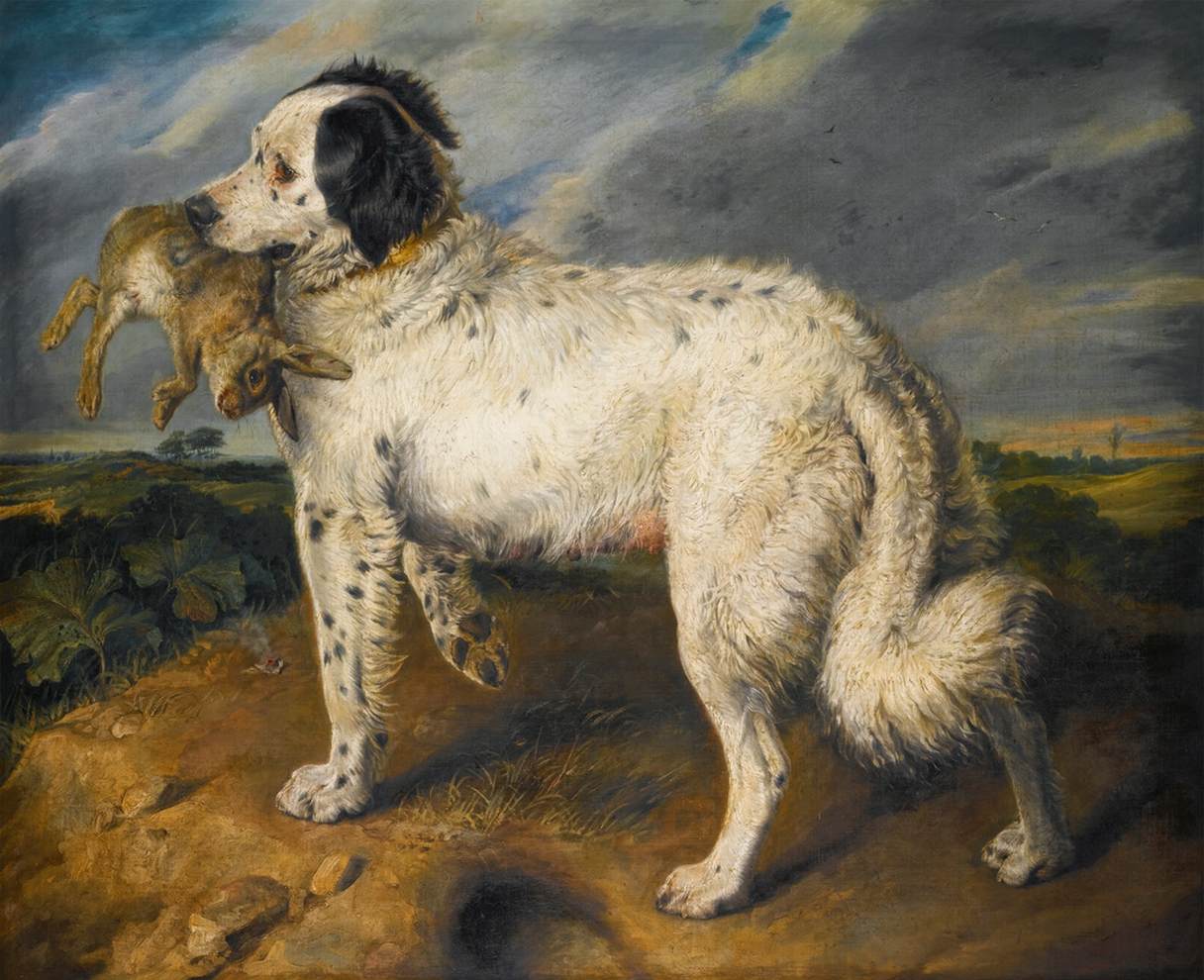} &
    \cellcolor{LightCyan}\includegraphics[width=0.21\linewidth]{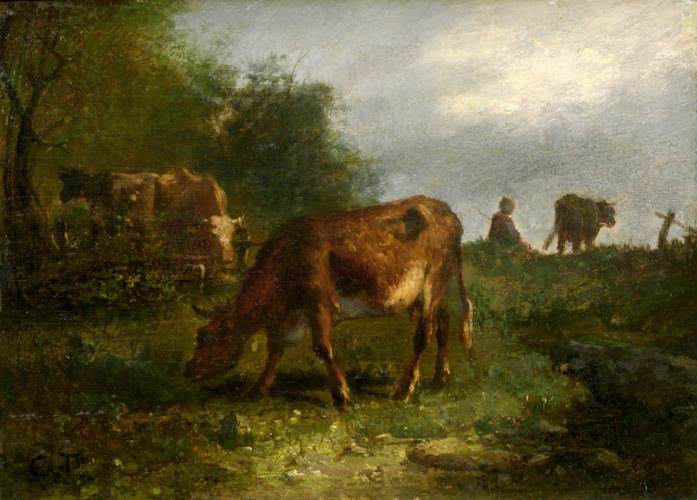} \\
    \cellcolor{LightPurple}\includegraphics[width=0.11\linewidth]{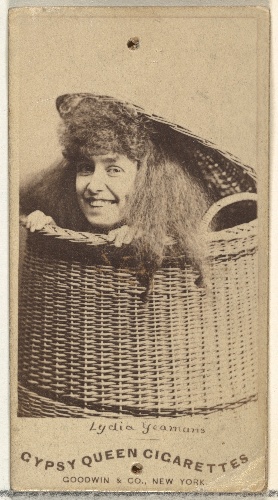} &
    \cellcolor{LightGreen}\includegraphics[width=0.21\linewidth]{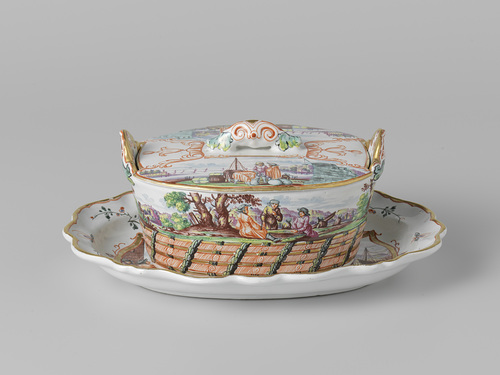} &
    \cellcolor{LightGreen}\includegraphics[width=0.21\linewidth]{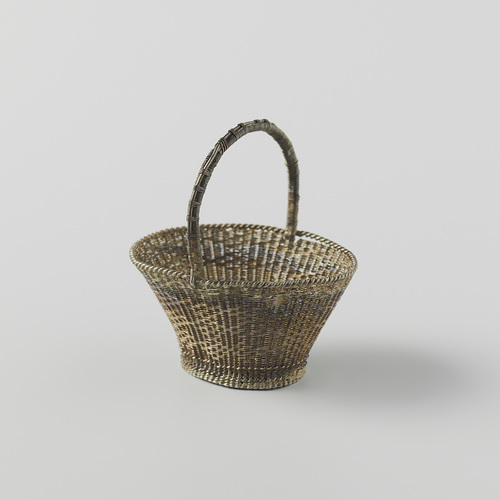} &
    \cellcolor{LightCyan}\includegraphics[width=0.21\linewidth]{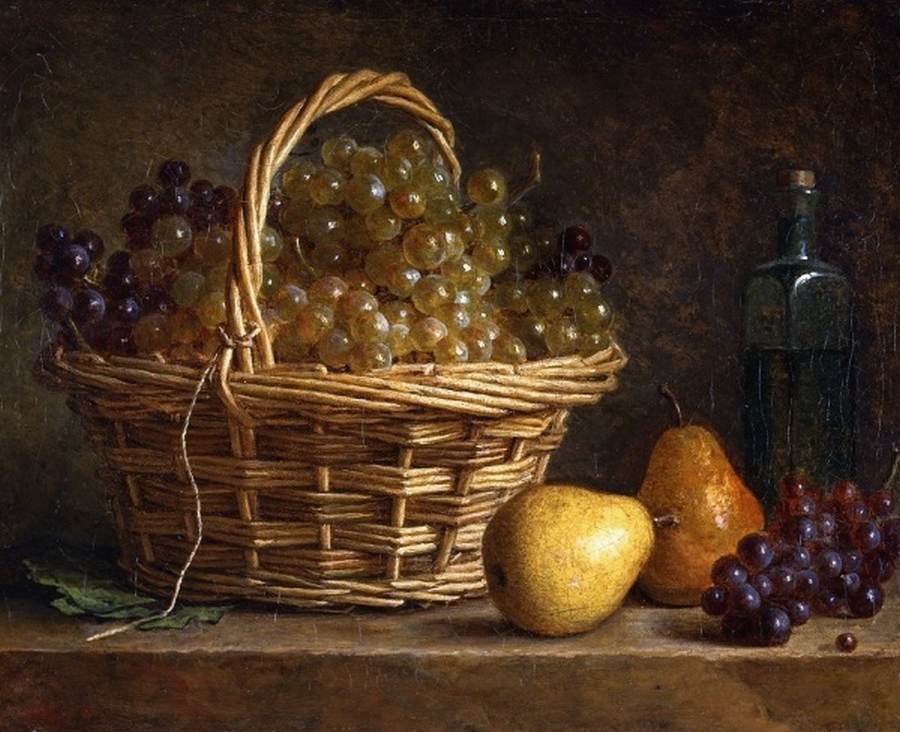} \\   
    \cellcolor{LightPurple}\includegraphics[width=0.21\linewidth]{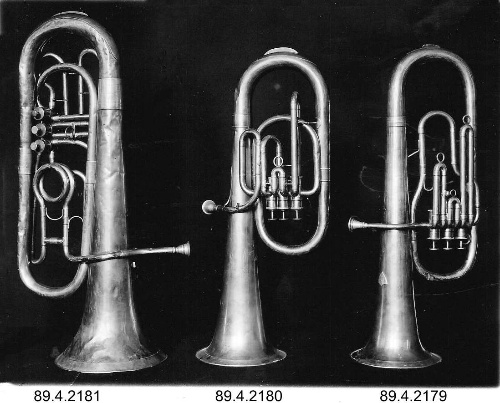} &
    \cellcolor{LightPurple}\includegraphics[width=0.21\linewidth]{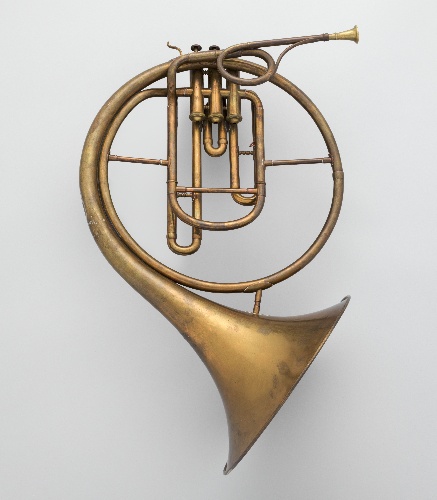} &
    \cellcolor{LightGreen}\includegraphics[width=0.21\linewidth]{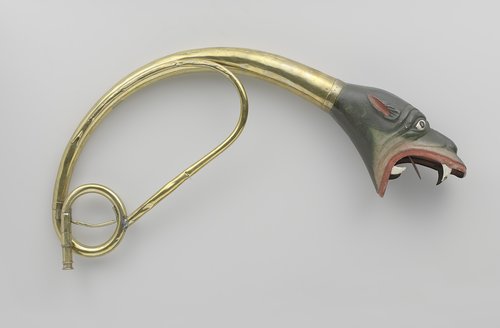} &
    \cellcolor{LightCyan}\includegraphics[width=0.21\linewidth]{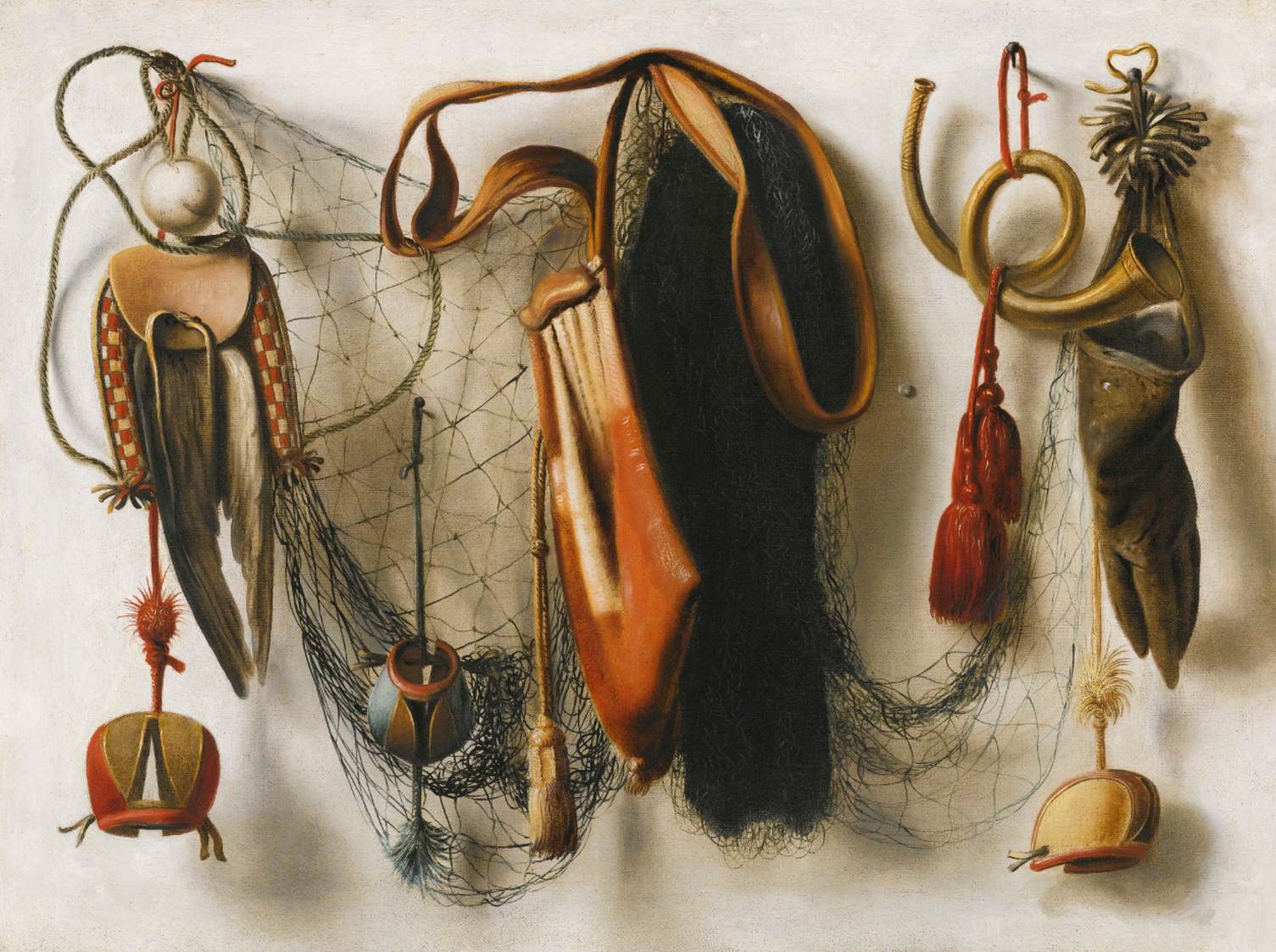} \\
    \end{tabular}}
    \vspace{-1.5em}
    \caption{Example images for nine prototypes, the first six rows are divided in three pairs of prototypes predominantly found in MET, Rijksmuseum, or SemArt respectively. The last three rows show prototypes common across the datasets. Examples with a pink, green, and blue background belong to the MET, Rijksmuseum, or Semart datasets respectively.} 
    \label{fig:artexamples}
\end{figure}
}

\section{Discussion}
\label{sec:discussion}

The prototypes obtained with ProtoSim enable dataset comparison, nonetheless, there are two limitations. Firstly, although explanation methods based on visual examples are preferred for visual data \cite{jeyakumarHowCanExplain2020}, there is a mandatory manual inference step to determine what a prototype represents. Whilst previous methods suffer from the same limitation \cite{chenThisLooksThat2019,kraftSPARROWSemanticallyCoherent2021} ProtoSim does offer an advantage in that the attention maps can be visualised (i.e., Figure~\ref{fig:heatmaps}). Based on example images and their attention maps it becomes possible to be relatively certain about what the prototype represents. 

Secondly, whilst the ViT backbone can be replaced, for our experiments we found the best results with an ImageNet pre-trained backbone. Since this dataset is one of the ones being compared, but also because this dataset has known biases \cite{Yang_Qinami_Fei-Fei_Deng_Russakovsky_2020} the choice for a ImageNet pre-trained backbone may influence the prototypes learned. During experiments no such influence was observed, instead it appears ProtoSim is flexible in learning diverse prototypes, nonetheless further investigation is necessary.

\section{Conclusion}
\label{sec:conclusion}

In this work we presented dataset comparison as a new direction for dataset inspection. To enable dataset comparison across large-scale datasets we introduce ProtoSim which leverages integral prototype-based learning and self-supervised learning to discover visual concepts across datasets. To evaluate our proposed approach we perform two case studies, in the first we compare the PASS and ImageNet datasets, and in the second we  we perform a three-way comparison between artwork datasets.

Based on these case studies we find that we can gain new insights into the datasets. In the first case study we find that ImageNet indeed contains many more images with persons, which is in line with the design goal of PASS. However, we still discovered partial and non-photorealistic depictions of persons in PASS, which were not discovered in its person-focused dataset curation process. When comparing the artwork datasets we find that each has a unique focus, but also that various semantic concepts are shared between them.

Overall, we presented an initial exploration of dataset comparison that will hopefully lead to greater attention for this topic, as it is becoming increasingly necessary to improve dataset inspection techniques.

{\small
\bibliographystyle{ieee_fullname}
\bibliography{lib}
}

\clearpage
\appendix
\section{Appendix}
\subsection{Single dataset summarisation}

In addition to comparison, we also used ProtoSim to summarise ImageNet and PASS independently. Nonetheless, summarising datasets independently does enable some comparison. For instance, ImageNet has a well-documented centre bias (\ie, the salient objects primarily occur in the centre of the image). With ProtoSim we can demonstrate that this is indeed the case, and additionally compare to other datasets. Additionally, we explore if there are structural differences between the prototypes learned for class and patch tokens, and to what extent the prototypes represent semantics. 

\begin{figure}[b]
    \centering
    \begin{tabular}{c c c c c}
        \raisebox{2\height}{Mask} &
        \includegraphics[width=0.15\linewidth]{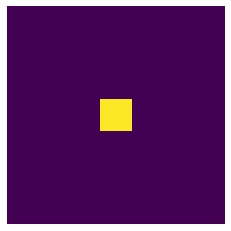} &
        \includegraphics[width=0.15\linewidth]{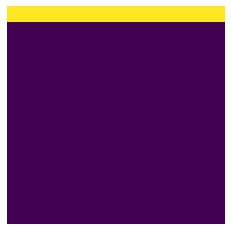} &
        \includegraphics[width=0.15\linewidth]{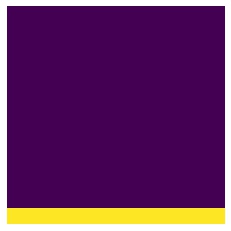} &
        \includegraphics[width=0.15\linewidth]{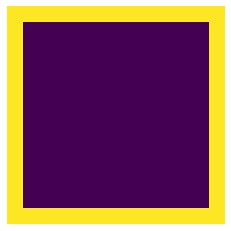} \\
        \raisebox{2\height}{ImageNet} &
        \includegraphics[width=0.15\linewidth]{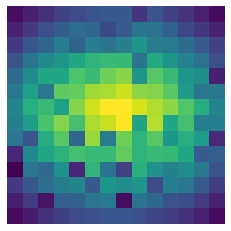} &
        \includegraphics[width=0.15\linewidth]{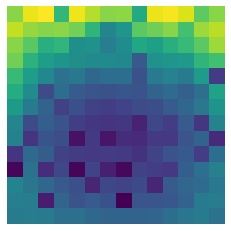} &
        \includegraphics[width=0.15\linewidth]{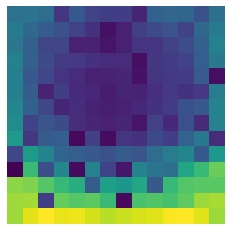} &
        \includegraphics[width=0.15\linewidth]{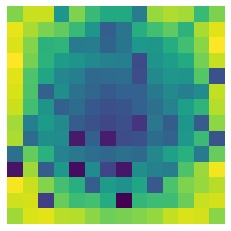} \\
        \raisebox{2\height}{PASS} &
        \includegraphics[width=0.15\linewidth]{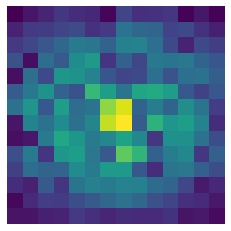} &
        \includegraphics[width=0.15\linewidth]{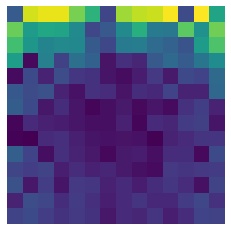} &
        \includegraphics[width=0.15\linewidth]{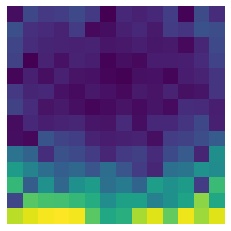} &
        \includegraphics[width=0.15\linewidth]{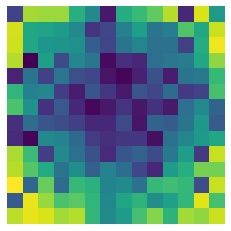} \\
    \end{tabular}
    \caption{Visualisation of between token correlations for selected tokens marked in yellow in the first row. The second row shows ImageNet and the strong (circular) centre bias between tokens, whereas the third row shows a minimal bias for the PASS dataset. A bright (yellow) colour indicates a (correlation) value of 1.} 
    \label{fig:bias}
\end{figure}

\textbf{Centre Bias}. Each of the $14 \times 14$ squares in Figure~\ref{fig:bias} represents a patch token, with its colour indicating the strength of the correlation with the pixels in the mask (first row). The first column shows the correlations between the $4$ central tokens with all other tokens, for ImageNet we observe a clear circular pattern to the correlations, with strong correlations between tokens in the centre. For PASS (third row) this pattern is much less pronounced as all correlations, except  those selected, are much lower. Similar observations can be made for the other columns, when selecting tokens at the top, bottom, or all edges respectively. 

\begin{figure}[t]
    \centering
    \includegraphics[width=\linewidth]{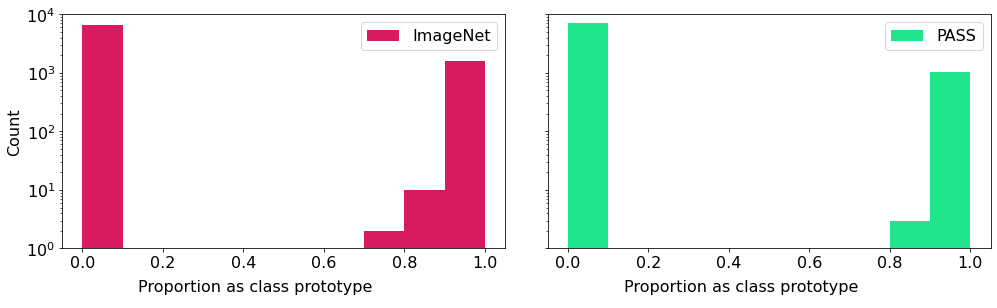}
    \caption{Histograms of the prototype's proportion as class prototype. A proportion of $1.0$ indicates a prototype is always a class prototype and $0$ means it is only ever a spatial prototype.}
    \label{fig:spatialclass}
\end{figure}

\textbf{Class vs. Patch tokens}. ProtoSim learns prototypes for patch tokens and the class token $z^0$, as such prototypes can represent global image information. As there is no architectural difference between how class tokens and patch tokens are treated we can find prototypes for both tokens. However, we find that there is a sharp distinction between \textit{class prototypes} and \textit{spatial prototypes}, with most being spatial tokens as illustrated in Figure~\ref{fig:spatialclass}. 
Notably, there are only a few prototypes that have proportion between $0$ and $0.9$.

\begin{figure}[t]
    \centering
    \includegraphics[width=\linewidth]{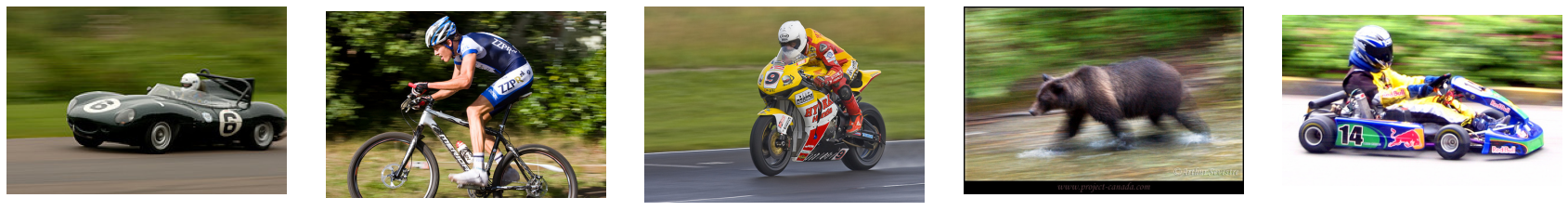}
    \includegraphics[width=\linewidth]{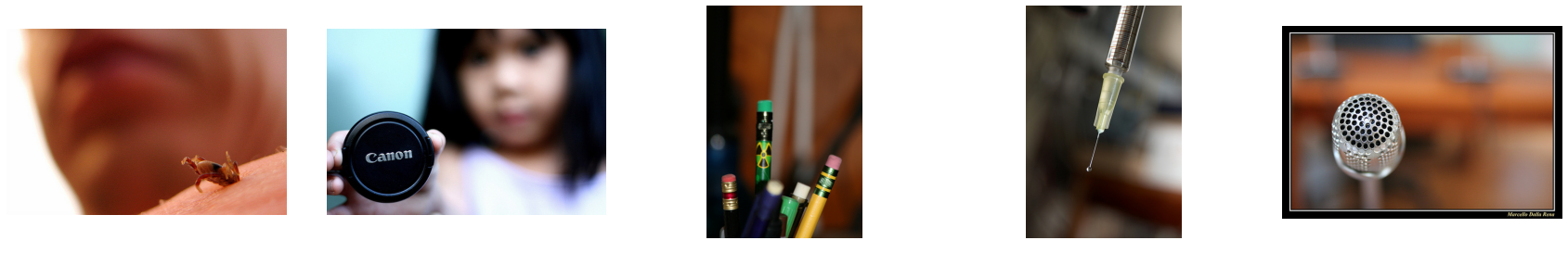}
    \caption{ImageNet prototypes that appear to represent motion blur (first row) and a shallow depth of field (last row).} 
    \label{fig:motionblur}
\end{figure}

\begin{figure}[ht]
    \centering
    \includegraphics[width=\linewidth]{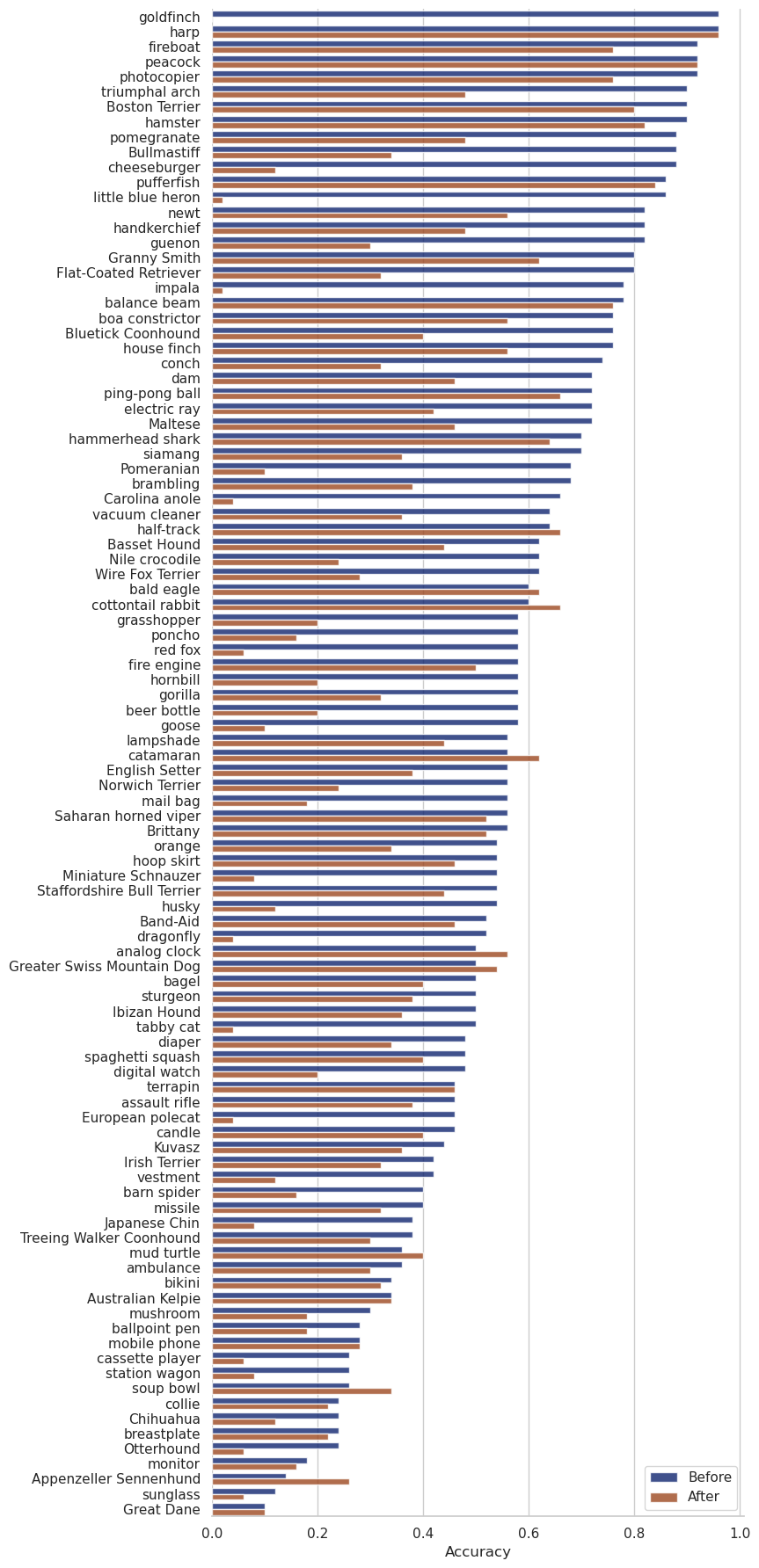}
    \caption{Before and after per-class accuracy of zeroing out the most frequent prototype for each of top 100 ImageNet classes ranked by the strength of their association to a prototype. For the majority of classes we see a stark drop in performance.}
    \label{fig:zeroing}
\end{figure}

\textbf{Semantic Prototypes}. Based on the ImageNet labels we can relate prototypes to categories and explore to what extent there is alignment with the semantic categories. For $874$ out of $1000$ categories the most frequent prototype also has that category as its most frequent category. The remaining $126$ categories that fall in the top 4 for a specific prototype mainly concern one-of a few fine-grained or highly related categories, such as: husky, Alaskan malamute, and Siberian husky; ambulance, police van, and minibus; beer bottle and soda bottle; photocopier and printer; or bikini and tank suit. We can thus conclude that on ImageNet the model learns highly semantic prototypes. In Figure~\ref{fig:zeroing} we experiment with zeroing out the most frequent prototypes for these classes and report their before and after accuracy, we can observe that for most classes this results in a stark drop in performance. On average there is a difference of 22.6 percentage points between before and after zeroing, thereby highlighting how strongly class-specific these prototypes are. In addition to semantic prototypes it also learns non-semantic visual effects such as motion blur or a shallow depth of field, as shown in Figure~\ref{fig:motionblur}.

\onecolumn
\subsection{Additional PassNet Prototypes}

%
%
%
%

\setlength{\tabcolsep}{0.4em} 
\begin{figure*}[h]
\sffamily
    \begin{tabular}{dddddd}
    \includegraphics[width=0.14\linewidth]{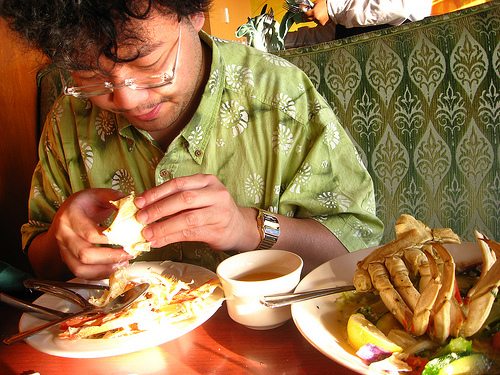} &
    \includegraphics[width=0.14\linewidth]{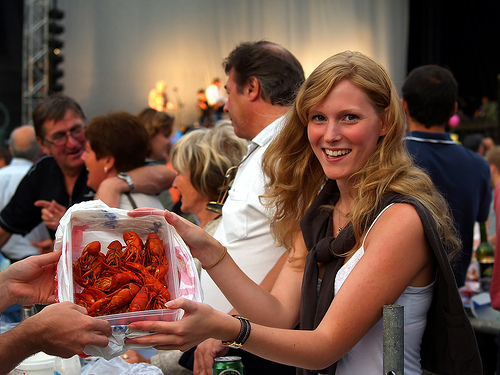} &
    \includegraphics[width=0.14\linewidth]{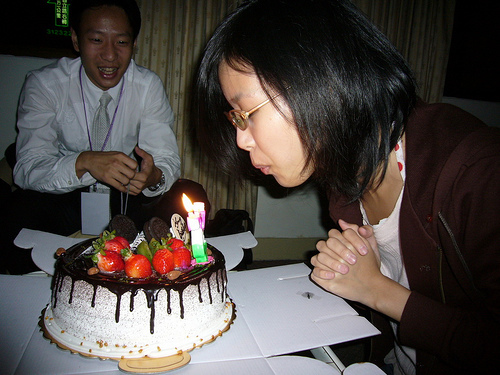} &
    \includegraphics[width=0.14\linewidth]{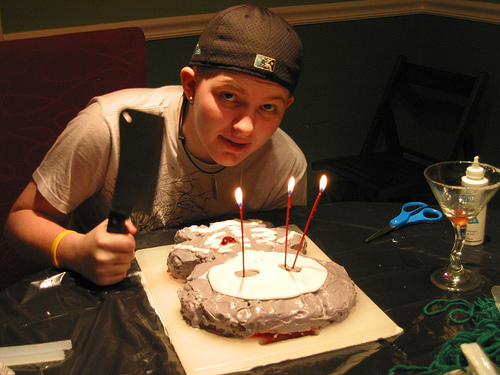} &
    \includegraphics[width=0.14\linewidth]{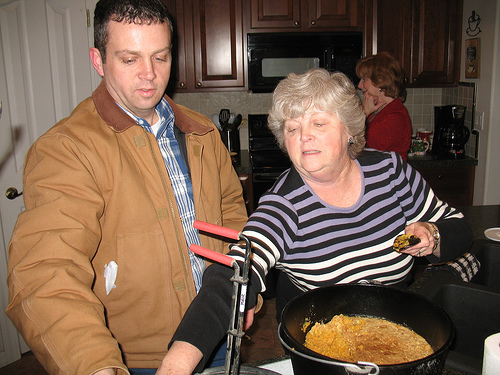} &
    \includegraphics[width=0.07\linewidth]{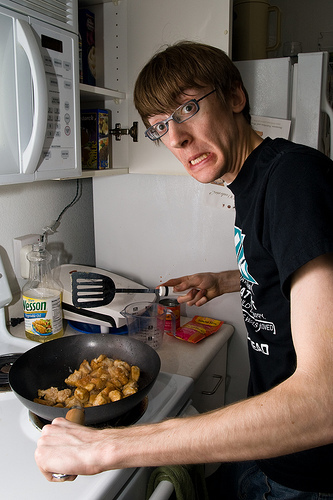} \\
    \includegraphics[width=0.14\linewidth]{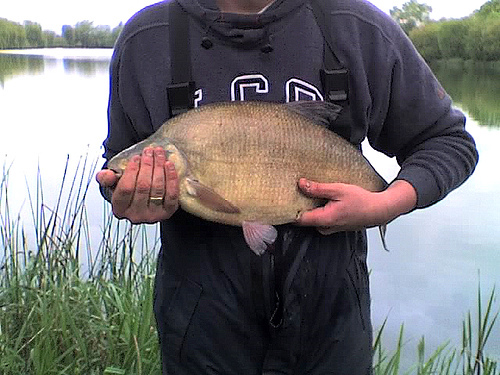} &
    \includegraphics[width=0.14\linewidth]{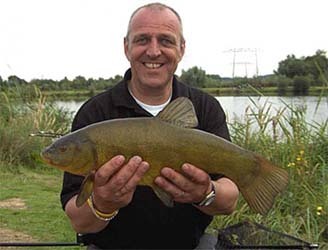} &
    \includegraphics[width=0.14\linewidth]{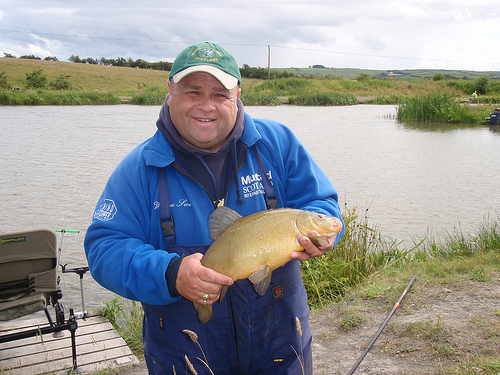} &
    \includegraphics[width=0.14\linewidth]{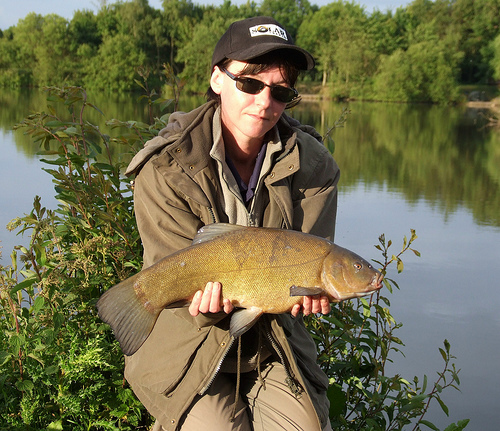} &
    \includegraphics[width=0.14\linewidth]{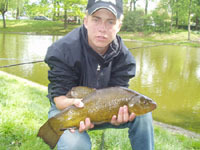} &
    \includegraphics[width=0.14\linewidth]{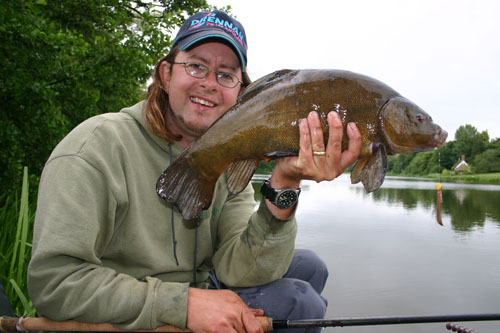} \\
    \includegraphics[width=0.14\linewidth]{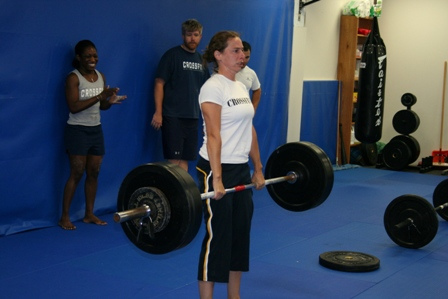} &
    \includegraphics[width=0.14\linewidth]{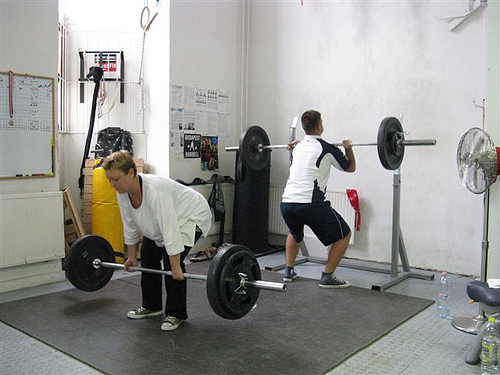} &
    \includegraphics[width=0.14\linewidth]{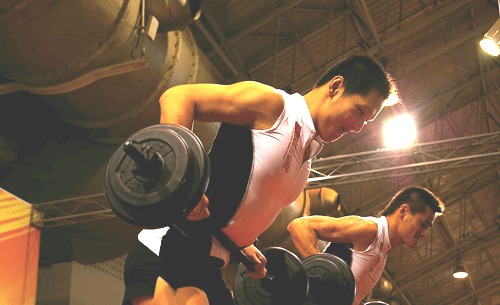} &
    \includegraphics[width=0.14\linewidth]{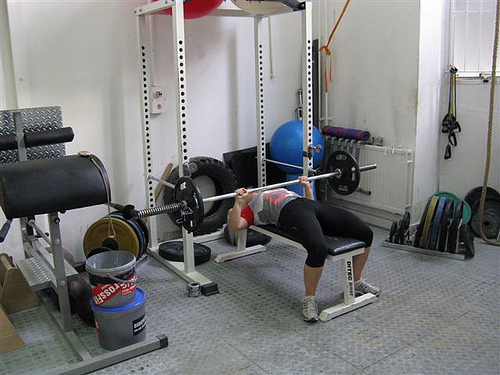} &
    \includegraphics[width=0.14\linewidth]{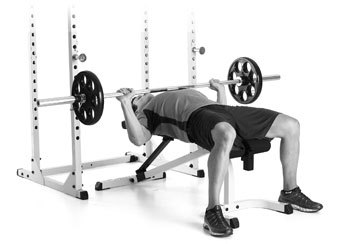} &
    \includegraphics[width=0.14\linewidth]{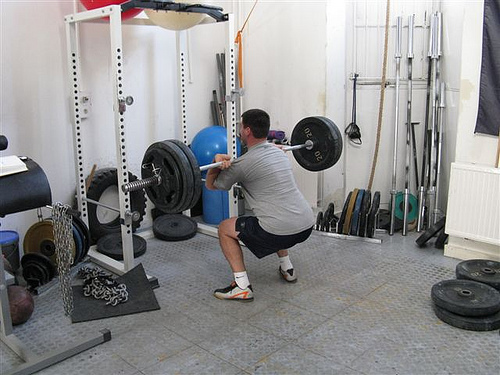} \\
    \end{tabular}
    \caption{Three prototypes only found in ImageNet: \textit{Persons eating food}, \textit{Person holding fish}, and \textit{Weightlifting} respectively.} 
\end{figure*}
\begin{figure*}[h]
\sffamily
    \begin{tabular}{dddd!{\color{white}\vrule width 4pt}bb}
    \includegraphics[width=0.14\linewidth]{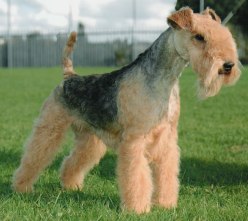} &
    \includegraphics[width=0.14\linewidth]{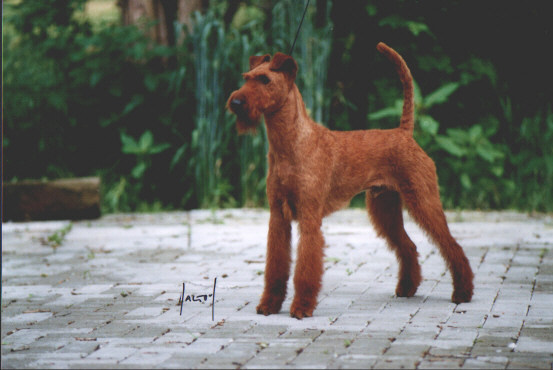} &
    \includegraphics[width=0.09\linewidth]{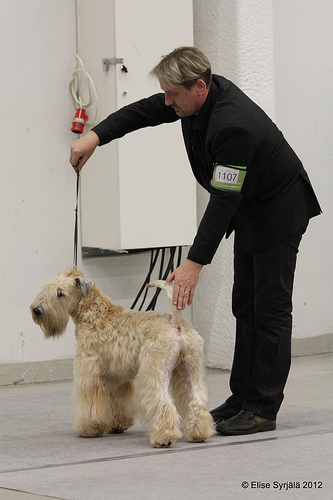} &
    \includegraphics[width=0.14\linewidth]{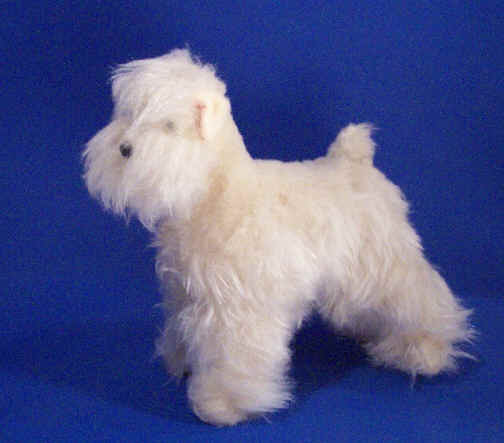} &
    \includegraphics[width=0.14\linewidth]{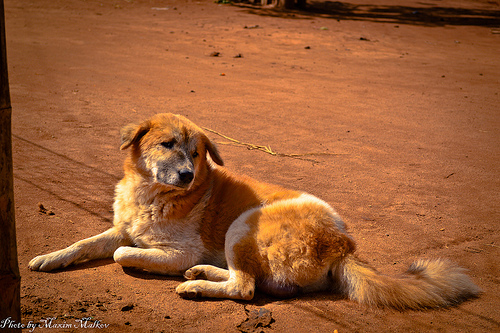} &
    \includegraphics[width=0.14\linewidth]{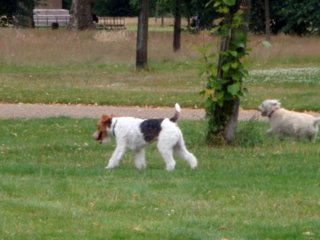} \\
    \includegraphics[width=0.10\linewidth]{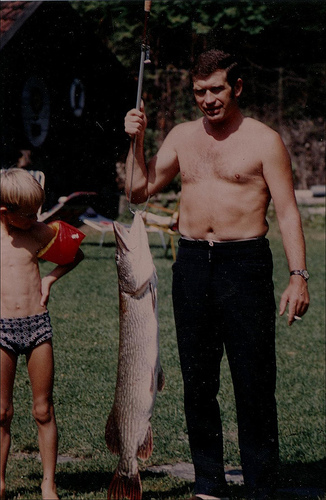} &
    \includegraphics[width=0.14\linewidth]{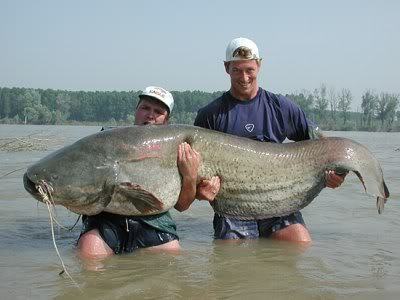} &
    \includegraphics[width=0.10\linewidth]{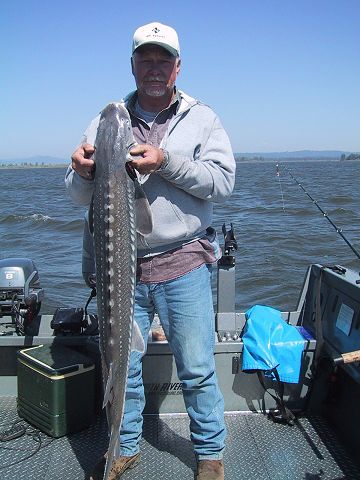} &
    \includegraphics[width=0.14\linewidth]{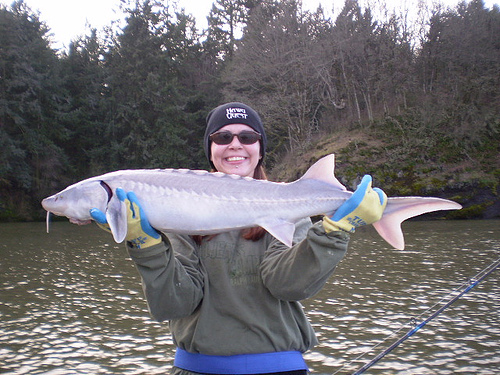} &
    \includegraphics[width=0.10\linewidth]{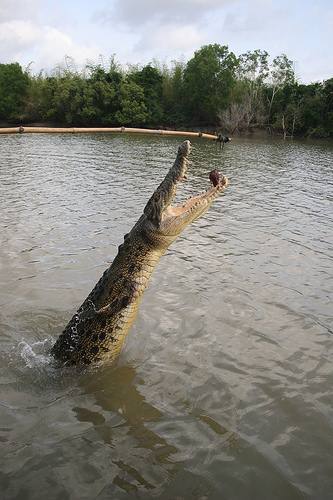} &
    \includegraphics[width=0.10\linewidth]{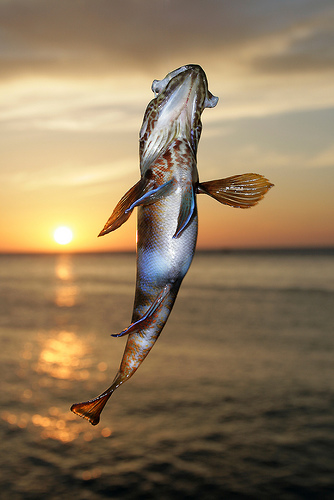} \\
    \includegraphics[width=0.10\linewidth]{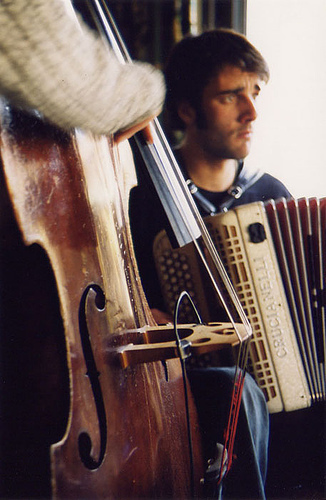} &
    \includegraphics[width=0.10\linewidth]{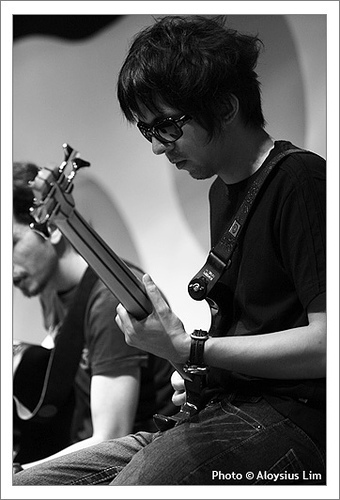} &
    \includegraphics[width=0.10\linewidth]{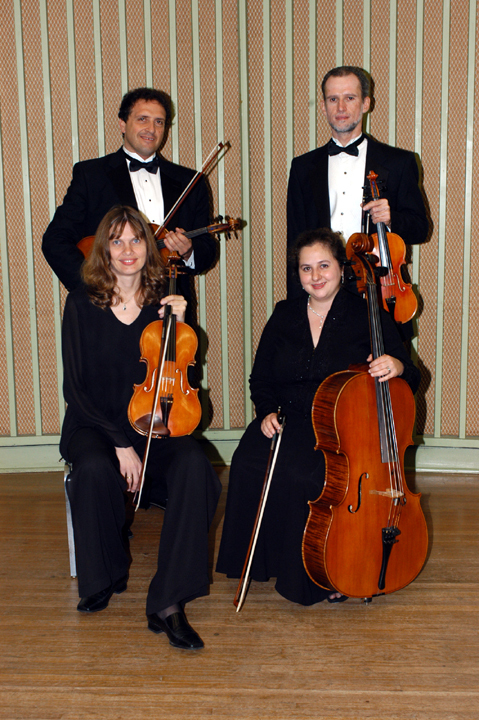} &
    \includegraphics[width=0.10\linewidth]{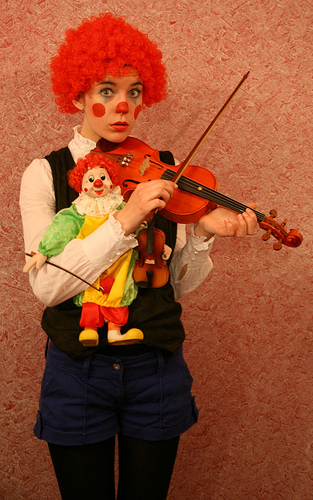} &
    \includegraphics[width=0.10\linewidth]{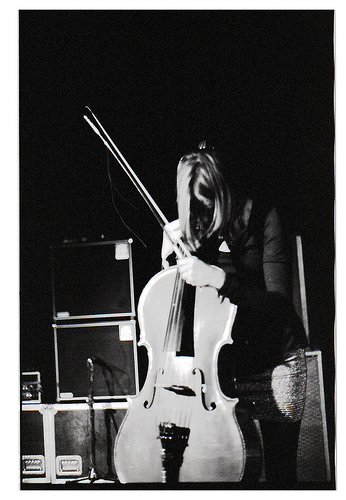} &
    \includegraphics[width=0.10\linewidth]{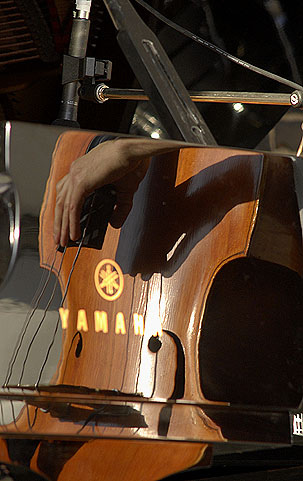} \\
    \end{tabular}
    \caption{Three predominantly ImageNet prototypes: \textit{Dogs}, \textit{Persons holding large fish}, and \textit{Persons with instrument} respectively. Last two columns are PASS images.} 
\end{figure*}

%
%
%
%
%

\begin{figure*}[b]
\sffamily
    \begin{tabular}{bbbbbb}
    \includegraphics[width=0.14\linewidth]{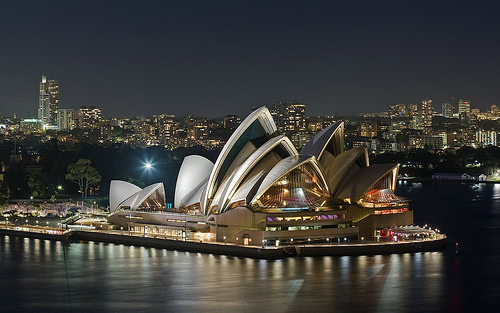} &
    \includegraphics[width=0.14\linewidth]{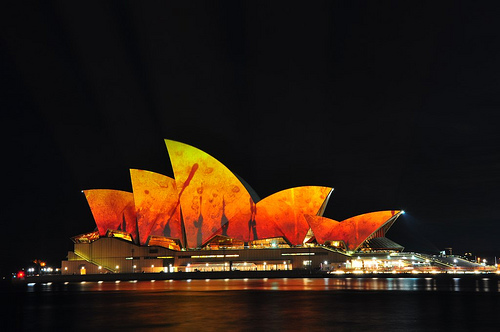} &
    \includegraphics[width=0.14\linewidth]{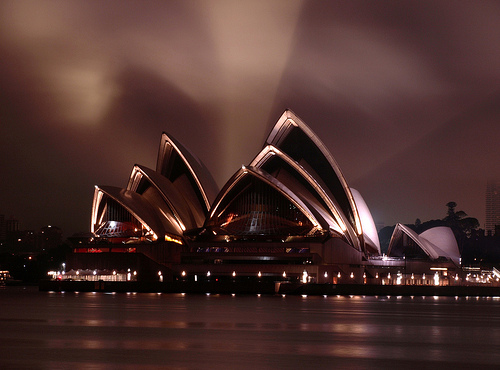} &
    \includegraphics[width=0.14\linewidth]{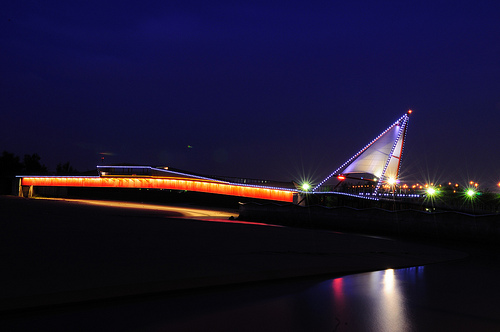} &
    \includegraphics[width=0.14\linewidth]{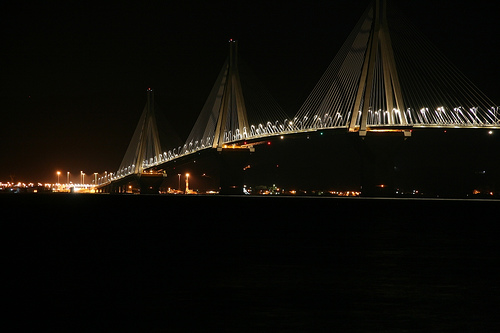} &
    \includegraphics[width=0.14\linewidth]{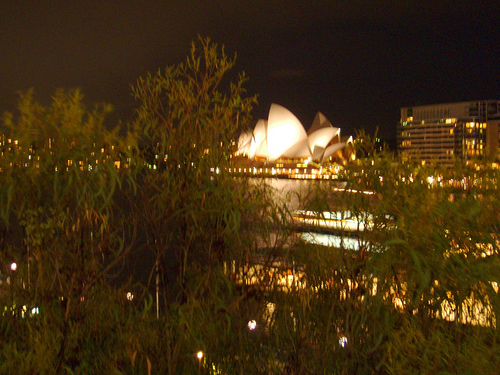} \\
    \includegraphics[width=0.14\linewidth]{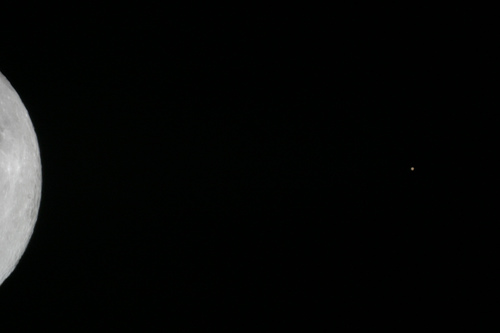} &
    \includegraphics[width=0.14\linewidth]{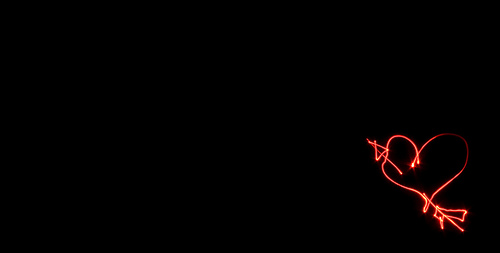} &
    \includegraphics[width=0.14\linewidth]{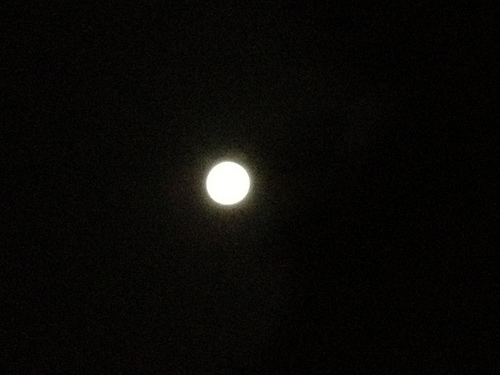} &
    \includegraphics[width=0.14\linewidth]{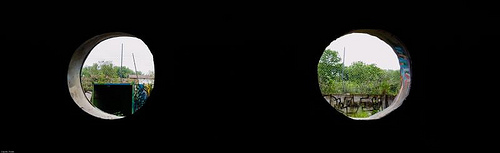} &
    \includegraphics[width=0.14\linewidth]{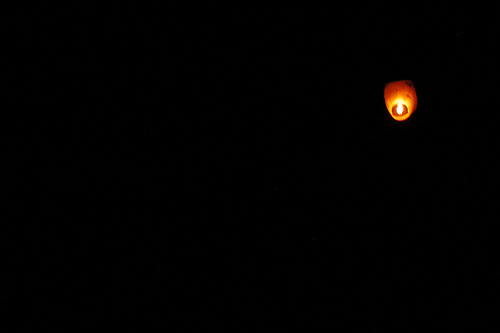} &
    \includegraphics[width=0.14\linewidth]{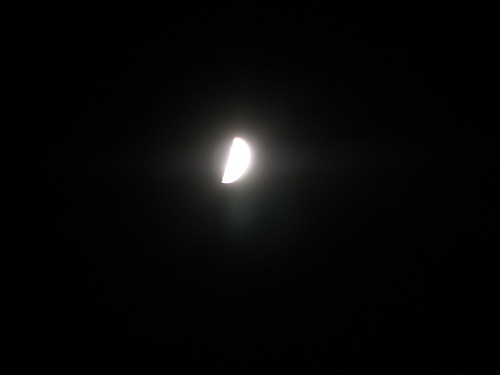} \\
    \includegraphics[width=0.14\linewidth]{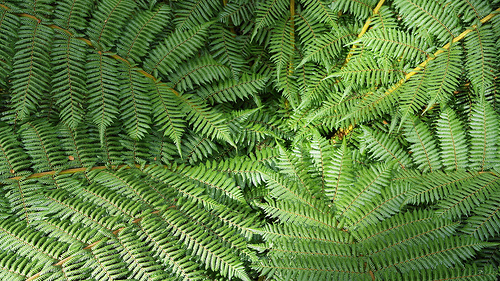} &
    \includegraphics[width=0.14\linewidth]{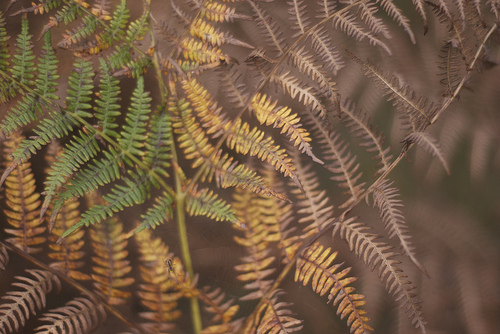} &
    \includegraphics[width=0.14\linewidth]{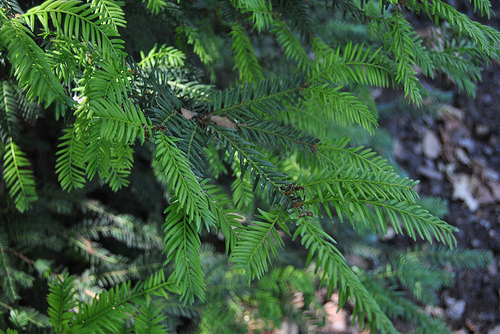} &
    \includegraphics[width=0.14\linewidth]{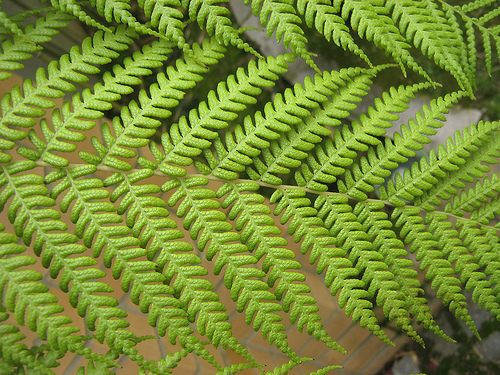} &
    \includegraphics[width=0.14\linewidth]{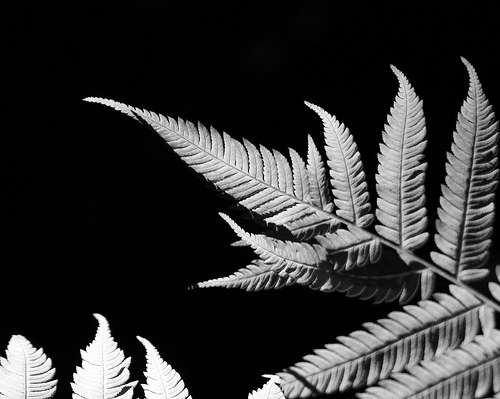} &
    \includegraphics[width=0.14\linewidth]{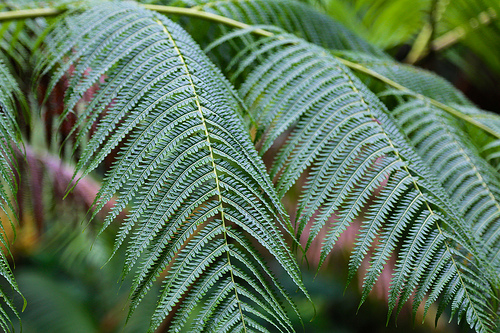} \\
    \end{tabular}
    \caption{Three prototypes only found in PASS: \textit{Illuminated structure}, \textit{Darkness with bright area/celestial body}, and \textit{Palm fronds} respectively.} 
\end{figure*}
\begin{figure*}[b]
\sffamily
    \begin{tabular}{bbbb!{\color{white}\vrule width 4pt}dd}
    \includegraphics[width=0.14\linewidth]{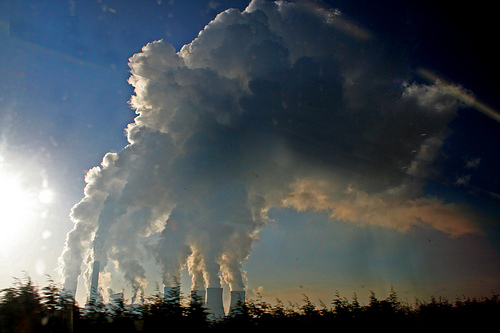} &
    \includegraphics[width=0.14\linewidth]{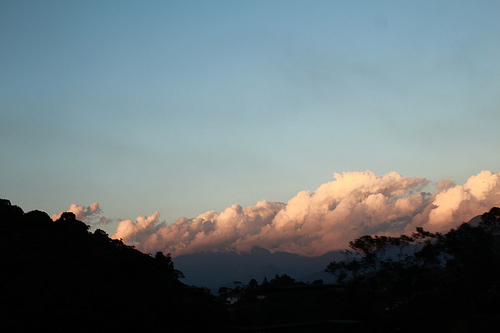} &
    \includegraphics[width=0.14\linewidth]{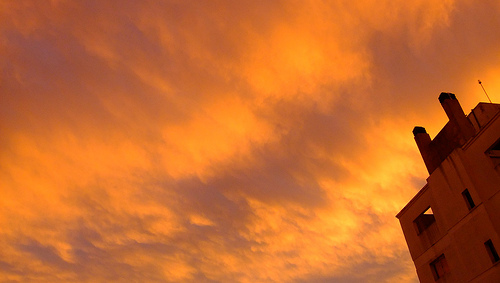} &
    \includegraphics[width=0.14\linewidth]{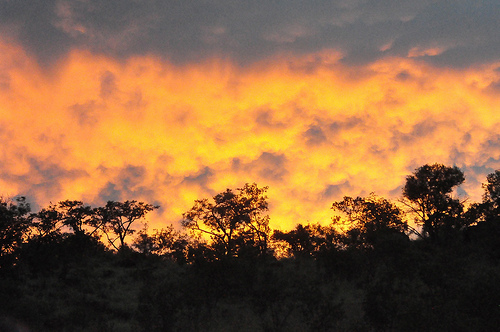} &
    \includegraphics[width=0.14\linewidth]{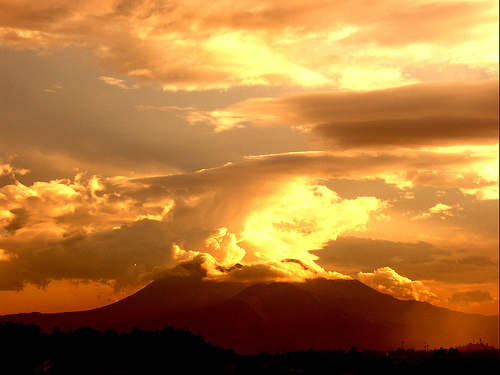} &
    \includegraphics[width=0.14\linewidth]{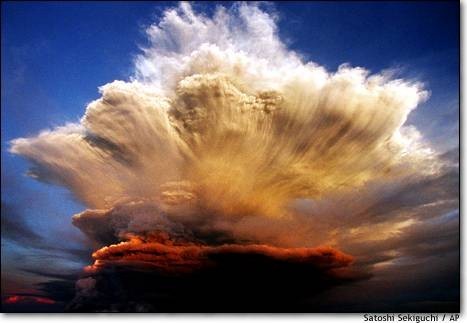} \\
    \includegraphics[width=0.14\linewidth]{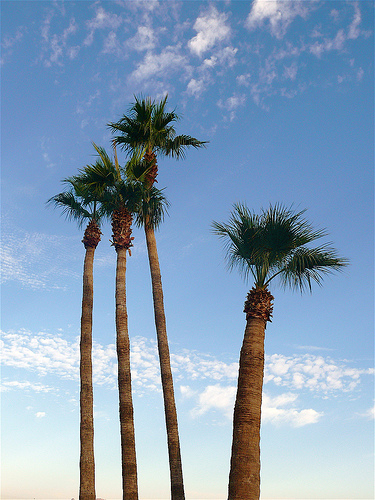} &
    \includegraphics[width=0.14\linewidth]{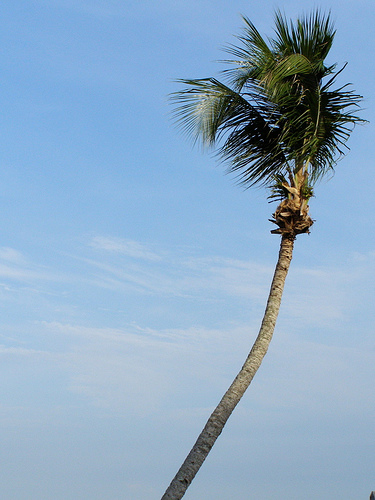} &
    \includegraphics[width=0.14\linewidth]{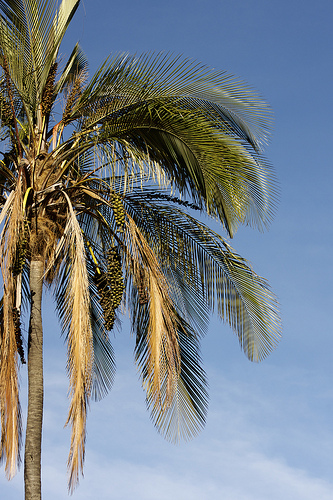} &
    \includegraphics[width=0.14\linewidth]{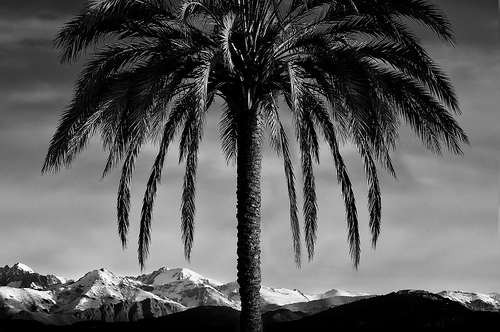} &
    \includegraphics[width=0.14\linewidth]{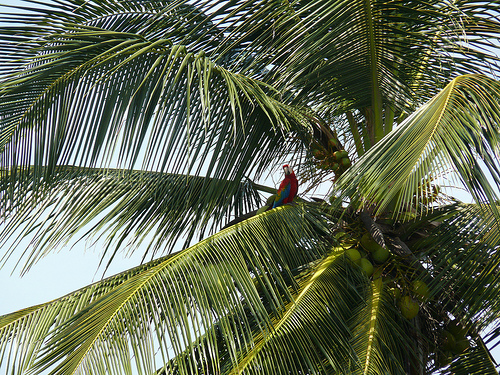} &
    \includegraphics[width=0.14\linewidth]{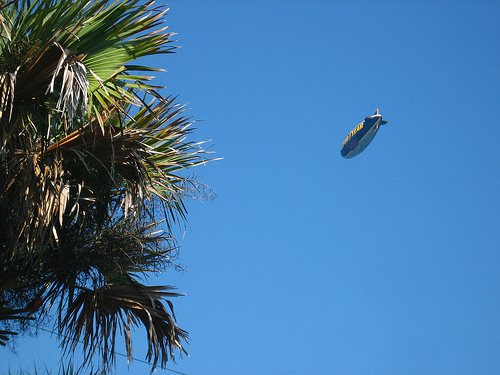} \\
    \includegraphics[width=0.14\linewidth]{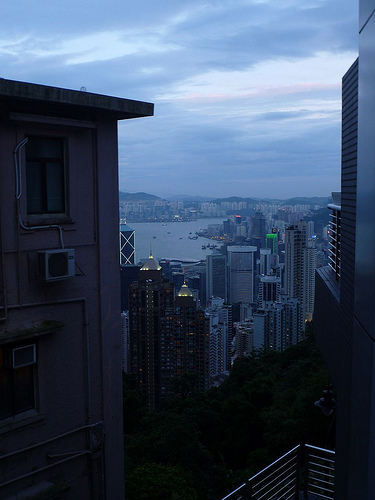} &
    \includegraphics[width=0.14\linewidth]{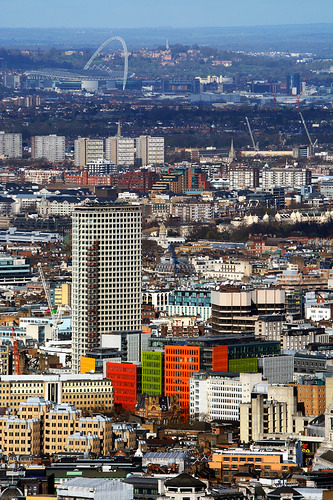} &
    \includegraphics[width=0.14\linewidth]{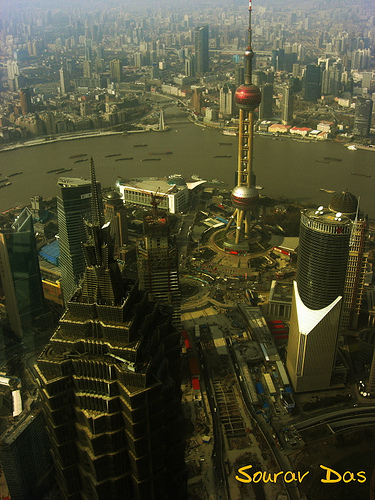} &
    \includegraphics[width=0.14\linewidth]{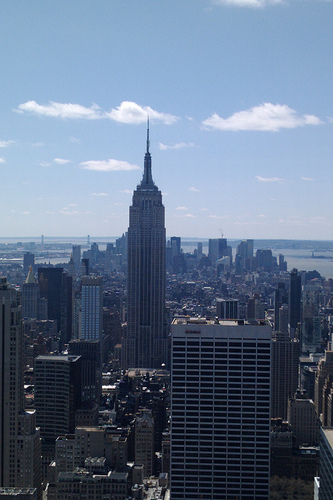} &
    \includegraphics[width=0.14\linewidth]{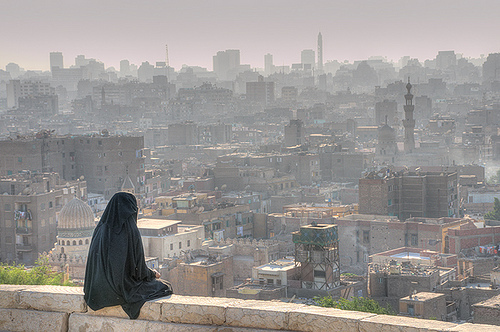} &
    \includegraphics[width=0.14\linewidth]{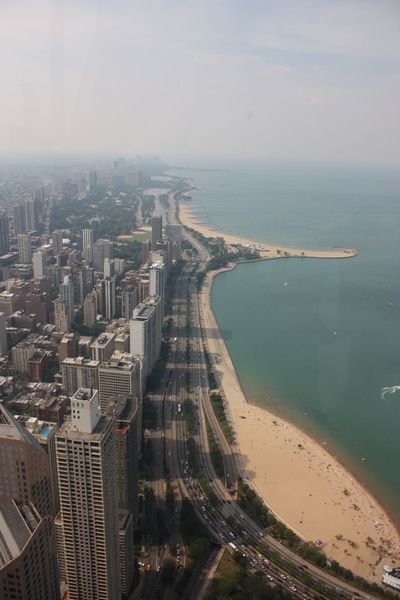} \\
    \end{tabular}
    \caption{Three prototypes predominantly found in PASS: \textit{Sunlit clouds}, \textit{Palm trees}, and \textit{Cityscape} respectively. The last two columns are images found in ImageNet.} 
\end{figure*}

%
\begin{figure*}[b]
\sffamily
    \begin{tabular}{ddd!{\color{white}\vrule width 4pt}bbb}
    \includegraphics[width=0.14\linewidth]{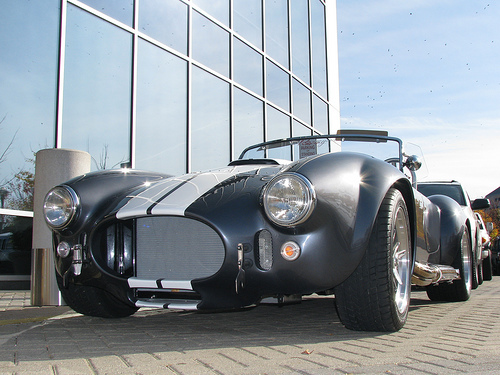} &
    \includegraphics[width=0.14\linewidth]{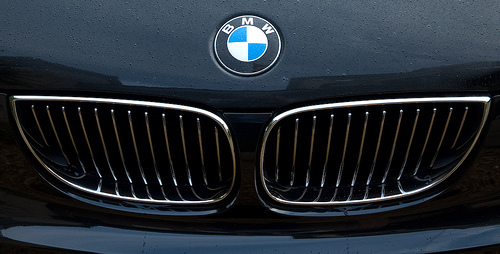} &
    \includegraphics[width=0.14\linewidth]{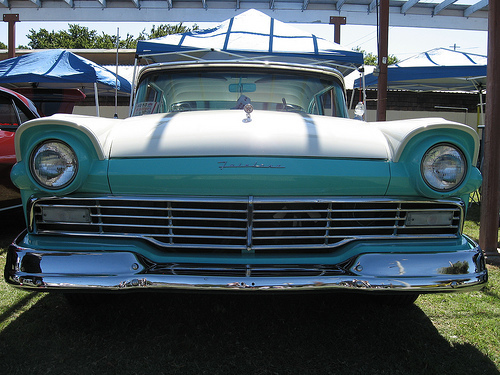} &
    \includegraphics[width=0.14\linewidth]{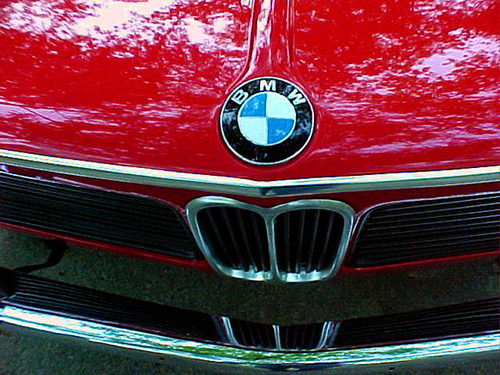} &
    \includegraphics[width=0.14\linewidth]{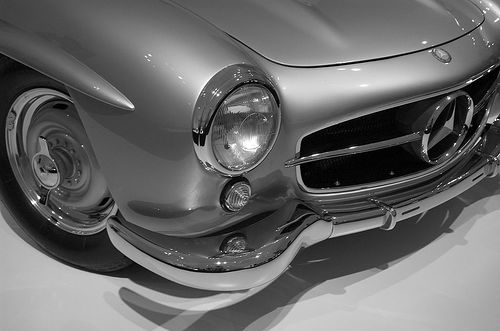} &
    \includegraphics[width=0.08\linewidth]{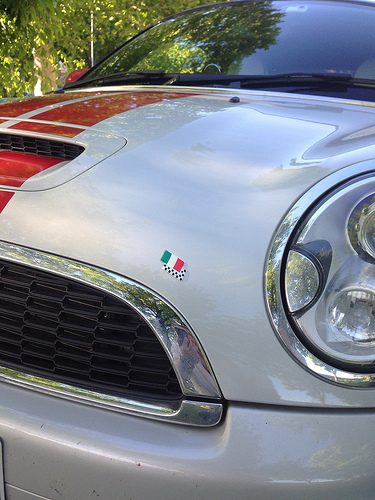} \\
    \includegraphics[width=0.14\linewidth]{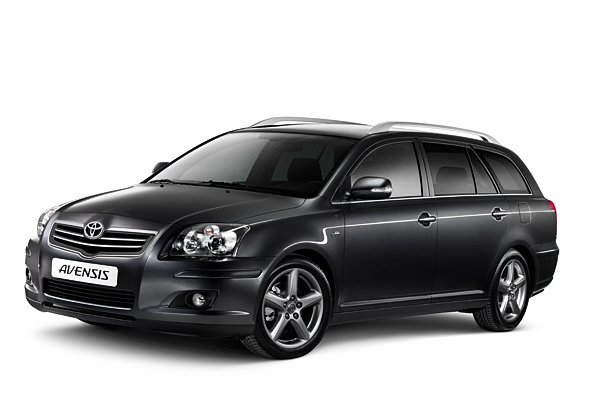} &
    \includegraphics[width=0.14\linewidth]{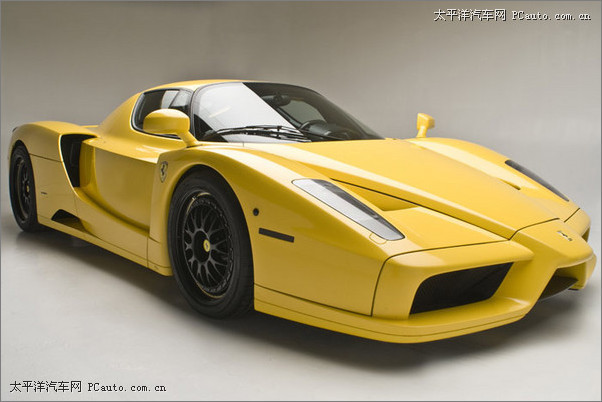} &
    \includegraphics[width=0.14\linewidth]{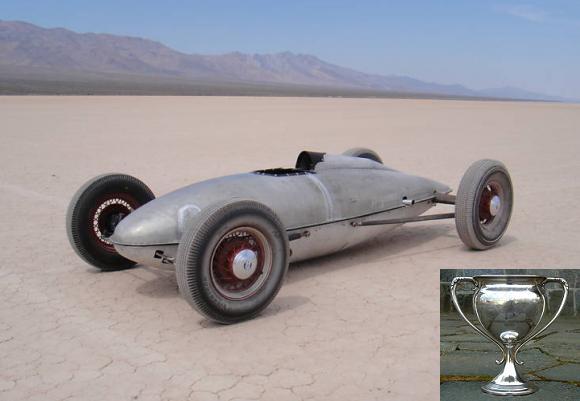} &
    \includegraphics[width=0.14\linewidth]{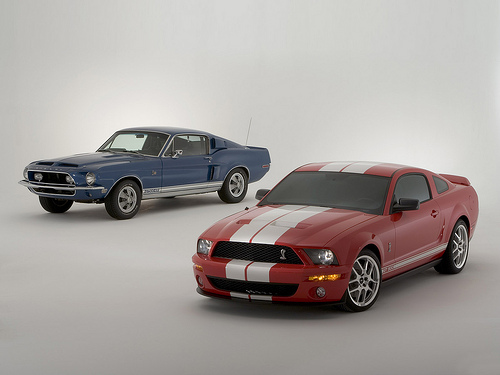} &
    \includegraphics[width=0.14\linewidth]{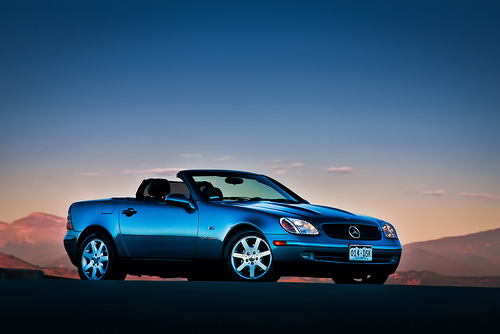} &
    \includegraphics[width=0.14\linewidth]{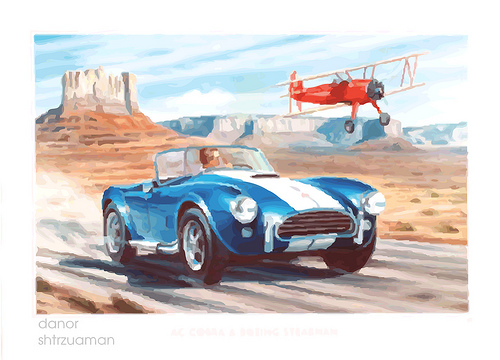} \\
    \includegraphics[width=0.14\linewidth]{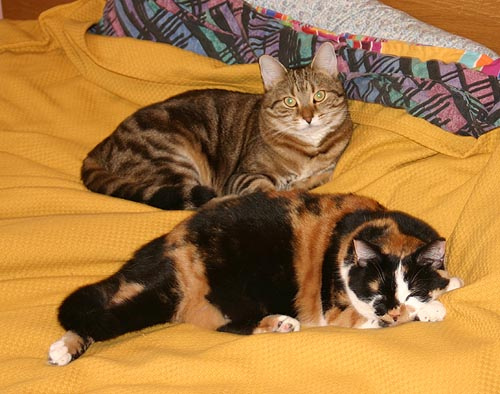} &
    \includegraphics[width=0.14\linewidth]{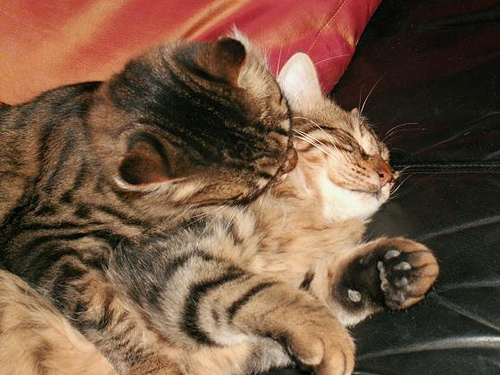} &
    \includegraphics[width=0.14\linewidth]{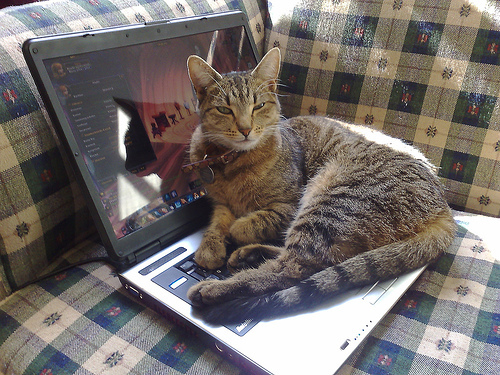} &
    \includegraphics[width=0.14\linewidth]{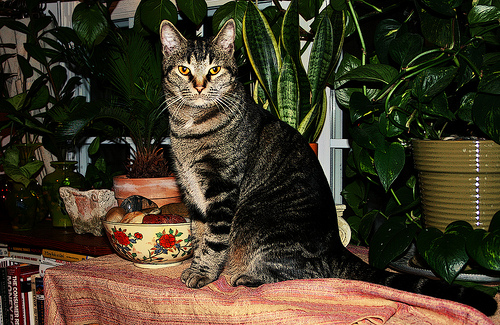} &
    \includegraphics[width=0.14\linewidth]{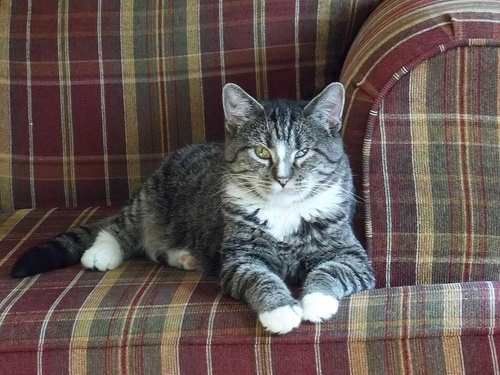} &
    \includegraphics[width=0.14\linewidth]{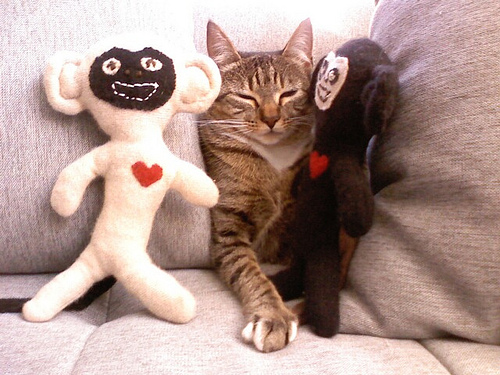} \\
    \includegraphics[width=0.14\linewidth]{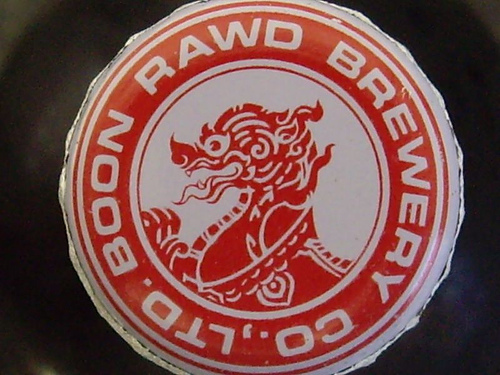} &
    \includegraphics[width=0.12\linewidth]{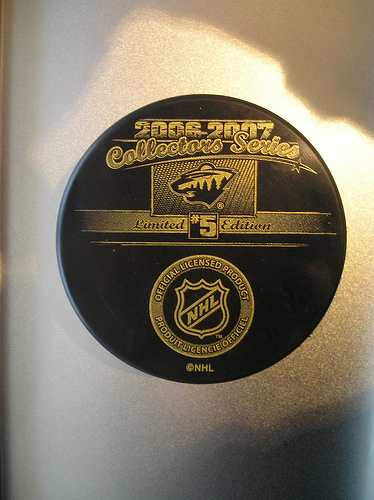} &
    \includegraphics[width=0.14\linewidth]{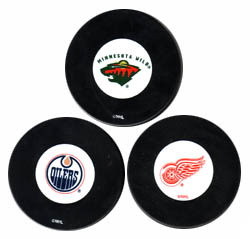} &
    \includegraphics[width=0.12\linewidth]{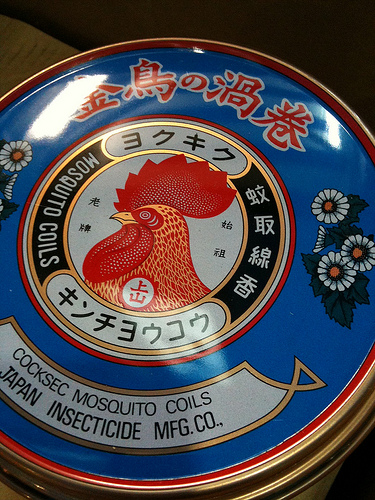} &
    \includegraphics[width=0.14\linewidth]{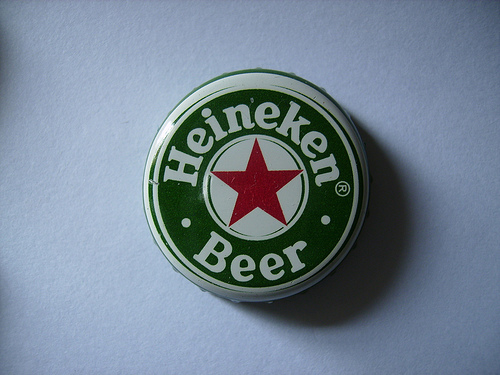} &
    \includegraphics[width=0.14\linewidth]{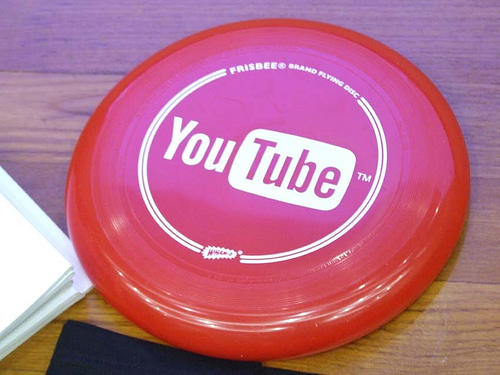} \\
    \includegraphics[width=0.14\linewidth]{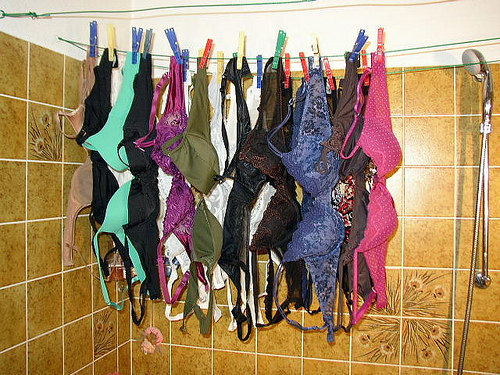} &
    \includegraphics[width=0.14\linewidth]{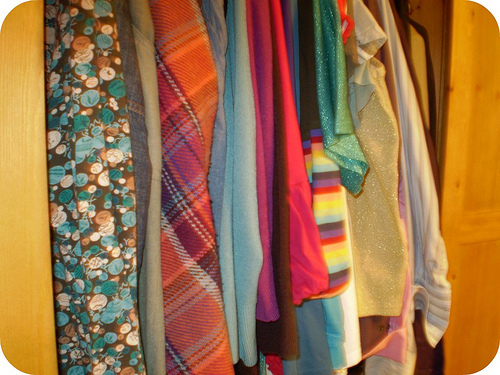} &
    \includegraphics[width=0.14\linewidth]{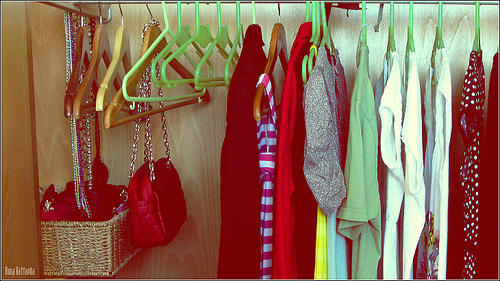} &
    \includegraphics[width=0.14\linewidth]{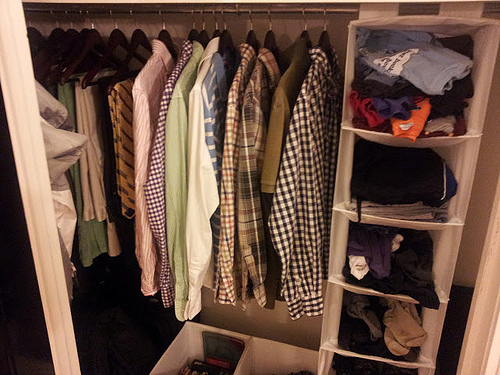} &
    \includegraphics[width=0.14\linewidth]{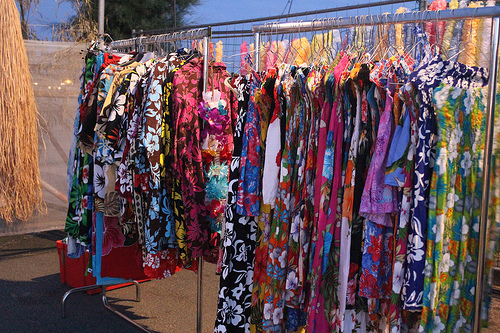} &
    \includegraphics[width=0.14\linewidth]{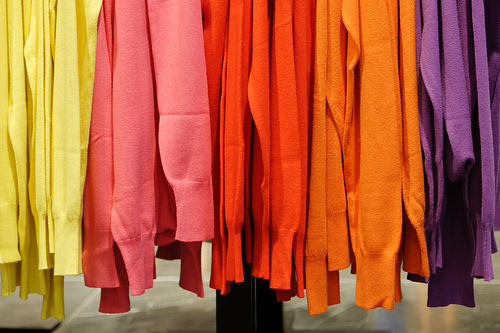} \\
    \includegraphics[width=0.14\linewidth]{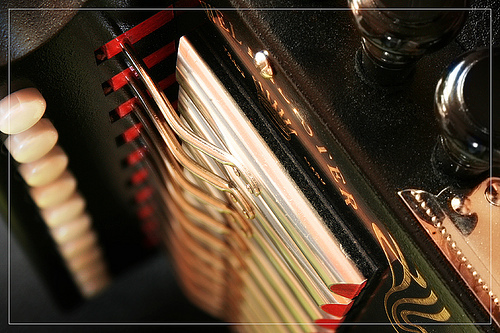} &
    \includegraphics[width=0.14\linewidth]{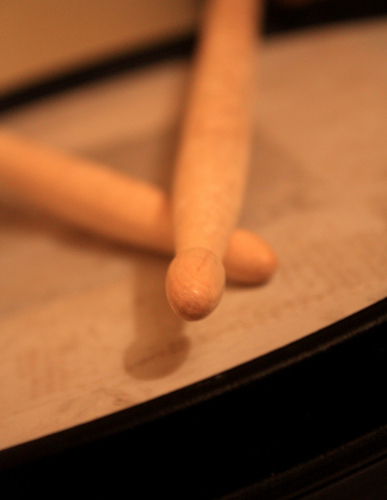} &
    \includegraphics[width=0.12\linewidth]{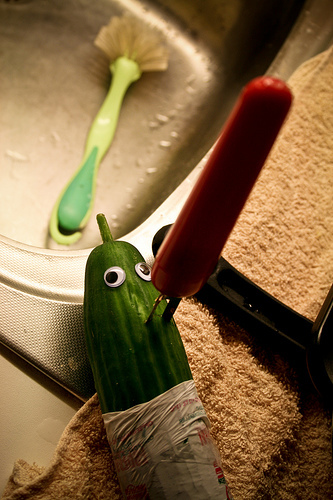} &
    \includegraphics[width=0.12\linewidth]{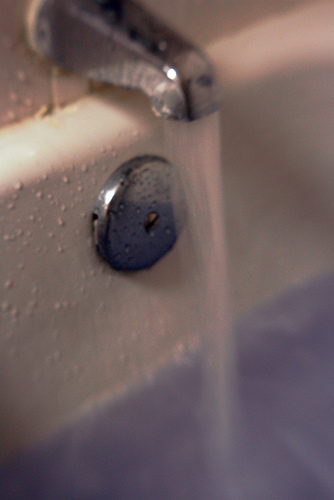} &
    \includegraphics[width=0.14\linewidth]{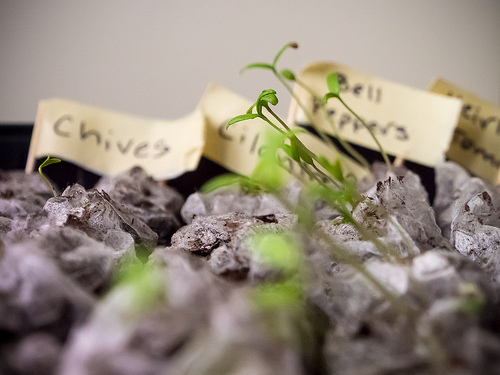} &
    \includegraphics[width=0.14\linewidth]{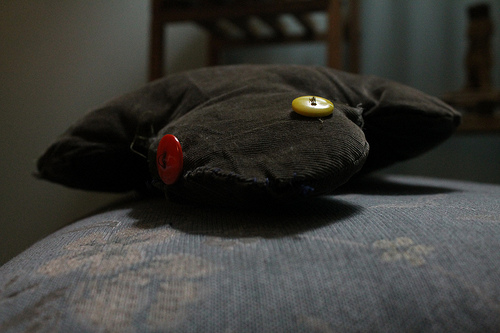} \\
    \includegraphics[width=0.14\linewidth]{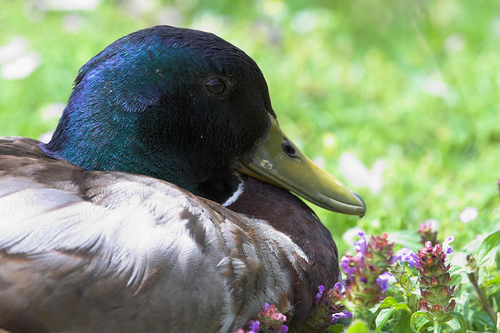} &
    \includegraphics[width=0.14\linewidth]{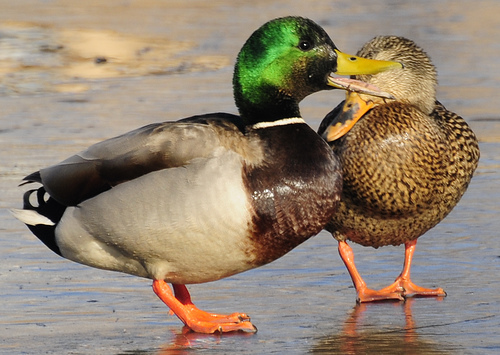} &
    \includegraphics[width=0.14\linewidth]{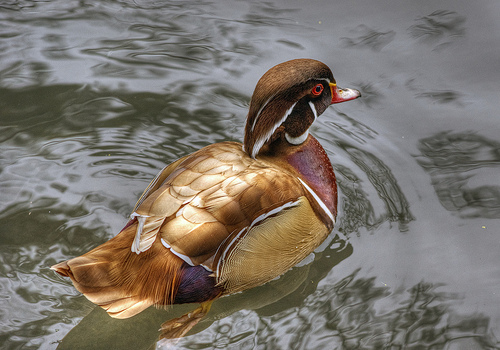} &
    \includegraphics[width=0.14\linewidth]{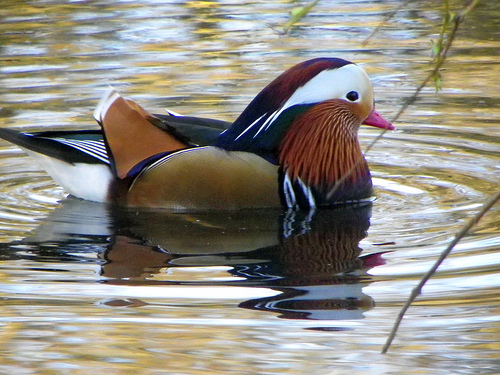} &
    \includegraphics[width=0.14\linewidth]{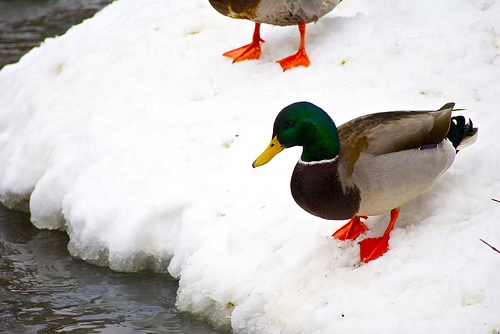} &
    \includegraphics[width=0.14\linewidth]{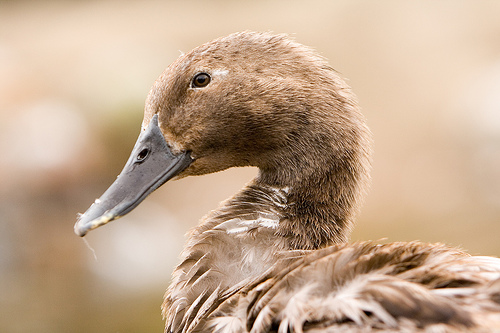} \\
    \includegraphics[width=0.14\linewidth]{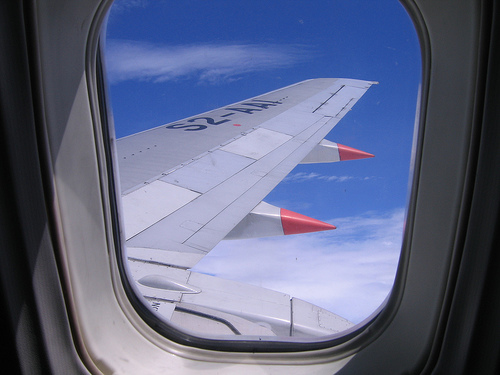} &
    \includegraphics[width=0.14\linewidth]{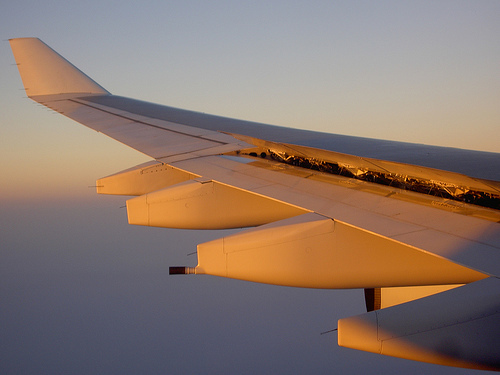} &
    \includegraphics[width=0.14\linewidth]{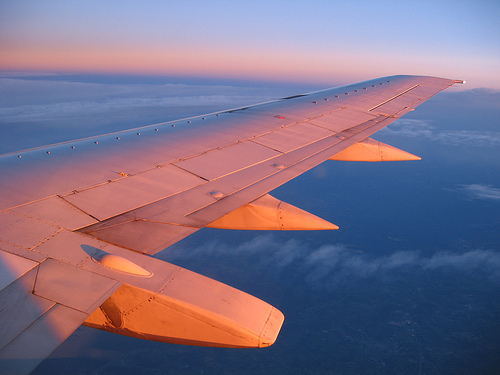} &
    \includegraphics[width=0.14\linewidth]{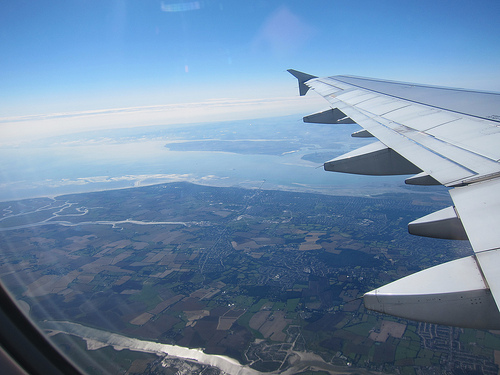} &
    \includegraphics[width=0.14\linewidth]{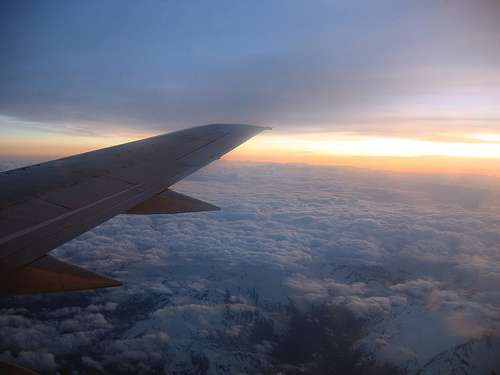} &
    \includegraphics[width=0.14\linewidth]{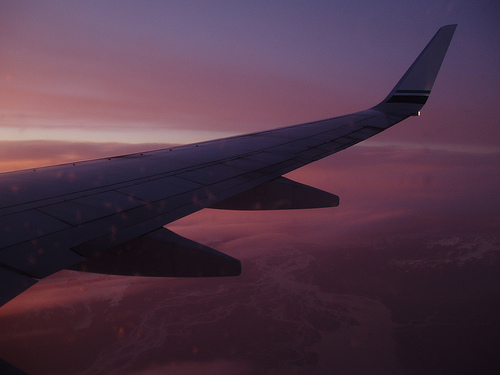} \\
    \end{tabular}
    \caption{Eight shared prototypes commonly found in both ImageNet and PASS: \textit{Car details}, \textit{Cars}, \textit{Cats}, \textit{Caps/lids}, \textit{Clothing rack}, \textit{Close-ups}, \textit{Ducks}, and \textit{Airplane wings} respectively. The first three columns show ImageNet images and the last three columns are PASS images.} 
\end{figure*}

\clearpage
\section{Additional Art Dataset Prototypes}

{\setlength{\tabcolsep}{0.4em} 
    \begin{figure}[!ht]
    \centering
    {\sffamily
    \begin{tabular}{dddddc}
    \multicolumn{6}{c}{MET prototypes} \\
    \includegraphics[width=0.15\linewidth]{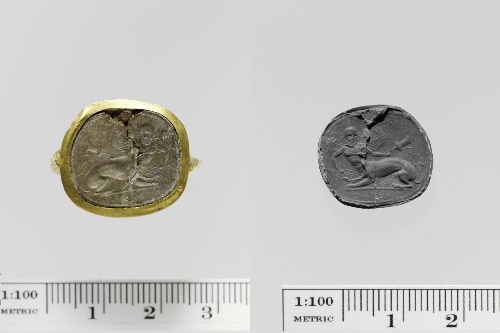} &
    \includegraphics[width=0.15\linewidth]{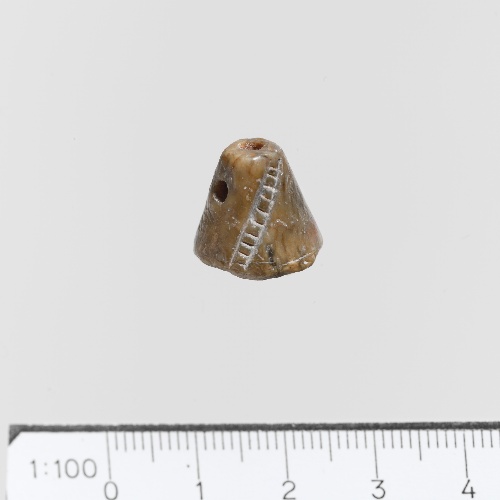} &
    \includegraphics[width=0.15\linewidth]{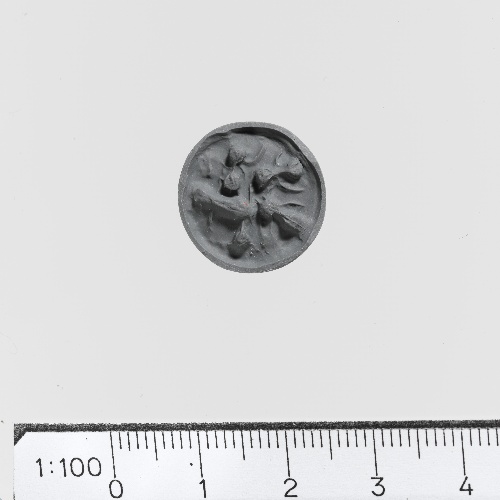} &
    \includegraphics[width=0.15\linewidth]{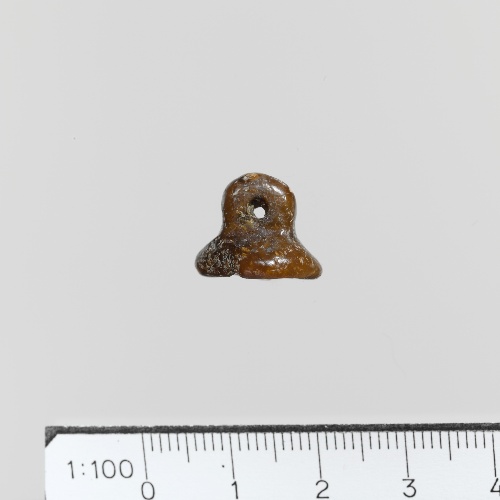} &
    \includegraphics[width=0.15\linewidth]{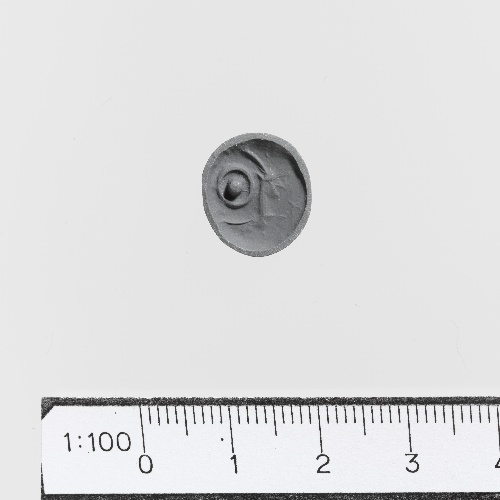} &
    \includegraphics[width=0.15\linewidth]{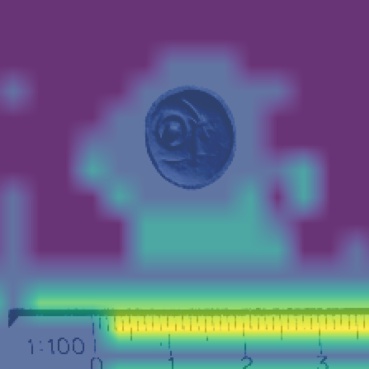} \\
    \includegraphics[width=0.13\linewidth]{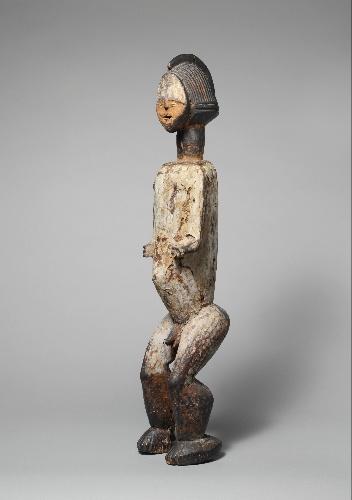} &
    \includegraphics[width=0.13\linewidth]{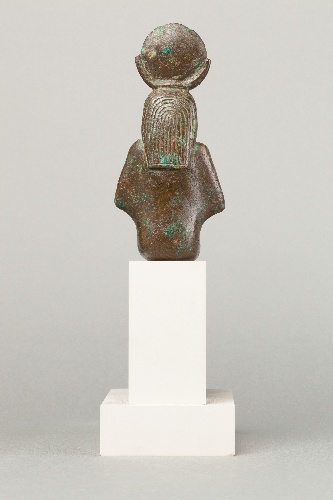} &
    \includegraphics[width=0.15\linewidth]{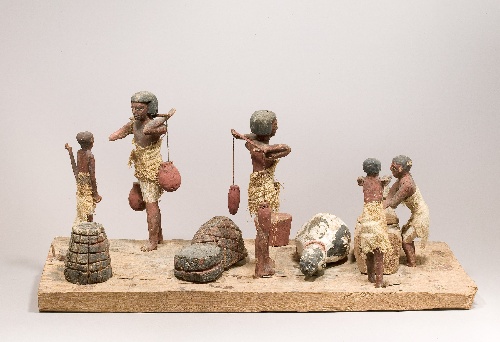} &
    \includegraphics[width=0.11\linewidth]{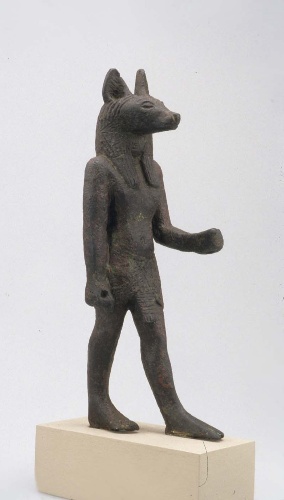} &
    \includegraphics[width=0.13\linewidth]{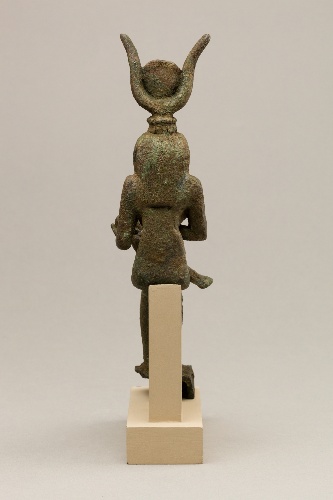} &
    \includegraphics[width=0.15\linewidth]{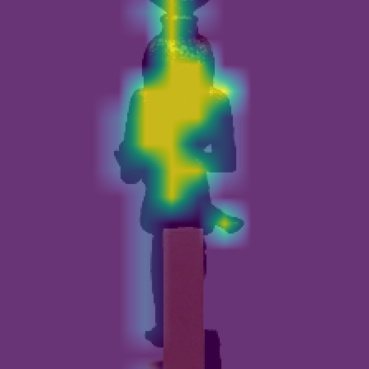} \\
    \includegraphics[width=0.15\linewidth]{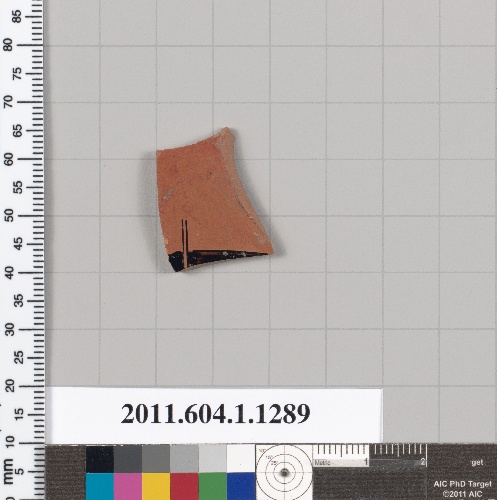} &
    \includegraphics[width=0.15\linewidth]{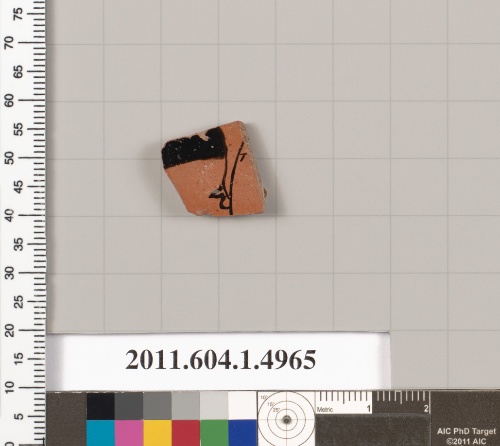} &
    \includegraphics[width=0.15\linewidth]{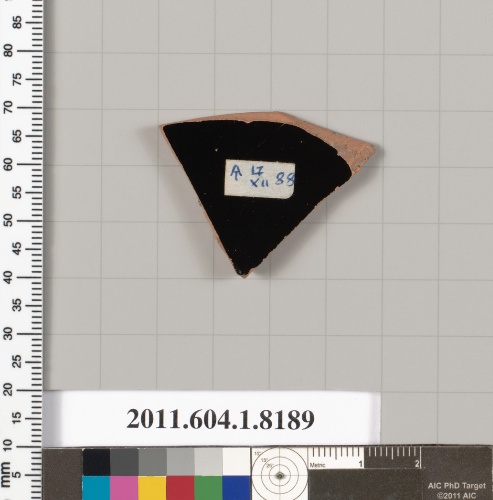} &
    \includegraphics[width=0.15\linewidth]{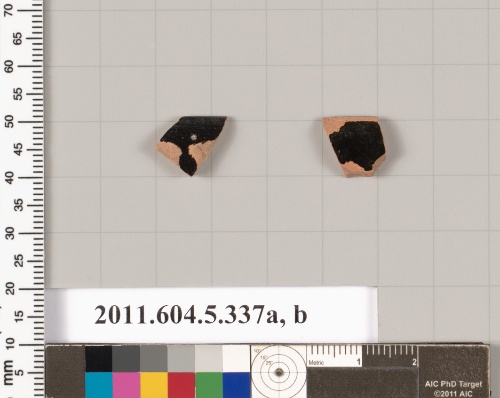} &
    \includegraphics[width=0.15\linewidth]{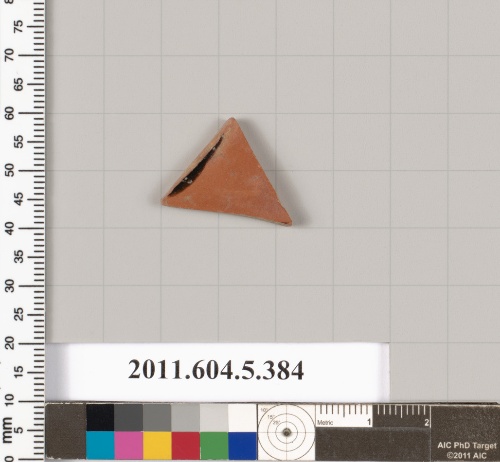} &
    \includegraphics[width=0.15\linewidth]{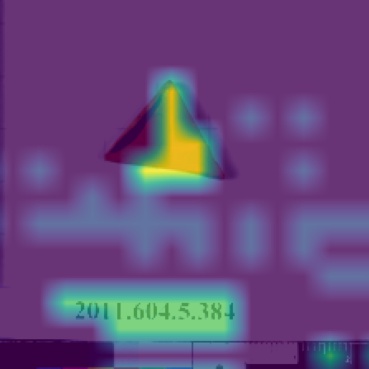} \\
    \end{tabular}}
    \vspace{.5em}
    \caption{Three predominantly MET prototypes: \textit{Tape measure}, \textit{Statues}, and \textit{Fragments with colour chart}. Per row five example images are shown, the sixth column contains the square-cropped attention mask of the prototype for the image in the fifth column.} 
    \label{fig:supp_met}
    \end{figure}

    \begin{figure}[!ht]
    \centering
    {\sffamily
    \begin{tabular}{bbbbbc}
    \multicolumn{6}{c}{Rijksmuseum prototypes} \\
    \includegraphics[width=0.15\linewidth]{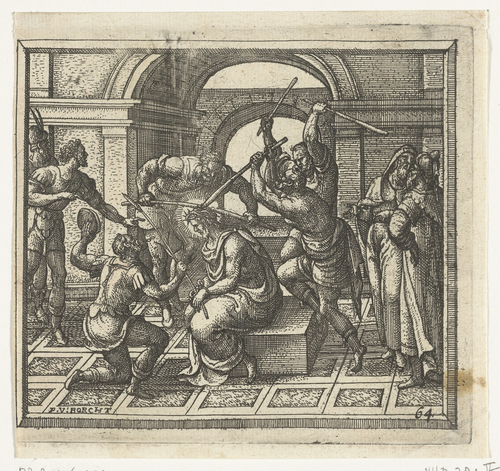} &
    \includegraphics[width=0.13\linewidth]{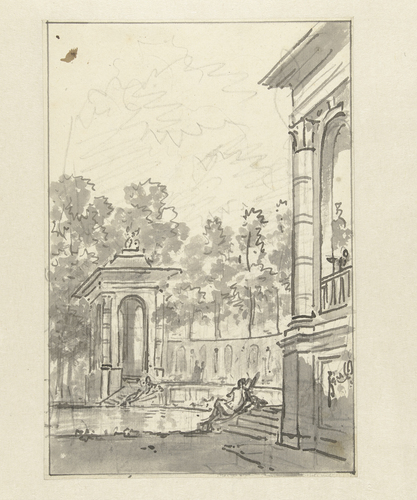} &
    \includegraphics[width=0.15\linewidth]{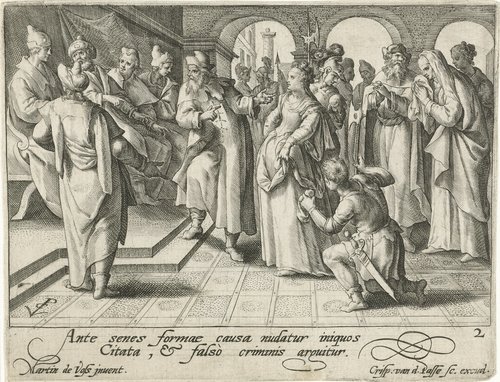} &
    \includegraphics[width=0.15\linewidth]{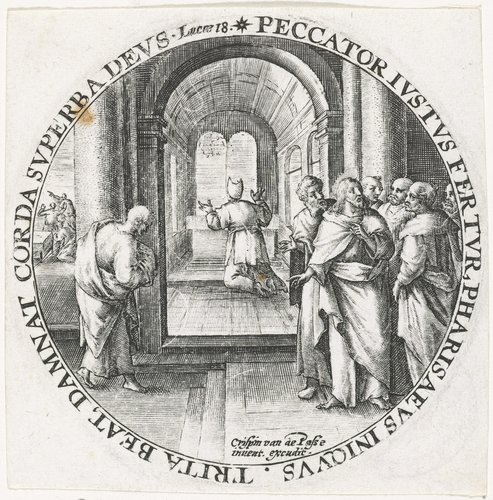} &
    \includegraphics[width=0.15\linewidth]{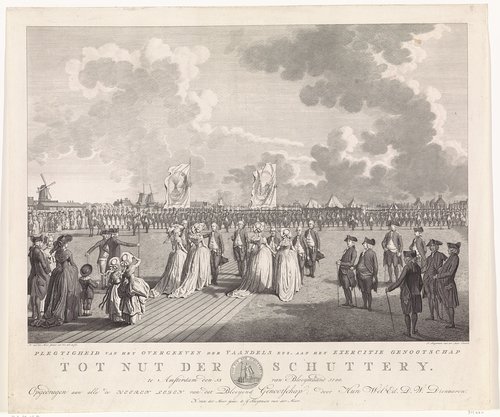} &
    \includegraphics[width=0.15\linewidth]{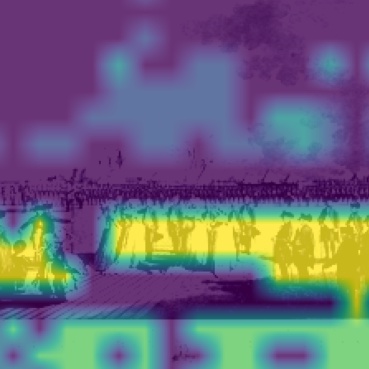} \\
    \includegraphics[width=0.15\linewidth]{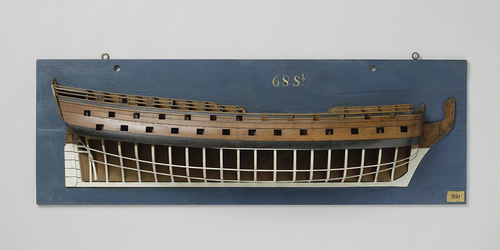} &
    \includegraphics[width=0.15\linewidth]{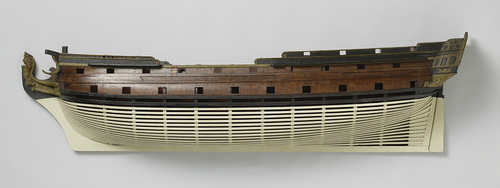} &
    \includegraphics[width=0.15\linewidth]{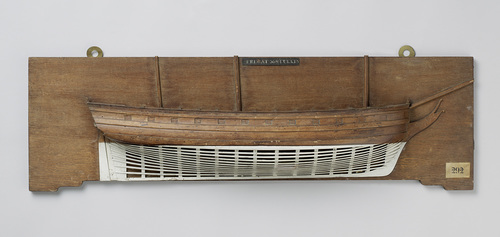} &
    \includegraphics[width=0.15\linewidth]{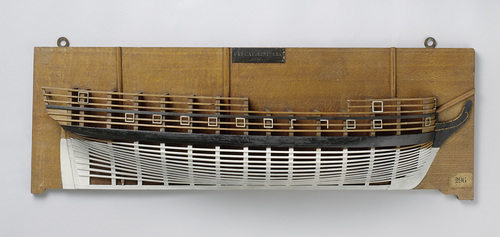} &
    \includegraphics[width=0.15\linewidth]{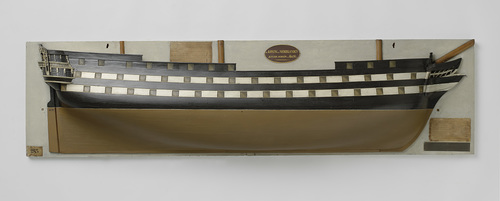} &
    \includegraphics[width=0.15\linewidth]{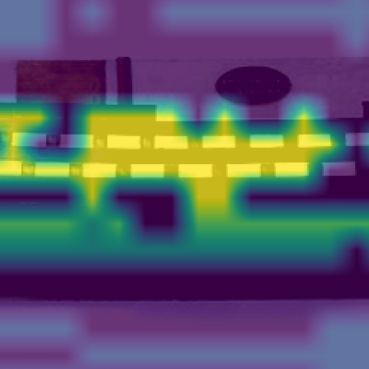} \\
    \includegraphics[width=0.15\linewidth]{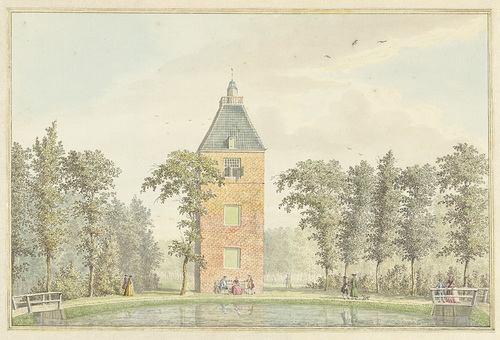} &
    \includegraphics[width=0.15\linewidth]{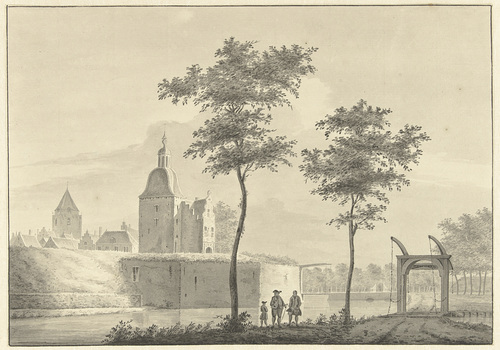} &
    \includegraphics[width=0.15\linewidth]{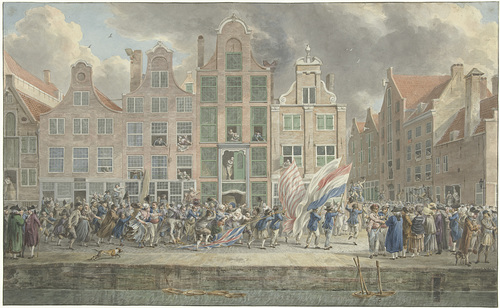} &
    \includegraphics[width=0.15\linewidth]{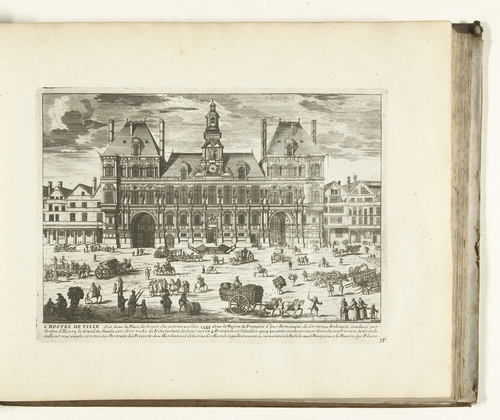} &
    \includegraphics[width=0.15\linewidth]{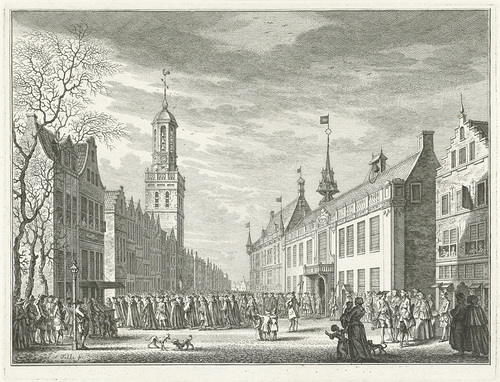} &
    \includegraphics[width=0.15\linewidth]{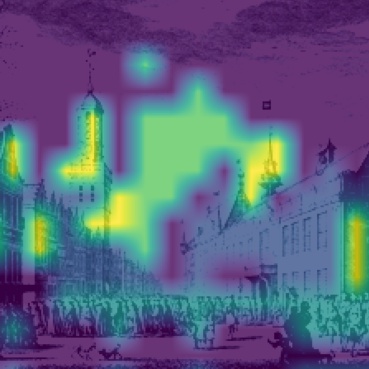} \\
    \end{tabular}}
    \vspace{.5em}
    \caption{Three predominantly Rijksmuseum prototypes: \textit{Crowds in prints}, \textit{Model boats}, and \textit{Scene with tower(s)}. Per row five example images are shown, the sixth column contains the square-cropped attention mask of the prototype for the image in the fifth column.} 
    \label{fig:supp_rijks}
    \end{figure}

    \begin{figure}[!ht]
    \centering
    {\sffamily
    \begin{tabular}{aaaaac}
    \multicolumn{6}{c}{SemArt prototypes} \\
    \includegraphics[width=0.15\linewidth]{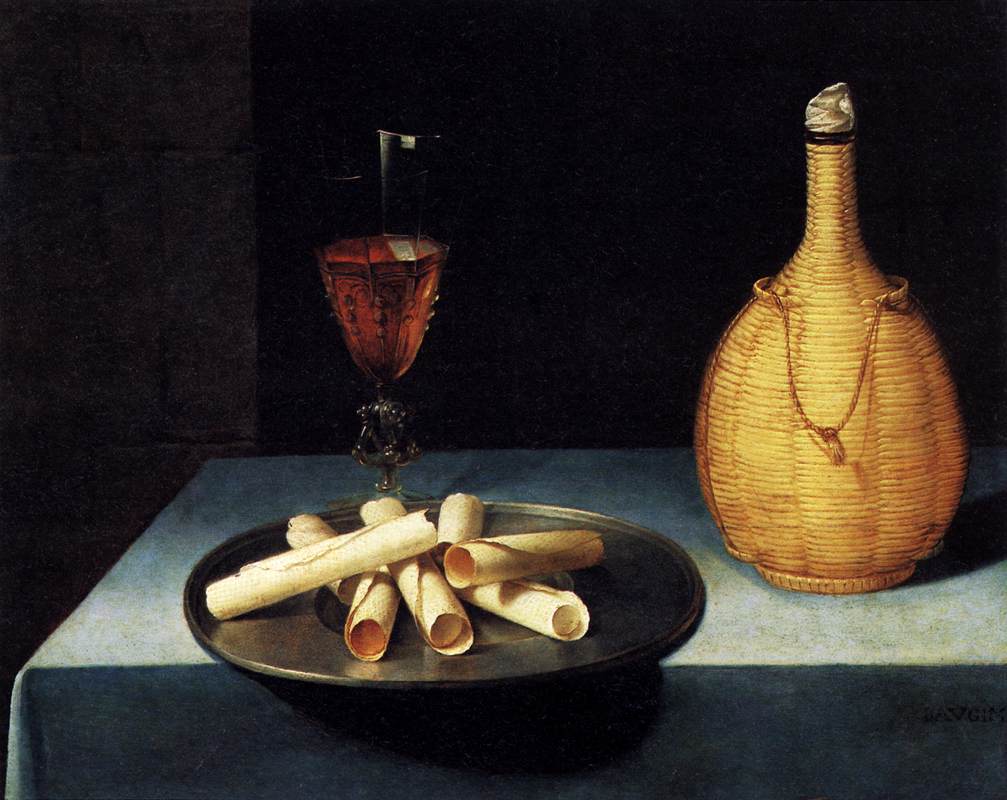} &
    \includegraphics[width=0.15\linewidth]{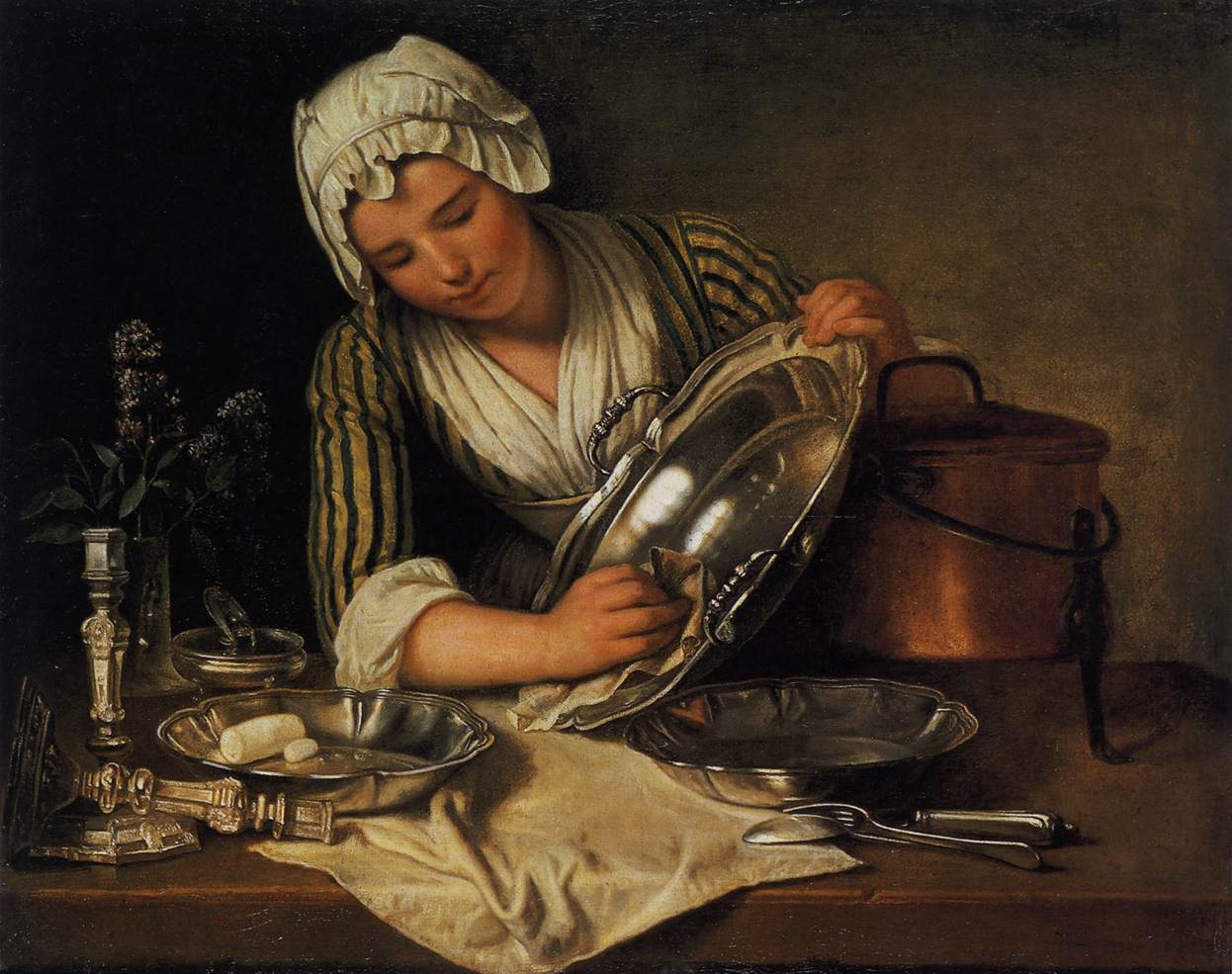} &
    \includegraphics[width=0.15\linewidth]{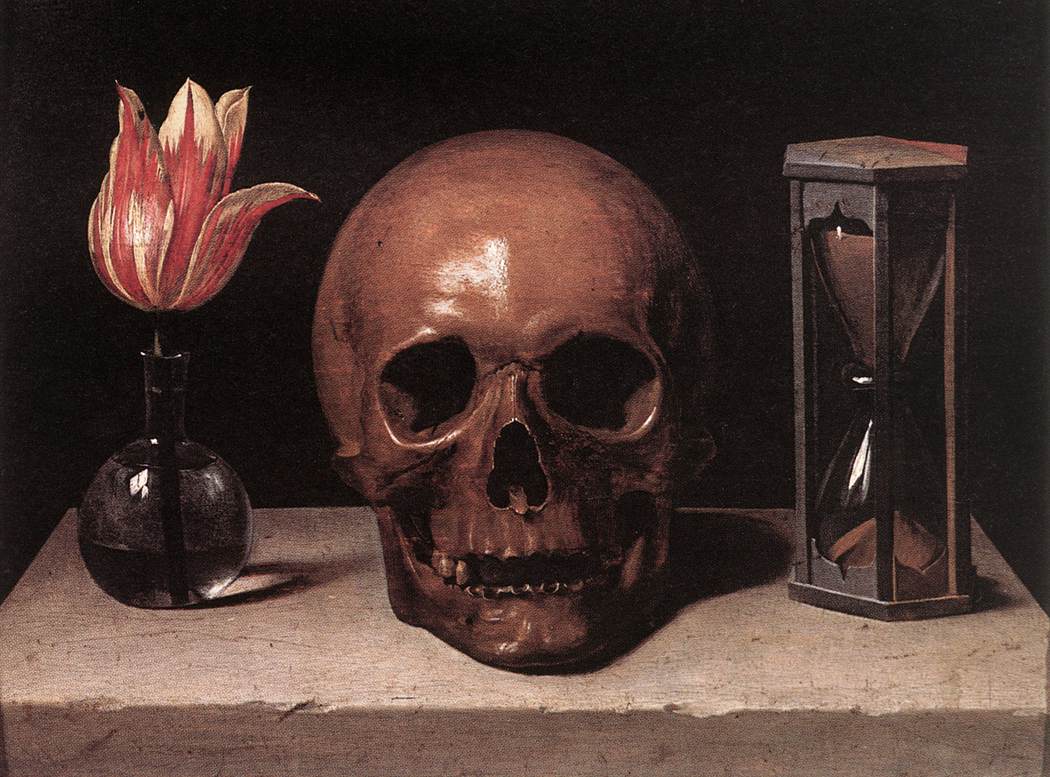} &
    \includegraphics[width=0.15\linewidth]{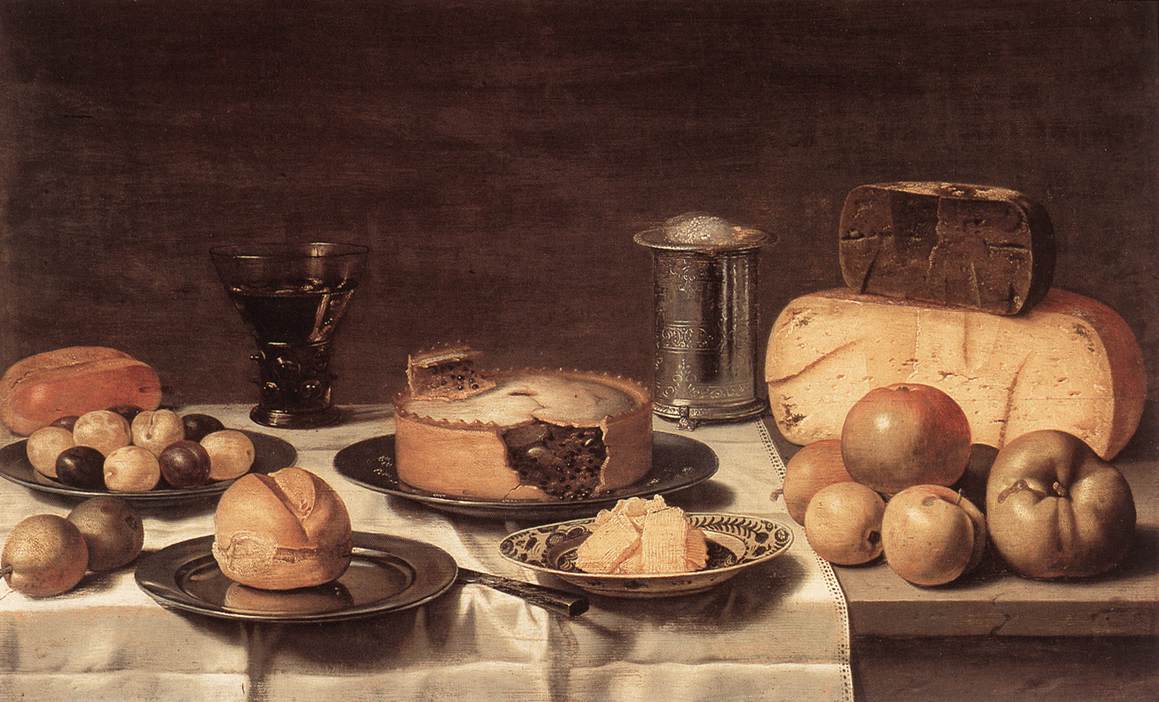} &
    \includegraphics[width=0.13\linewidth]{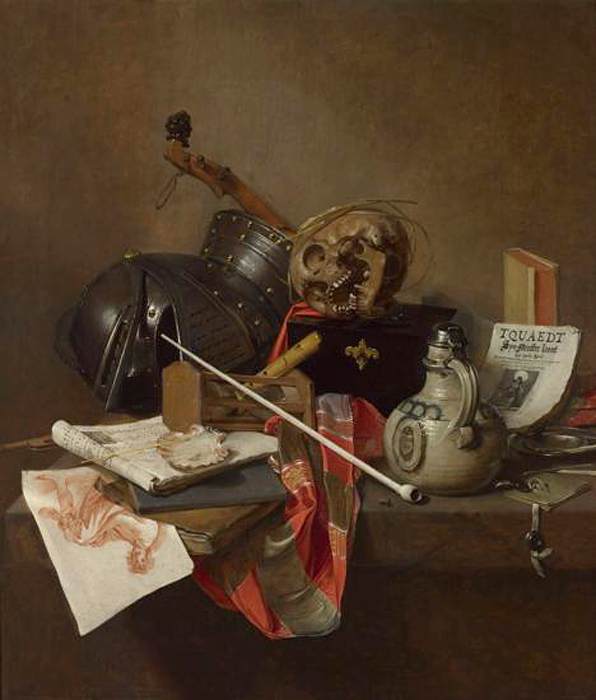} &
    \includegraphics[width=0.15\linewidth]{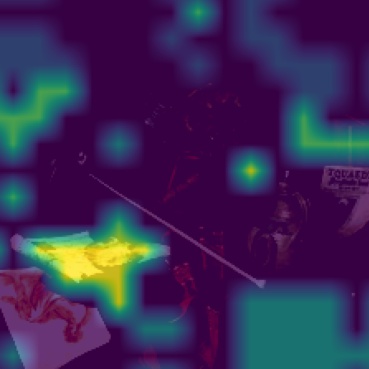} \\
    \includegraphics[width=0.15\linewidth]{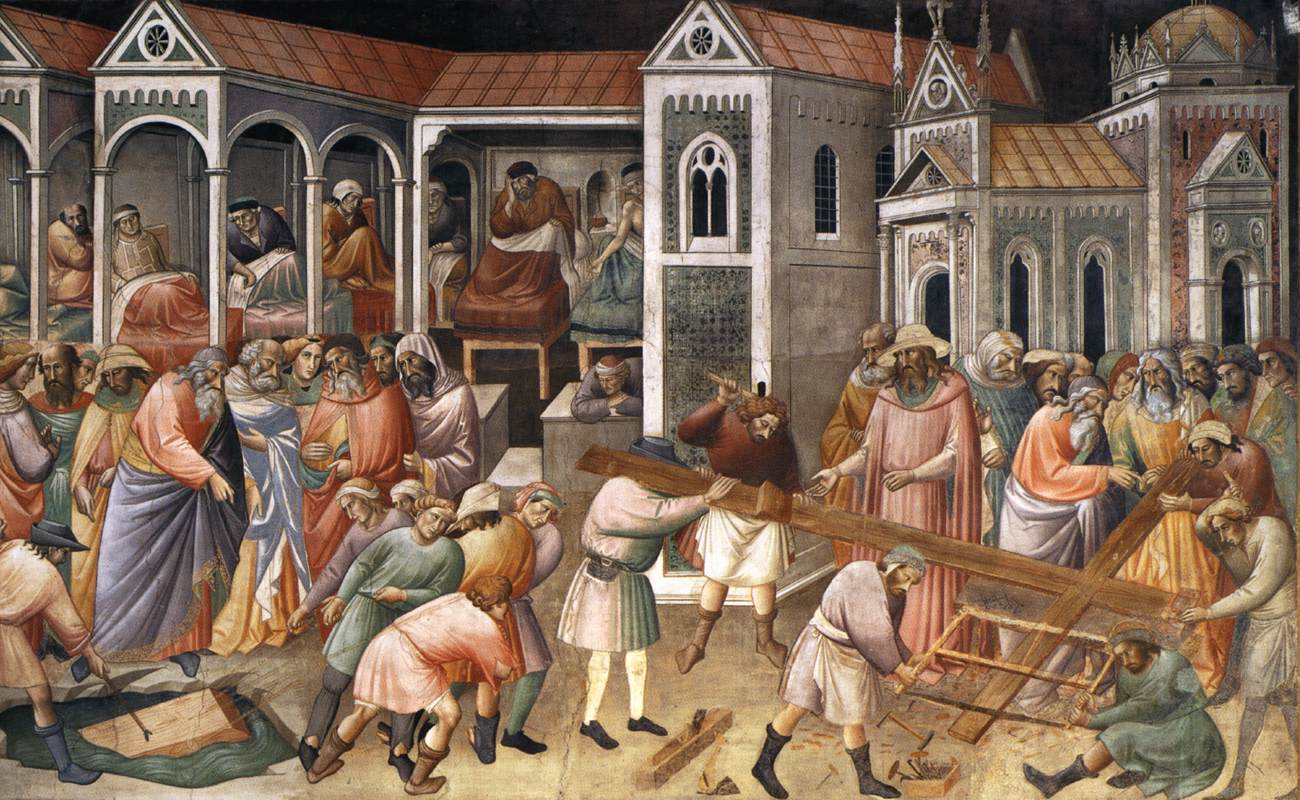} &
    \includegraphics[width=0.13\linewidth]{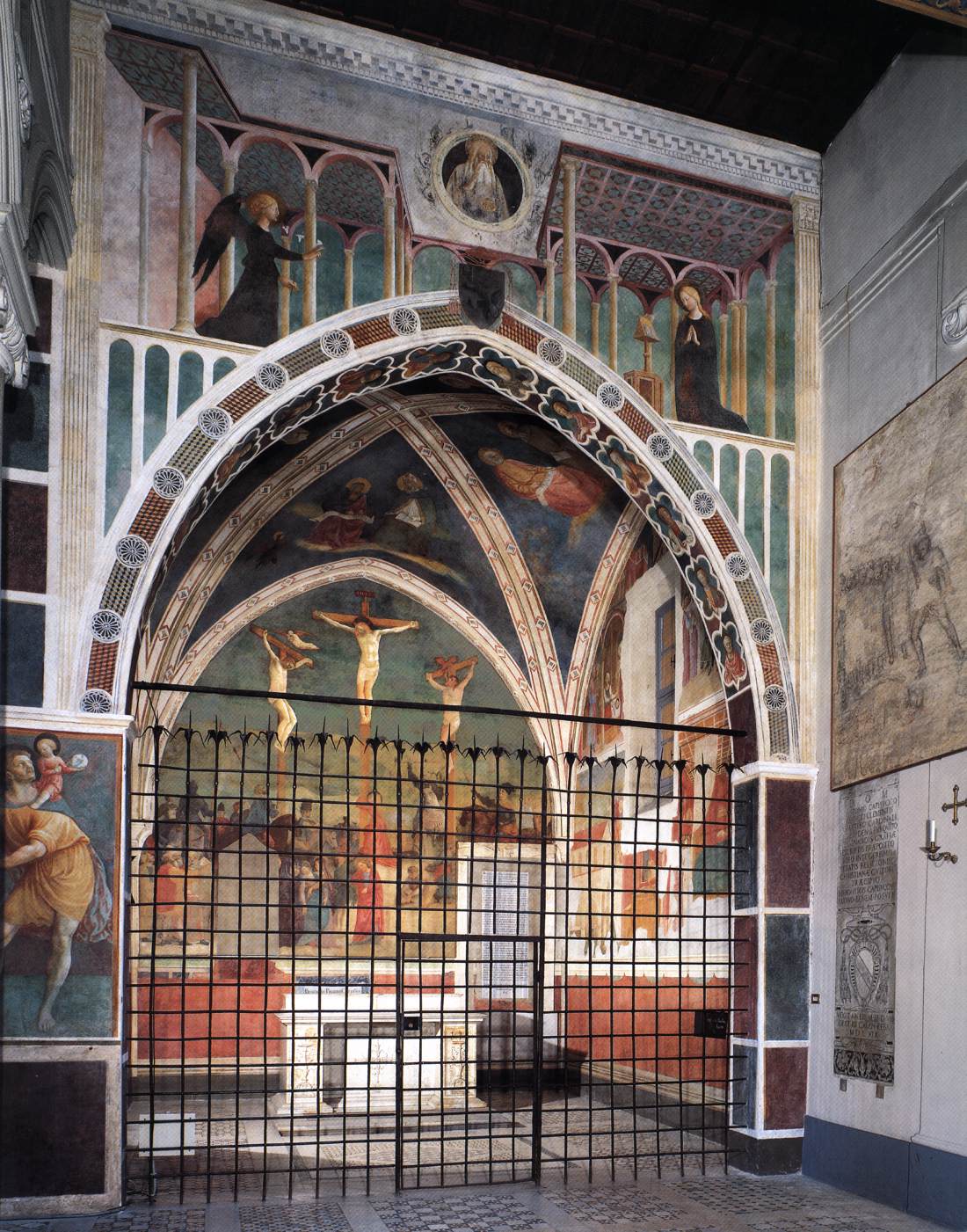} &
    \includegraphics[width=0.13\linewidth]{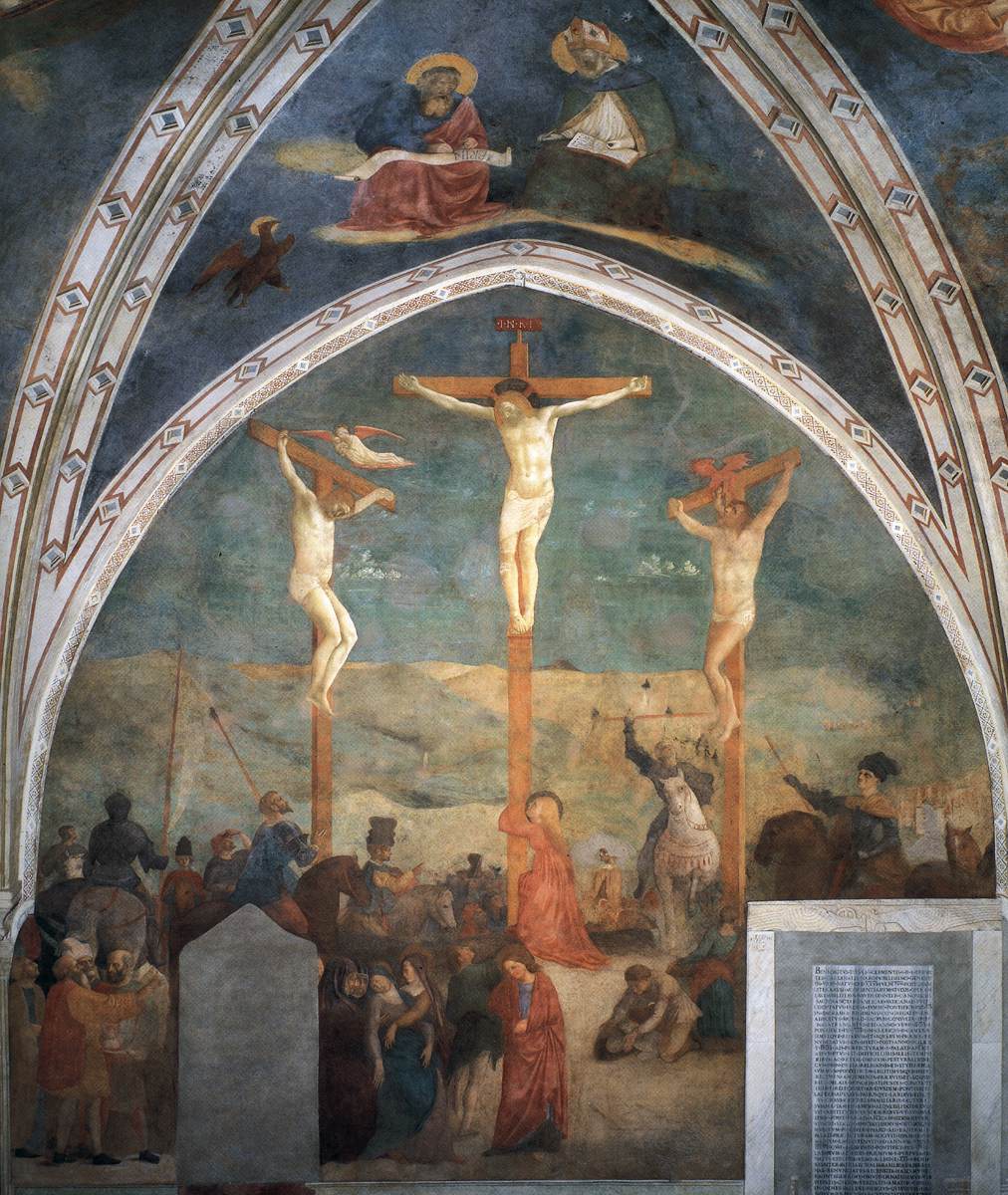} &
    \includegraphics[width=0.15\linewidth]{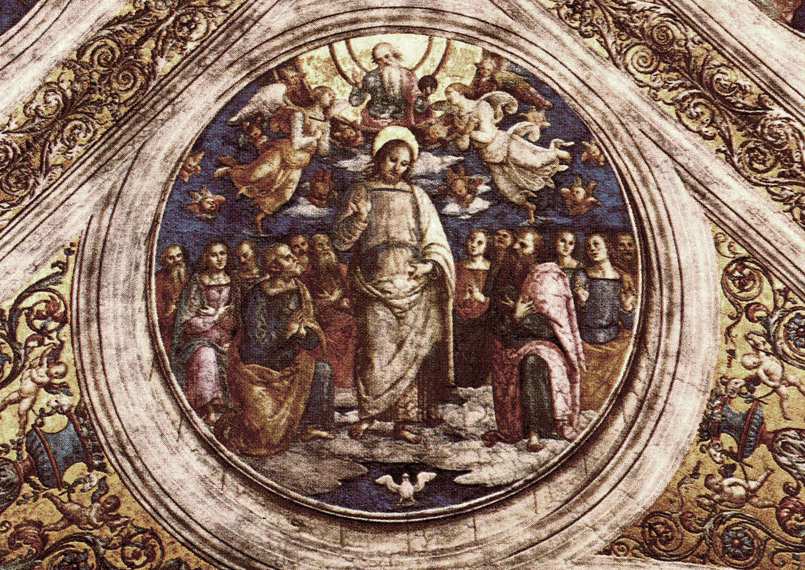} &
    \includegraphics[width=0.15\linewidth]{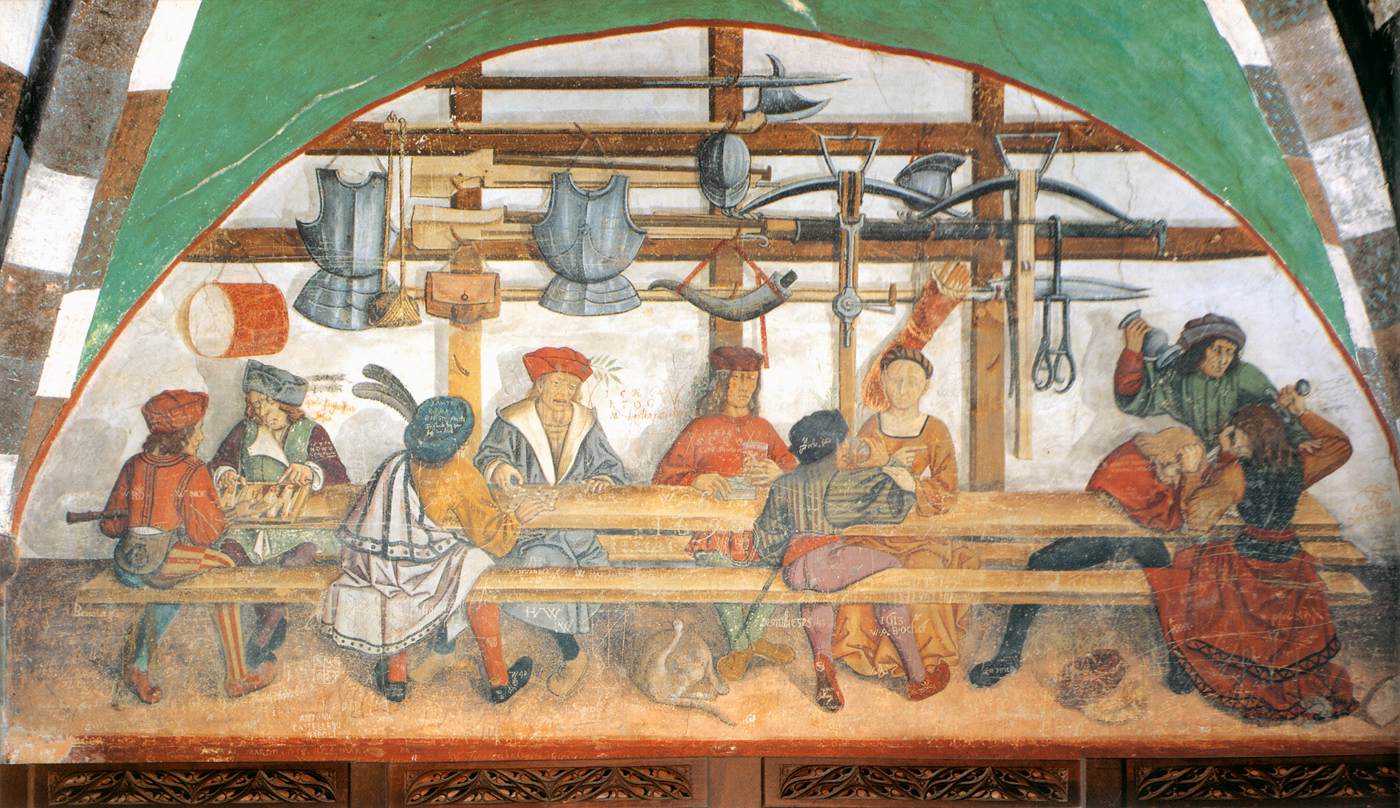} &
    \includegraphics[width=0.15\linewidth]{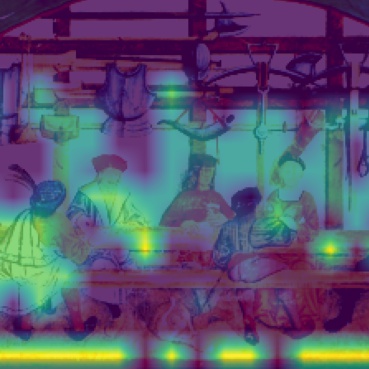} \\
    \includegraphics[width=0.15\linewidth]{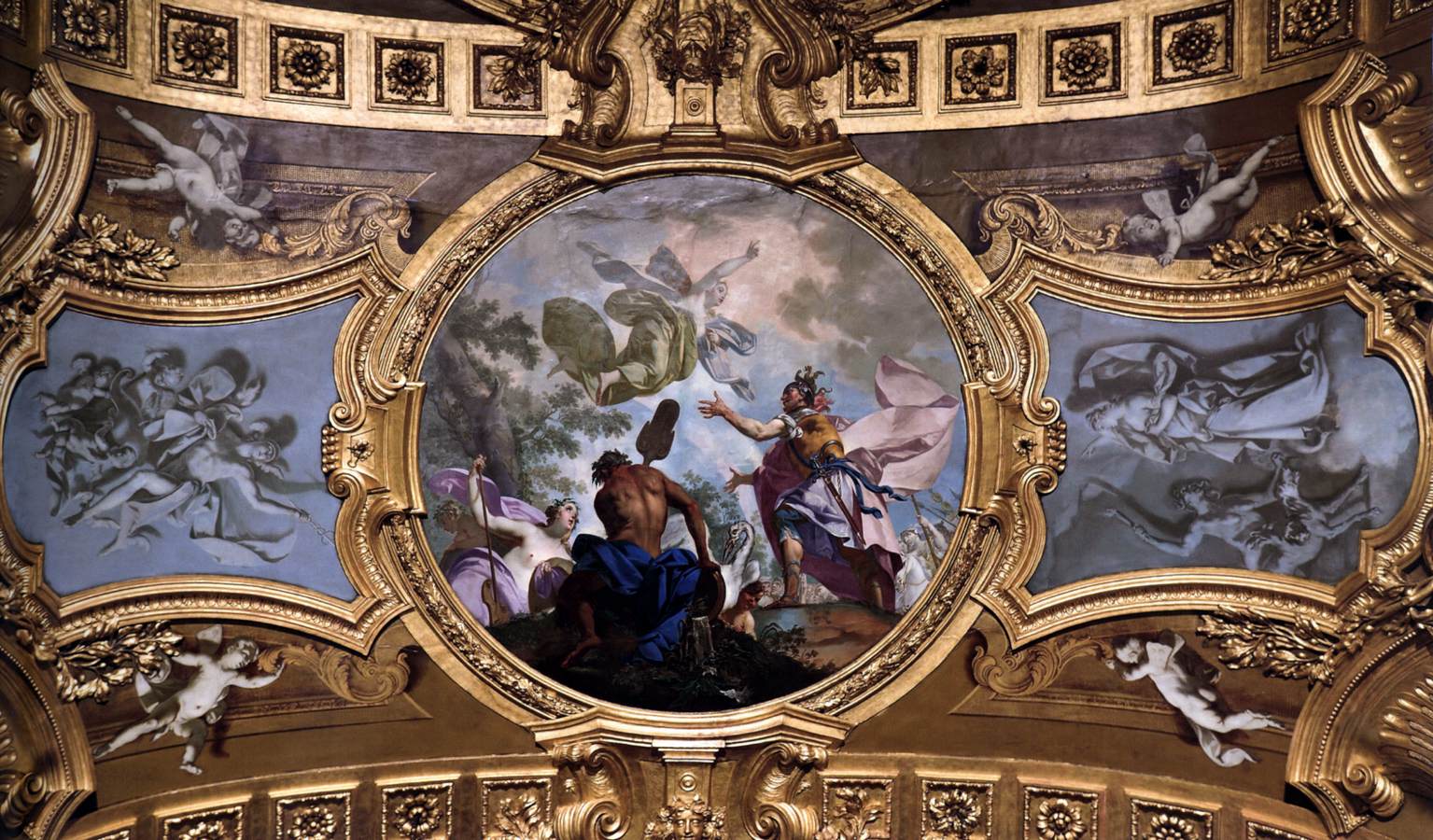} &
    \includegraphics[width=0.15\linewidth]{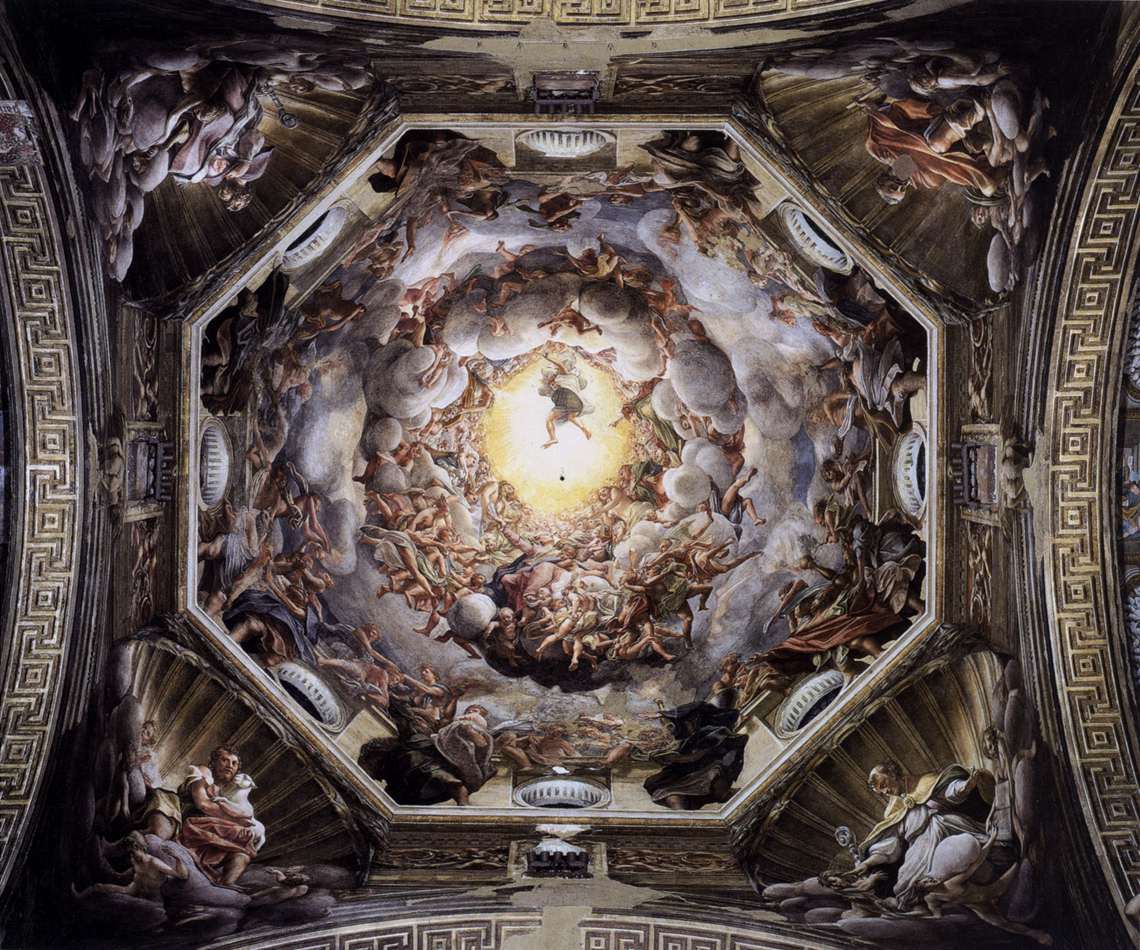} &
    \includegraphics[width=0.13\linewidth]{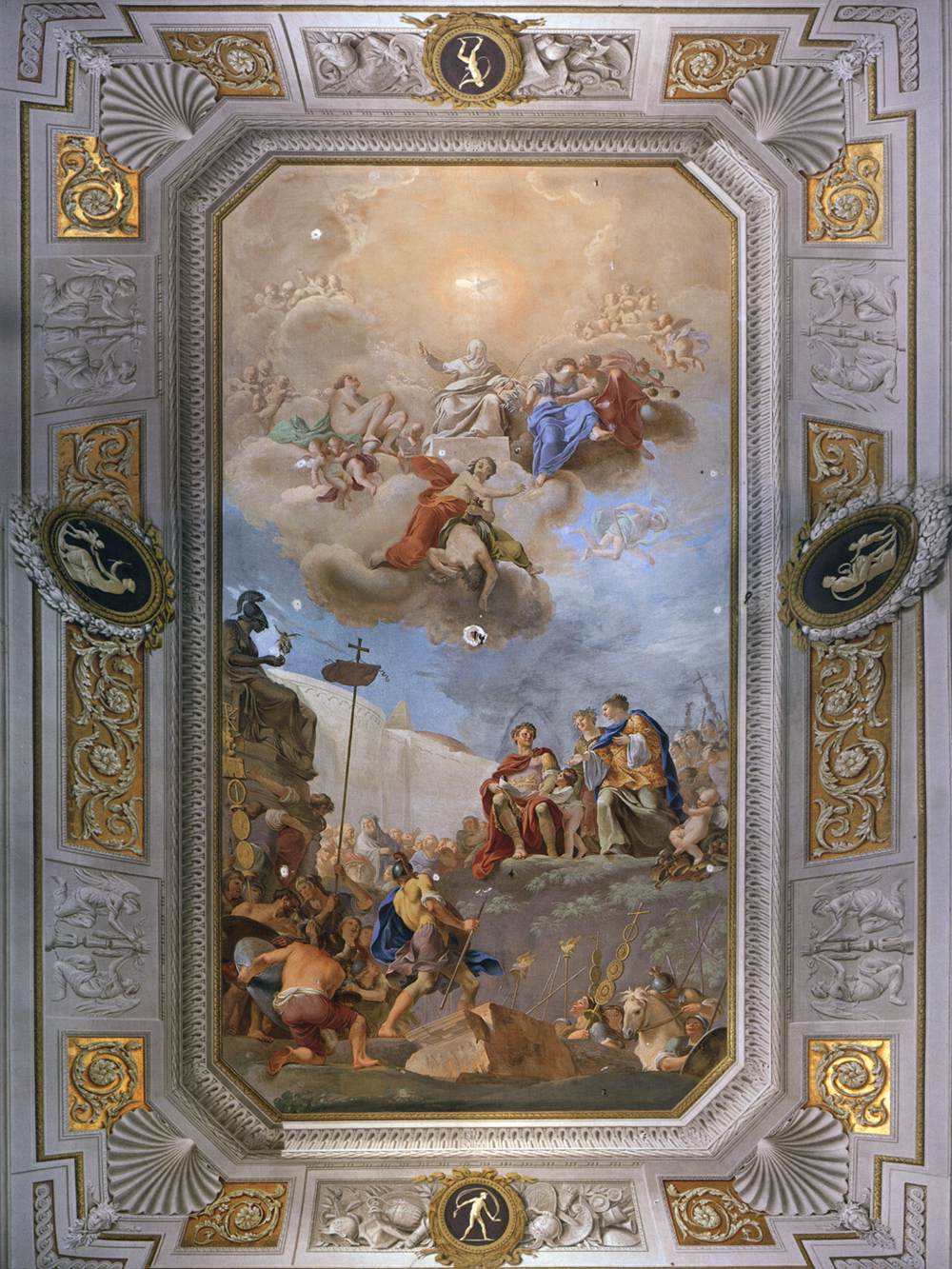} &
    \includegraphics[width=0.13\linewidth]{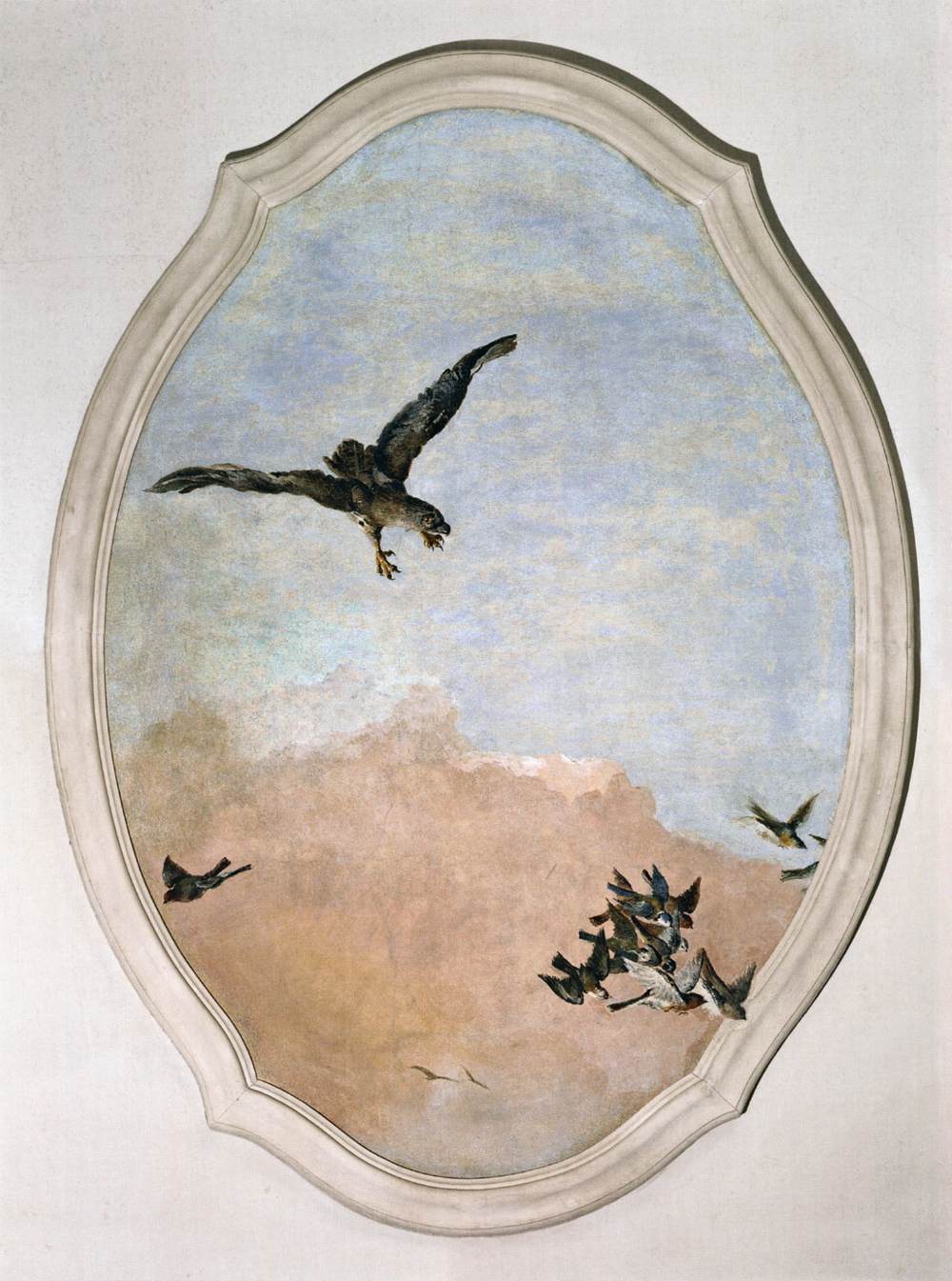} &
    \includegraphics[width=0.15\linewidth]{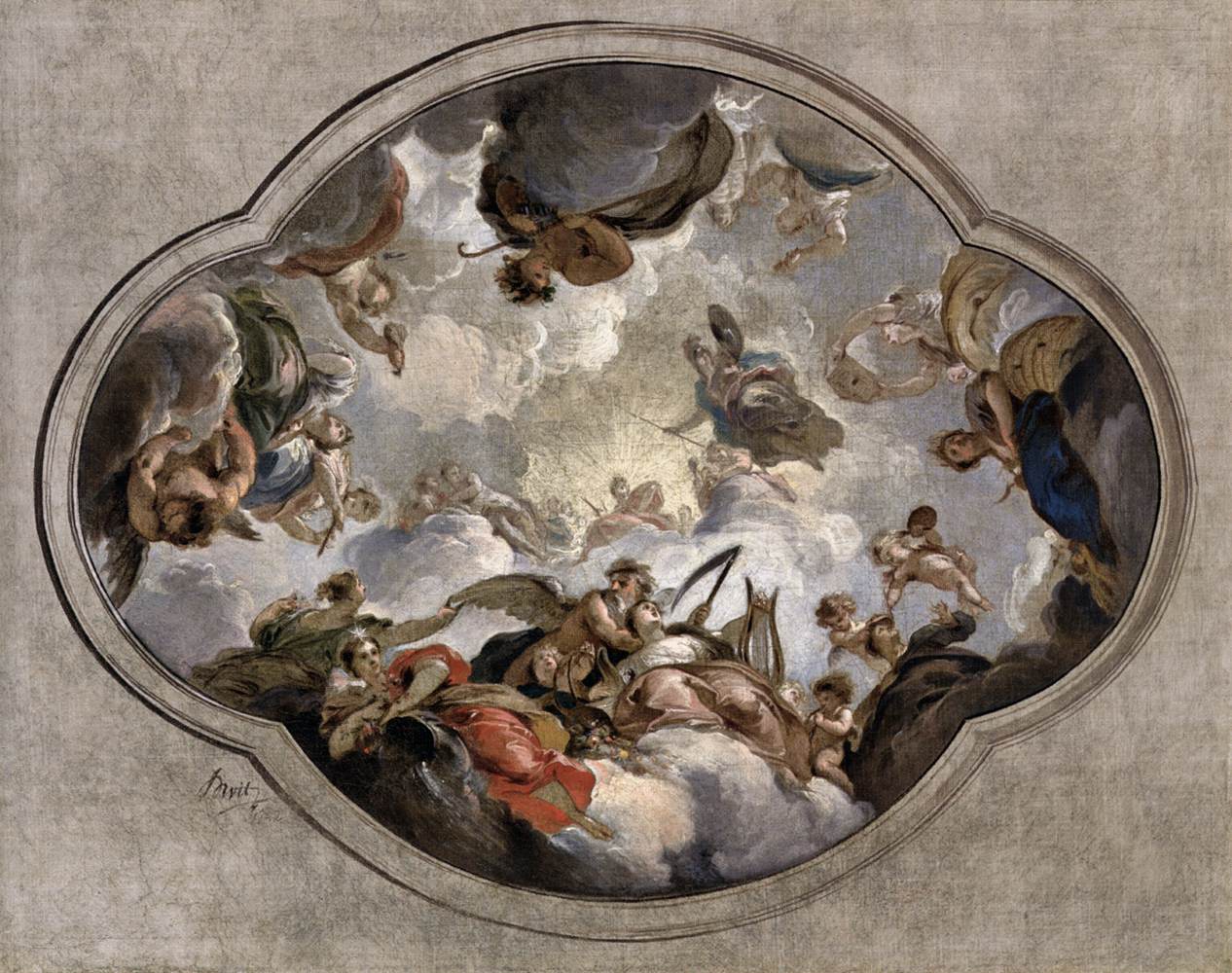} &
    \includegraphics[width=0.15\linewidth]{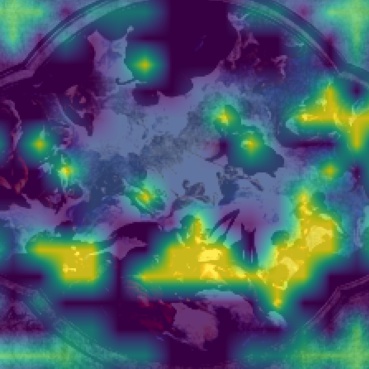} \\
    \end{tabular}}
    \vspace{.5em}
    \caption{Three predominantly SemArt prototypes: \textit{Still life painting}, \textit{Panel painting}, and \textit{Ceiling Fresco}. Per row five example images are shown, the sixth column contains the square-cropped attention mask of the prototype for the image in the fifth column.} 
    \label{fig:supp_sem}
    \end{figure}
    
    \begin{figure}[!ht]
    \centering
    {\sffamily
    \begin{tabular}{ddbbaa}
    \multicolumn{6}{c}{Shared prototypes} \\
    \includegraphics[width=0.15\linewidth]{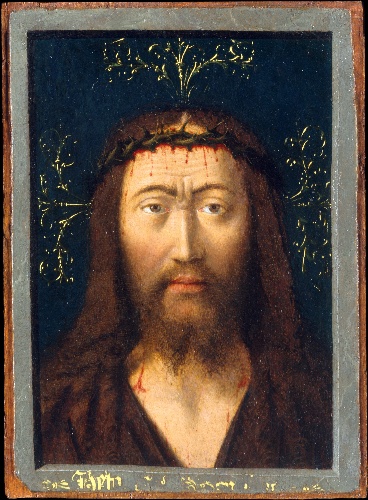} &
    \includegraphics[width=0.15\linewidth]{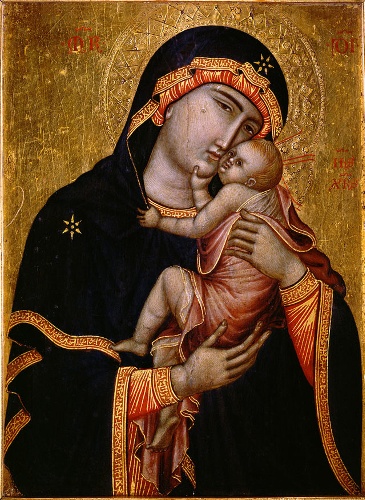} &
    \includegraphics[width=0.15\linewidth]{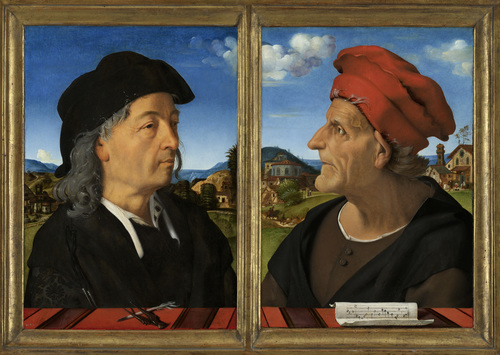} &
    \includegraphics[width=0.15\linewidth]{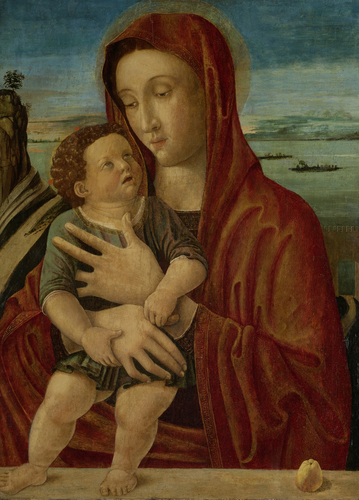} &
    \includegraphics[width=0.15\linewidth]{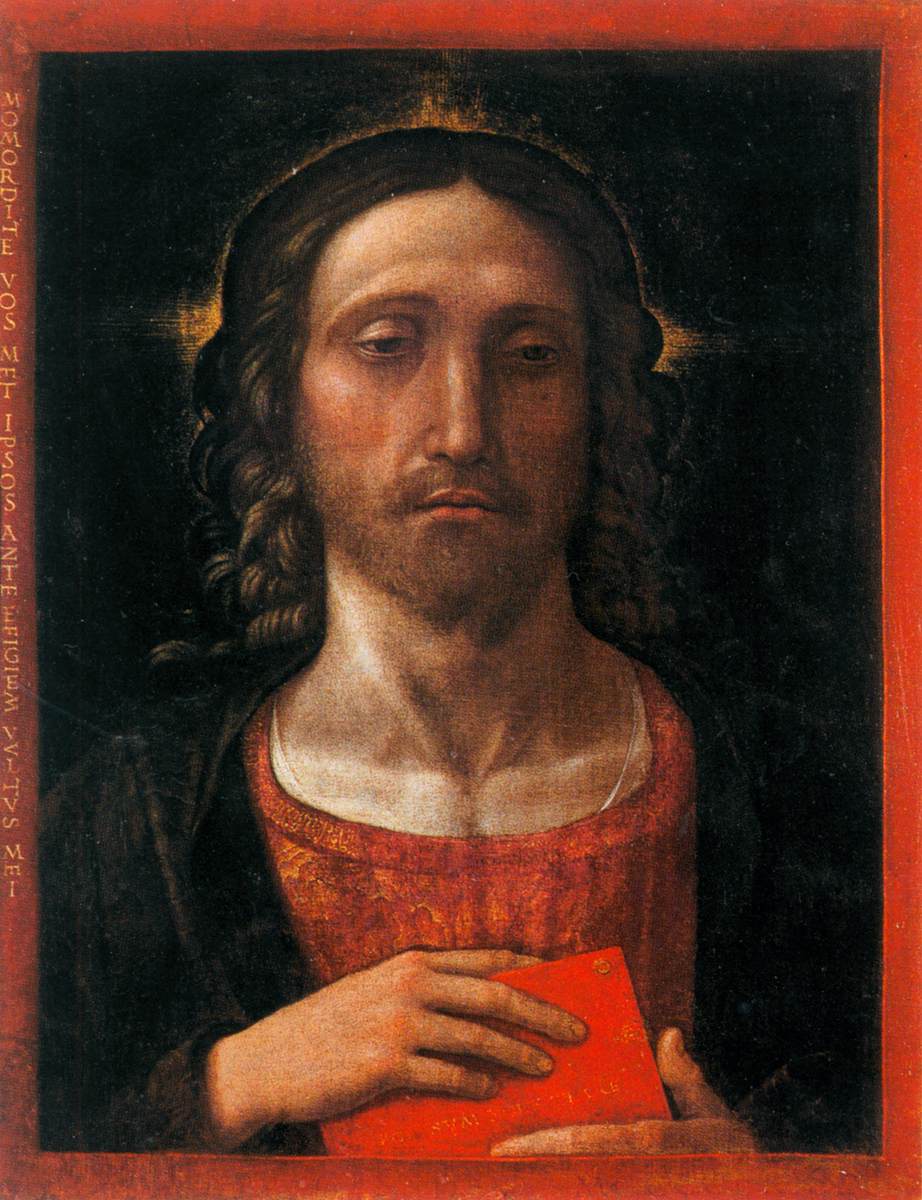} &
    \includegraphics[width=0.15\linewidth]{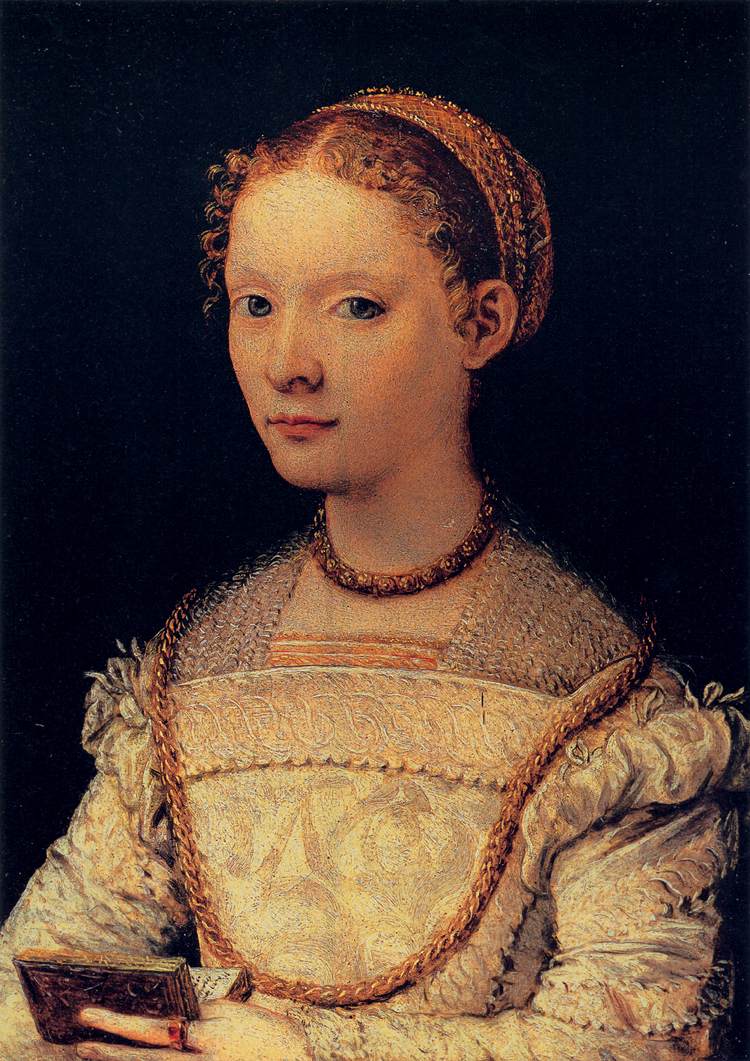} \\
    \includegraphics[width=0.15\linewidth]{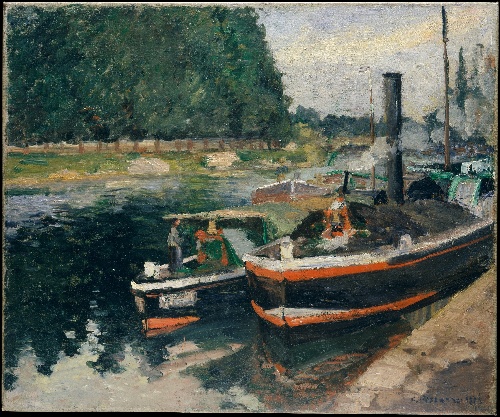} &
    \includegraphics[width=0.15\linewidth]{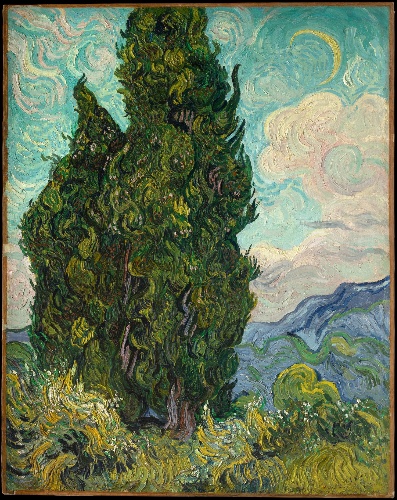} &
    \includegraphics[width=0.15\linewidth]{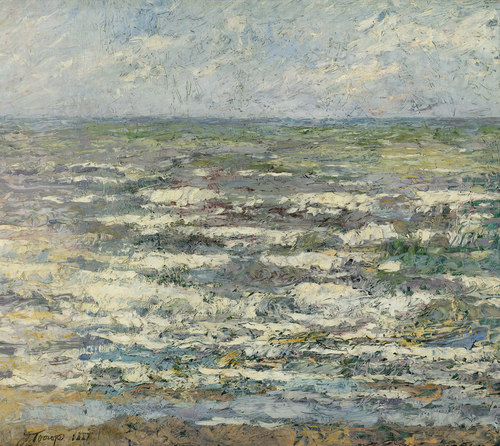} &
    \includegraphics[width=0.15\linewidth]{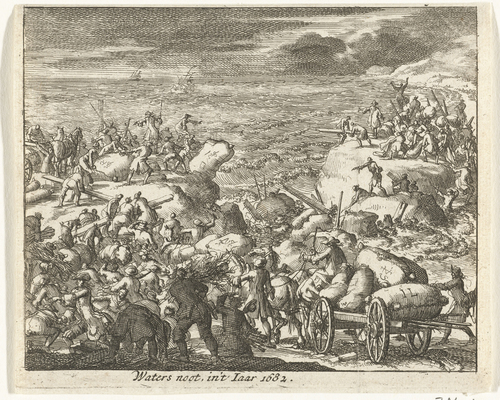} &
    \includegraphics[width=0.15\linewidth]{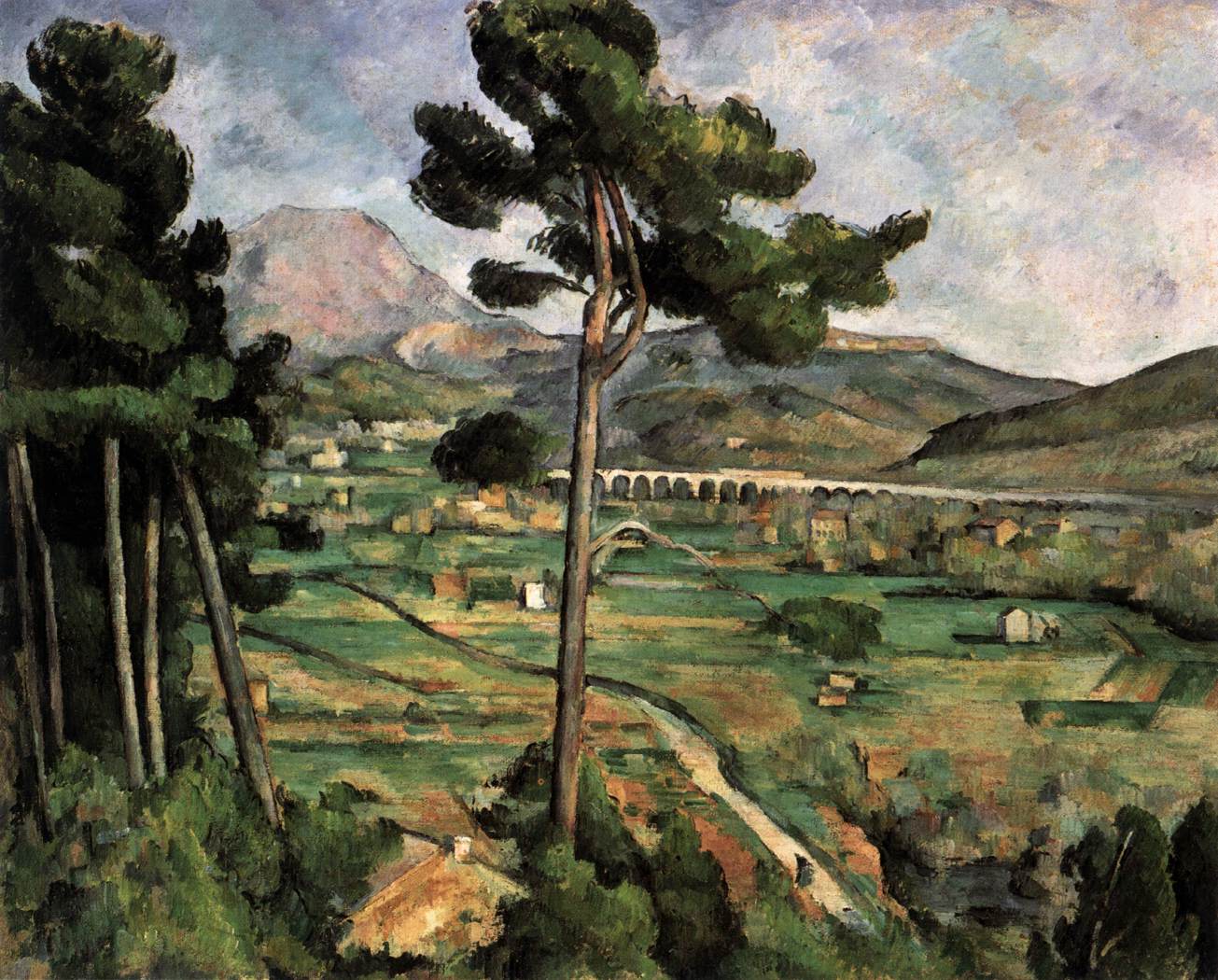} &
    \includegraphics[width=0.15\linewidth]{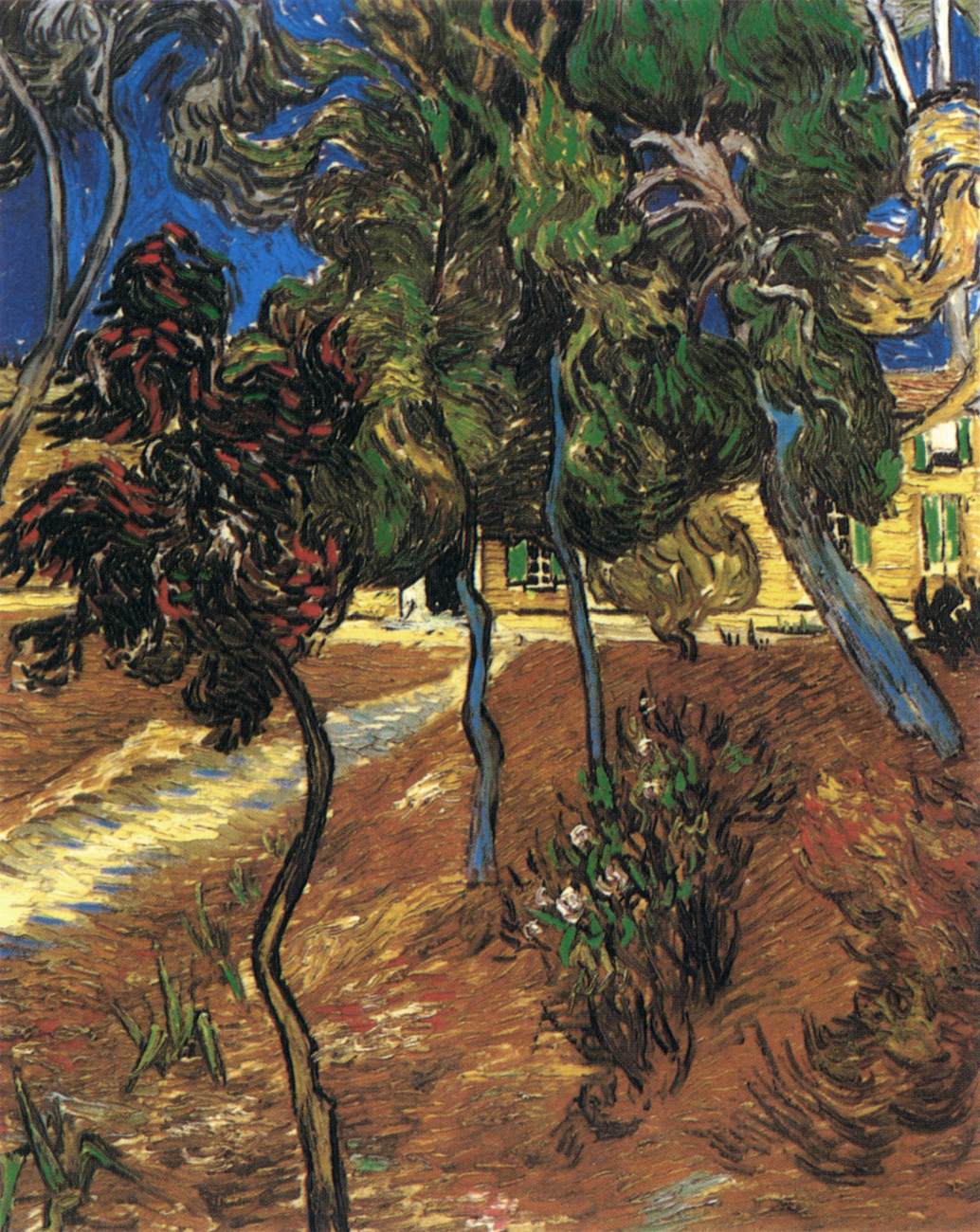} \\
    
    \end{tabular}}
    \caption{Two shared prototypes: \textit{Icon paintings}, \textit{Impressionist paintings}. The first two columns contain examples from MET, the second two from Rijksmuseum, and the last two columns from SemArt. The second prototype is rarely found in the Rijksmuseum dataset, mostly activating for drawings (as shown in second row, fourth image).} 
    \label{fig:artexamples}
\end{figure}}

\end{document}